\documentclass[nohyperref]{article}

\usepackage{microtype}
\usepackage{graphicx}
\usepackage{subcaption}
\usepackage{booktabs} 

\usepackage{hyperref}

\usepackage[accepted]{icml2022}

\usepackage{amsmath}
\usepackage{amssymb}
\usepackage{mathtools}
\usepackage{amsthm}

\usepackage[capitalize,noabbrev]{cleveref}

\theoremstyle{plain}

\theoremstyle{definition}

\theoremstyle{remark}

\usepackage[textsize=tiny]{todonotes}

\definecolor{link-color}{RGB}{25,33,205}
\definecolor{dark-blue}{RGB}{4,13,112}
\usepackage{url}
\usepackage{wrapfig}
\usepackage{graphicx}
\usepackage{subcaption}
\usepackage{array}
\usepackage{bm}

\usepackage{soul} 
\usepackage{xcolor} 
\usepackage{multicol} 
\usepackage{xspace} 
\usepackage{multirow}
\usepackage{paralist}
\usepackage{tikz}

\DeclareMathOperator*{\argmin}{\arg\!\min}

\newcommand{\state}{X}
\newcommand{\pred}{\hat{\state}}
\newcommand{\context}{\state_{\leq t}}
\newcommand{\update}{Y}
\newcommand{\updatehat}{\hat{\update}}
\newcommand{\simulator}{s}
\newcommand{\explicitsim}{f_\text{D}}
\newcommand{\iterativesim}{f_\text{DI}}
\newcommand{\constr}{f_\text{C}}
\newcommand{\staticstate}{Z}

\newcommand{\predictor}{\textsc{Predictor}}
\newcommand{\updater}{\textsc{Updater}}

\newcommand{\forwardbaselinename}{Forward GNN\xspace}
\newcommand{\constraintmodel}{C-GNS\xspace}
\newcommand{\iterativemodel}{Iterative GNN\xspace}

\newcommand{\dataset}[1]{\textsc{#1}}

\newcommand{\myhref}[3][blue]{\href{#2}{\color{#1}{#3}}}
\newcommand{\videopagewithlink}{\myhref{https://sites.google.com/view/constraint-based-simulator}{sites.google.com/view/constraint-based-simulator}}
\newcommand{\videos}{\myhref{https://sites.google.com/view/constraint-based-simulator}{Videos}}

\icmltitlerunning{Constraint-based graph network simulator}

\begin{document}

\twocolumn[
\icmltitle{Constraint-based graph network simulator}



\icmlsetsymbol{equal}{*}

\begin{icmlauthorlist}
\icmlauthor{Yulia Rubanova}{equal,deepmind}
\icmlauthor{Alvaro Sanchez-Gonzalez}{equal,deepmind}
\icmlauthor{Tobias Pfaff}{deepmind}
\icmlauthor{Peter Battaglia}{deepmind}
\end{icmlauthorlist}

\icmlaffiliation{deepmind}{DeepMind, London, UK}

\icmlcorrespondingauthor{Yulia Rubanova}{rubanova@deepmind.com}
\icmlcorrespondingauthor{Alvaro Sanchez-Gonzalez}{alvarosg@deepmind.com}

\icmlkeywords{Machine Learning, ICML}

\vskip 0.3in
]



\printAffiliationsAndNotice{\icmlEqualContribution} 

\begin{abstract}
In the area of physical simulations, nearly all neural-network-based methods directly predict future states from the input states. However, many traditional simulation engines instead model the constraints of the system and select the state which satisfies them. 
Here we present a framework for constraint-based learned simulation, where a scalar constraint function is implemented as a graph neural network, and future predictions are computed by solving the optimization problem defined by the learned constraint. Our model achieves comparable or better accuracy to top learned simulators on a variety of challenging physical domains, and offers several unique advantages. We can improve the simulation accuracy on a larger system by applying more solver iterations at test time. We also can incorporate novel hand-designed constraints at test time and simulate new dynamics which were not present in the training data. Our constraint-based framework shows how key techniques from traditional simulation and numerical methods can be leveraged as inductive biases in machine learning simulators.
\end{abstract}

\section{Introduction}
Consider a bowling ball colliding with a bowling pin. You might explain this event through a pair of forces being generated: one causes the pin to move, the other one causes the ball to careen away. 
This approach is analogous to physical simulators that apply an explicit forward model to calculate a future state directly from the current one, i.e. by numerically integrating equations of motion. An alternative, but equally valid, way to explain the collision is in terms of constraint satisfaction: the ball and pin cannot occupy the same location at the same time, and their combined energies and momenta must be conserved. The post-collision trajectories are the only way the future can unfold without violating these constraints. This approach is analogous to the  constraint-based simulators that generate a prediction by searching for a future state that respects all the constraints.

Both families of simulators---those based on explicit, forward functions versus those which define the dynamics implicitly, via constraints---are widely used in physics, engineering, and graphics. In principle they can model the same types of dynamics. In practice these simulators strike different trade-offs that determine which one is preferred in different domains. Explicit methods are popular for large systems with (mostly) independent local effects where space and time derivatives are relatively smooth.
By contrast, implicit approaches are often preferred for systems with strong interactions, such rigid and stiff dynamics, and more accurate solutions can often be found by using more solver iterations or more sophisticated solvers. In machine learning, so far almost all methods for learned simulation have focused on explicit forward models \cite{sanchezgonzalez20a,pfaff2021learning}, with few exceptions~\citep{neural_projections}.

\begin{figure}[t]
\centering
\includegraphics[width=0.95\linewidth]{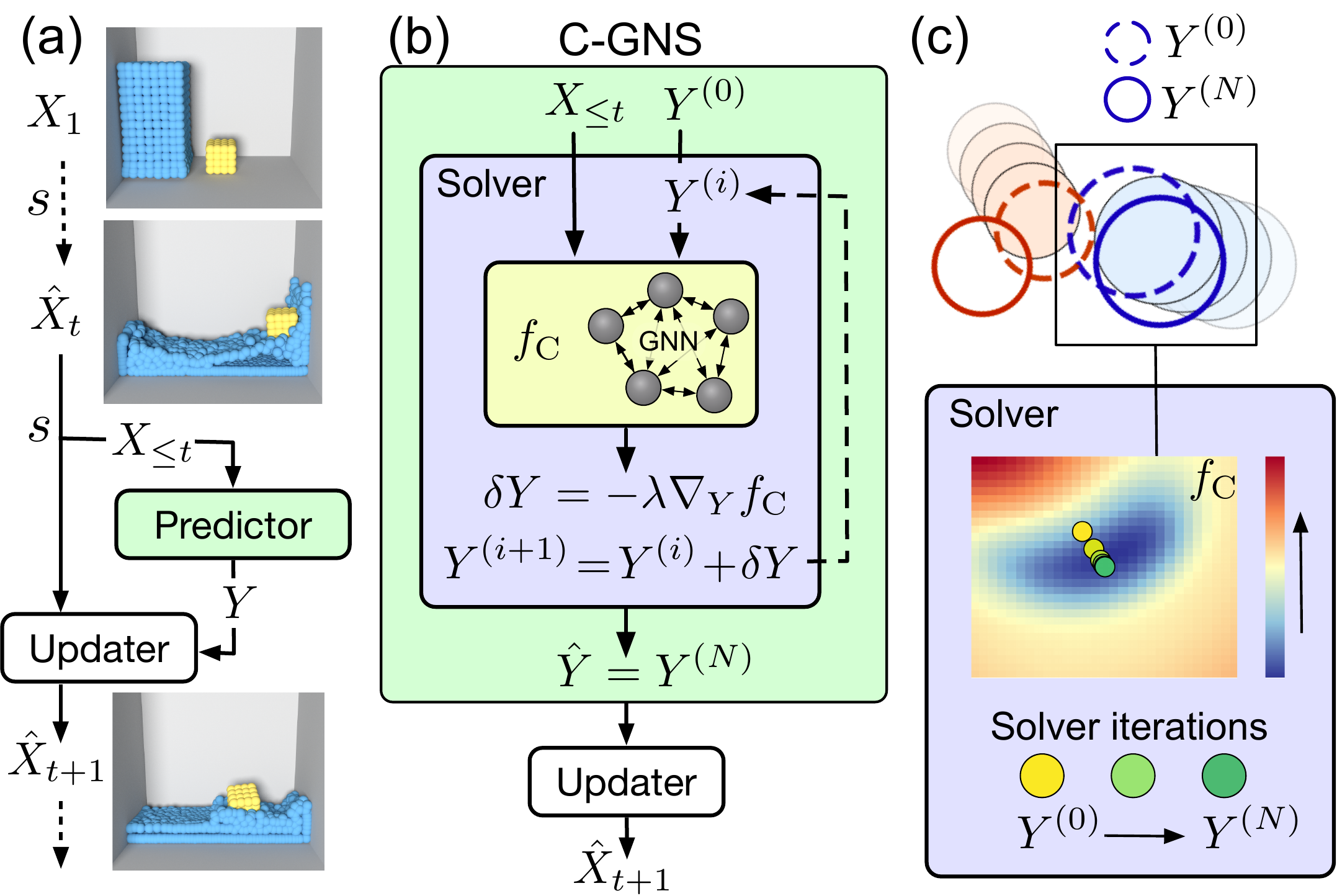}
\caption{\textbf{(a) Learned simulator schematics.} A simulator $s$ maps $\state_{\leq t}$ to a future state $\hat{X}_{t+1}$. The \predictor{} takes $\state_{\leq t}$ and returns $\hat{Y}$ which represents information about the system's temporal evolution. An \updater{} uses $\hat{Y}$ to update $\state_t$ to $\hat{X}_{t+1}$. 
\textbf{(b)~Constraint-based Graph Network simulator~(\constraintmodel)}. The \predictor{} iteratively solves for a $\hat{Y}$ to satisfy a constraint function $\constr$ using $\nabla_\update \constr$.
\textbf{(c)~Constraint optimization on two colliding balls}. The~heatmap color shows the value of learned constraint $\constr$ as we vary the update $Y$ for the blue ball. The colored points on the heatmap show the iterations of the solver as it minimizes the constraint $\constr$ (from $Y^{(0)}$ to $Y^{(N)}$), indicating that the blue ball should bounce downwards to resolve the collision. The learned $\constr$ has a "funnel" shape around the correct next state of the ball.}
\label{fig:schematic}
\end{figure}

Here we present a framework for learning to simulate complex dynamics via constraint satisfaction. Our ``Constraint-based Graph Network Simulator''~(\constraintmodel) defines a single learned constraint function that indicates whether a future state is consistent with the current and previous states. Conditioning on the previous states allows our method to capture the time dynamics within its learned constraint function. We implement the constraint function as a Graph Neural Network (GNN,~\citet{bronstein2017geometric,battaglia2018relational}).
To predict the future state, we use a gradient-based solver that iteratively refines a proposed state to minimize the learned constraint. We train the model end-to-end by backpropagating the loss gradients through the solver.
Crucially, our model is trained directly on observed trajectory data and does not require knowledge of the true underlying constraints which govern the system dynamics. The learned constraint function only needs to yield the same solution as the true constraints, while their objective landscapes may differ (e.g., the learned constraint function may be convex, as in Figure \ref{fig:schematic}(c), while the true constraint may not be).

We tested \constraintmodel on several physical simulation domains: ropes, bouncing balls and irregular rigids, and splashing fluids.
We found that  \constraintmodel produced more accurate rollouts than the state-of-the-art Graph Net Simulator~\cite{sanchezgonzalez20a} with a comparable number of parameters, and than Neural Projections \cite{neural_projections}.

We demonstrate several unique features of our model. First, the constraint function is decoupled from the procedure for satisfying it. The user can choose different solvers or invest different amounts of computation to improve the solution. We show that \constraintmodel can use additional solver iterations at test time to improve its predictive accuracy, striking desired speed-accuracy trade-offs. Second, our model allows to incorporate new, hand-designed constraints at test time and satisfy them jointly alongside its learned constraints. These properties have not been reported previously for explicit forward models or Neural Projections by \citet{neural_projections}.

\section{Background and related work}
\label{sec:relatedwork}

Constraint solvers are central to many physics simulators. Most rigid-body and game engines use constraints to model joints, collision and contact \citep{baraff1994fast}.
Position-based \citep{muller2007position} and projective dynamics \citep{bouaziz2014projective} are popular methods that express the time dynamics purely as constraints and can simulate a wide range of physical systems such as rigids, soft-bodies, fluids and cloth \citep{macklin2014unified, thomaszewski2009continuum}.

There is a rapid growth of machine learning methods for accelerating scientific simulation of complex systems, such as turbulence \citep{stachenfeld2021learned, kochkov2021machine} and aerodynamics \citep{thuerey2020deep,zhang2018airfoil}. Particularly, a learned simulator based on graph neural networks is a popular approach for modelling a wide range of systems, from articulated dynamics \citep{sanchez2018graph} to particle-based physics \citep{mrowca2018flexible,li2019learning,sanchezgonzalez20a} and mesh-based continuum systems \citep{pfaff2021learning,belbuteperes20a}. Combining learning algorithms with principles from physics and numerical methods can improve sample complexity, computational efficiency, and generalization~\citep{wu2018physics,karniadakis2021physics,chen2018neural,latent_ode}. Imposing Hamiltonian~\citep{greydanus2019hamiltonian,sanchez2019hamiltonian,chen2019symplectic} and Lagrangian~\citep{lutter2019deep,cranmer2020lagrangian,finzi2020simplifying} mechanics in learned simulators offers unique speed/accuracy tradeoffs and can preserve symmetries more effectively.

Outside of the scope of physical simulations, recent methods were proposed for learning implicit functions (see ``Deep Implicit Layers'' tutorial by \citet{duvenaud_kolter_johnson_2020} for an excellent survey). Such models can play games~\citep{amos2017optnet,wang2019satnet}, optimize power flow~\citep{donti2021dc3}, support robotic planning \citep{loula2020learning}, and perform combinatorial optimization~\citep{bartunov2020continuous}. Deep Equilibrium Models~\citep{bai2019deep,bai2020multiscale} use implicit differentiation technique to reduce the cost of computing the gradients through the solver.

Despite the popularity of traditional constraint-based simulators, only a single work that projects positional variables on a learned constraint manifold has been reported~\citep{neural_projections}. See Section~\ref{sec:advances_over_neural_proj} for a detailed comparison between our model and \citet{neural_projections}.

\section{Model Framework}
\label{sec:framework}

\subsection{Simulation basics}
A physical trajectory, measured at discrete time intervals, is a sequence of states, $(\state_1, \dots, \state_ T)$, where $\state_t$ may contain properties of elements of the system such as the positions, instantaneous velocities, masses, etc.
A physical simulator~$\simulator$ is a function that maps current and/or previous state(s), which we term the \textit{context} $\context$\footnote{Despite that physics is Markovian, we use $\context$ as input because our framework can also apply to non-Markovian dynamic processes. Providing previous states can also be helpful when there are hidden properties of the system which are only identifiable over a sequence of observed states, for example when a state does not contain instantaneous velocities, such as in our environments.}, to a predicted future state $\pred_{t+1} = s(\context)$\footnote{We loosely use the hat notation (e.g. $\pred$) for the quantities that are predicted by the model.} (see Figure~\ref{fig:schematic}a). 
A simulated physical trajectory termed a \textit{rollout} $(\state_t, \pred_{t+1} \pred_{t+2}, \dots )$, can be generated by repeatedly applying $\simulator$ to its own predicted state, $\pred_{t+1} = \simulator(\context)$. 

Simulators are often comprised of a \predictor{} and an \updater{} mechanism. The \predictor{} maps the context $\context$ to an update value $\updatehat$ that represents information about the system's temporal evolution at the current time (e.g., new positions, velocities or accelerations). Then the \updater{} mechanism uses $\updatehat$ to update the current state to the next state: $\pred_{t+1} = \updater(\context, \updatehat)$, e.g. updating current positions and velocities represented by $\state_t$ with new velocities and accelerations represented by $\updatehat$.

\subsection{Explicit simulators}
Across science, engineering, and graphics, a popular class of simulators \citep{todorov2012mujoco, Smoothed_Particle_Hydrodynamics, Impulse_Based_Simulation, Interactive_Dynamics} are defined \textit{explicitly}: the state update $\updatehat$ is predicted directly from $\context$ using an explicit forward function $\updatehat = \explicitsim(\context)$. Among the learned simulators, the forward function $\explicitsim$ is typically implemented using a graph neural network (GNN) that allows simulators to scale well to large graphs of 1000s of nodes and support generalization to systems with different shapes and sizes~\citep{sanchezgonzalez20a, pfaff2021learning, Battaglia2016InteractionNF}.
We call the explicit~GNN-based~model~\textbf{\forwardbaselinename}.

\subsection{Constraint-based implicit simulators}
\label{sec:methods:ourmodel}
In this paper we explore the learned simulators based on \textit{implicit} formulations of the dynamics. Instead of predicting the desired state directly, our implicit simulator uses a differentiable constraint function $c = \constr(\context, \update)$, where $c$ is a scalar that quantifies how well a proposed state update $\update$ agrees with $\context$. A future prediction is generated in two stages: 1) apply a solver (gradient descent or a zero-finding algorithm) to find a $\updatehat$ that satisfies the constraint function, and 2) use the value $\updatehat$ in the \updater{} to update $\state_t$ to $\pred_{t+1}$. Our constraint function $\constr$ is defined as a trainable neural network with a non-negative scalar output. It represents an approximation for all the physical constraints in the system, including the time dynamics. 

As illustrated in Figure~\ref{fig:schematic}(b), we formulate our constraint-solving procedure via an iterative method that starts with an initial proposal $\update^{(0)}$. On the $i$-th iteration, the solver uses the gradient of $\constr$ w.r.t. $\update$ to compute a change to the proposal, $\delta\update = -\lambda \left. \nabla_\update \constr(\context, \update) \right\rvert_{\update=\update^{(i)}}$. Then, $\delta\update$ is used to revise the proposal: $\update^{(i+1)}=\update^{(i)} + \delta\update$. This process repeats for $N$ steps, and the final proposal value is considered to be the \predictor{}'s output $\updatehat=\update^{(N)}$. 

We define the solution as the minimum of the constraint function $\updatehat = \argmin_\update \constr(\context, \update)$. We use gradient descent with the fixed step size $\lambda$ to find the solution $\updatehat$. We refer to our Constraint-based Graph Network Simulator with gradient descent solver as \textbf{\constraintmodel-GD}.

This general formulation of constraint-based learned simulation can be trained by backpropagating loss gradients through the solver iterations\footnote{Implicit differentiation at the solution point, mentioned in Section~\ref{sec:relatedwork}, is applicable as well but we did not explore this direction.}.
The computational budget of the forward pass can be varied via the number of solver iterations $N$, as we further explore in Section~\ref{sec:generalization}.

\subsection{Explicit iterative simulators}
\label{sec:iterative}
As a hybrid between forward and constraint-based simulators, we also introduce \textbf{\iterativemodel} model. Similarly to \constraintmodel-GD, this model iteratively refines the proposed state update, but at each iteration $\delta\update$ is predicted directly by a graph neural network $\delta\update = \iterativesim(\context, \updatehat)$, instead of being computed through a gradient. We use this hybrid model to separately study the effect of pure iterations versus iterative constraint-based optimization (Section~\ref{sec:generalization} and Figure~\ref{fig:suppl_barplots}).

\begin{figure}
\centering
\includegraphics[width=0.92\linewidth]{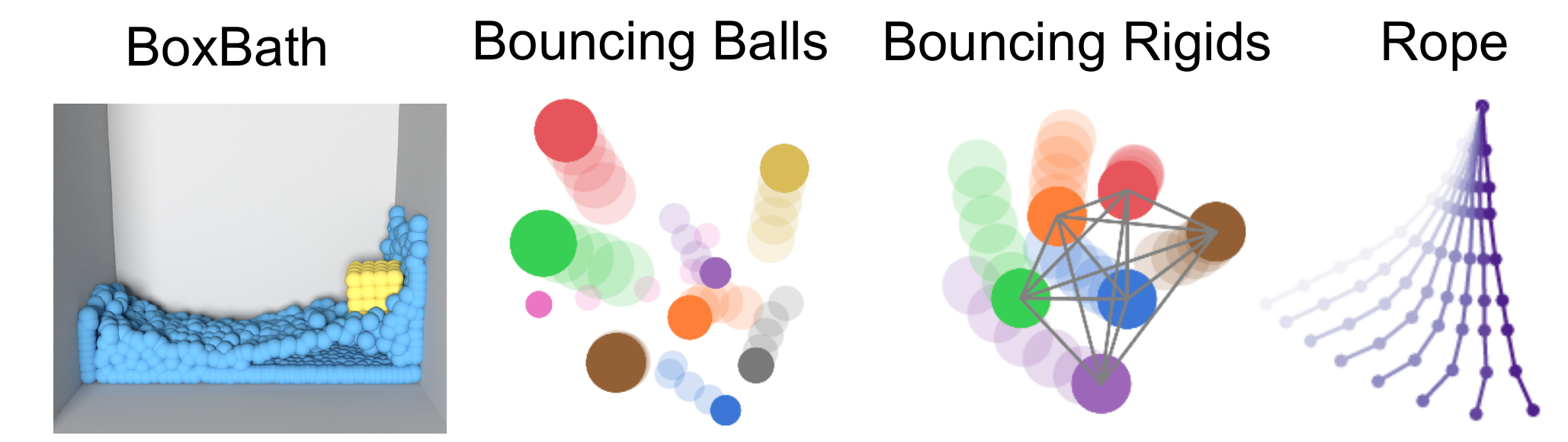}
\caption{\textbf{Renderings of the physical environments.}\\
Videos of the model rollouts are available at:\\\videopagewithlink{}.}
\label{fig:rollout_examples_main}
\end{figure}

\section{Experiments}
\label{sec:experiments}

\subsection{Experimental task domains}

We tested our approach on a variety of physical environments, shown in Figure~\ref{fig:rollout_examples_main}. We generated the data for our \dataset{Rope}, \dataset{Bouncing Balls} and \dataset{Bouncing Rigids} datasets using the MuJoCo physics simulator \cite{todorov2012mujoco}. We also tested our model on \dataset{BoxBath} dataset with 1024 particles from \cite{li2019learning} to explore the scaling capabilities of the model. These environments involve a diverse set of physical constraints: `hard' constraints on preserving the rigid shapes and resolving collisions, and `soft' constraints on fluid movement, handling gravity and preserving the momentum. The simulations consist of 150 time steps for \dataset{BoxBath} and 160 time steps for other datasets.

\textbf{Representing the physical system}
Our experimental domains consist of interacting point-like elements: sized objects, fluid particles or mesh vertices. The datasets contain the positions of each element: $ P_{t} = (p^j_t)^{j=1\dots J}$, where $J$ is the number of elements, and $p^j_t$ is the $j$-th element's position at time $t$. Note that our datasets do not contain the instantaneous velocities. Instead, the velocity information can be estimated by changes in the position, as described below. Additionally, we represent the static properties of the physical elements (masses, material types, etc.) as $\staticstate$ to keep it distinct from the dynamic state information.

\subsection{Implementation of the \constraintmodel model}
\label{sec:implementation_and_notation}

In our implementation, the state $\state_t$ consists of the positions $ P_t$ and the static information~$Z$. The input context is a sequence of the most recent positions and the static properties: $\context~:=~(\staticstate, P_{t-3}, P_{t-2}, P_{t-1}, P_{t})$.

To represent the dynamics, we set the update $Y$ to be the change in position over time $V_{t+1} = (v_{t+1}^j)^{j=1\dots J} \equiv (p^j_{t+1} - p^j_{t})^{j=1\dots J}$, which we informally call ``velocity'', estimated as a backward difference. The update mechanism $\pred_{t+1} = \updater(\context, \updatehat)$ simply becomes  $\hat{P}_{t+1} = P_{t} + \hat{V}_{t+1}$, where $\hat{V}_{t+1}$ is the output of a \predictor{}. For \dataset{BoxBath}, we set the update $\update$ to the acceleration, for the sake of consistency with \citet{sanchezgonzalez20a} (see details in Supplementary Section \ref{sec:suppl_model_details}).

\textbf{GNN-based constraint function} We represent the context~$\context$ and a proposed update $\update^{(i)}$ as a graph where the nodes correspond to different elements, such as objects or particles, and the edges correspond to the possible pairwise interactions between them. The function $\constr$ takes the input graph containing $\context$ and $\update^{(i)}$ and outputs a scalar value $c$.

To construct the graph features from the context $\context$, we enforce translation-invariance and do not explicitly provide absolute positions $P_t$ as the input to the network. The features for the node~$j$, $[z^j, v^j_{t-2}, v^j_{t-1}, v^j_{t}]$, include a sequence of three most recent position changes (i.e. velocities)  $v^j_{t} = p^j_{t} - p^j_{t-1}$ and static properties $z^j$. To construct the edge feature from nodes $j$ to $k$, we provide the relative displacement vector between the nodes' positions, $e^{jk}_{t} = p^k_{t} - p^j_{t}$. Finally, to represent $\update^{(i)}$, we concatenate the proposed update for node $j$ from the $i$-th solver iteration~$y^{j,(i)}$ (velocity or acceleration) to the node features.

We implement the function $\constr$ as a graph network (GNN) similar to \cite{sanchezgonzalez20a}. We encode nodes and edges of the input graph using MLPs. Then, we process the graph with a GNN consisting of a  sequence of message-passing (MP) layers without global updates. Next, we compute the scalar values $c^j$ for each node by running an MLP decoder on the node outputs of the GNN. We square the values $c^j$ to make them non-negative. Finally, we obtain a single scalar constraint~$c$ for the entire graph by averaging the per-node values $c = \constr(\context, \updatehat) = \frac{1}{J} \sum_{j=1}^{J} (c^j)^2$.

 \paragraph{Optimizing the constraint} We initialize $\update^{(0)}$ to the most recent velocity $V_t$, as we expect it to be a good prior for the future velocity. In \dataset{Boxbath}, where $\update$ represents the acceleration, we initialize $\update^{(0)}$ to a zero vector. We use auto-differentiation in JAX to compute the constraint gradient $\nabla_\update \constr$. For the gradient descent solver, we use a fixed step size $\lambda= 0.001$. We used $N=5$ iterations during training.

\subsection{Training and evaluation}
\label{sec:training_and_loss}
 
We compute the $L_2$ loss between the predicted update $\updatehat$ from the last solver iteration and the corresponding ground-truth update, averaged over all nodes. Note that it is straight-forward to compute the ground-truth update from the dataset. For example, if $\update$ represents the velocity, the update is simply the difference between the future and current positions.
 
To further incentivize to convergence to the ground-truth, we experimented with an additional loss between each intermediate update $\update^{(i)}$ and the ground-truth with exponential decay weights (details in Section~\ref{sec:suppl_model_implementation}). We use the additional loss only in the generalization experiments in Section~\ref{sec:generalization}, labeled with $\alpha=0.25$. For other experiments, the additional loss had little effect on the MSE error (Figure~\ref{fig:model_ablations_yang}).

We train the model on one-step prediction task using standard backpropagation with the Adam optimizer. At test time, we evaluate 1-step and rollout errors between predicted and ground truth trajectories. The rollout is computed by iteratively applying the model on the previous predictions, starting from the initial time step sequence.

\begin{figure}[t]
\centering
\includegraphics[width=\linewidth]{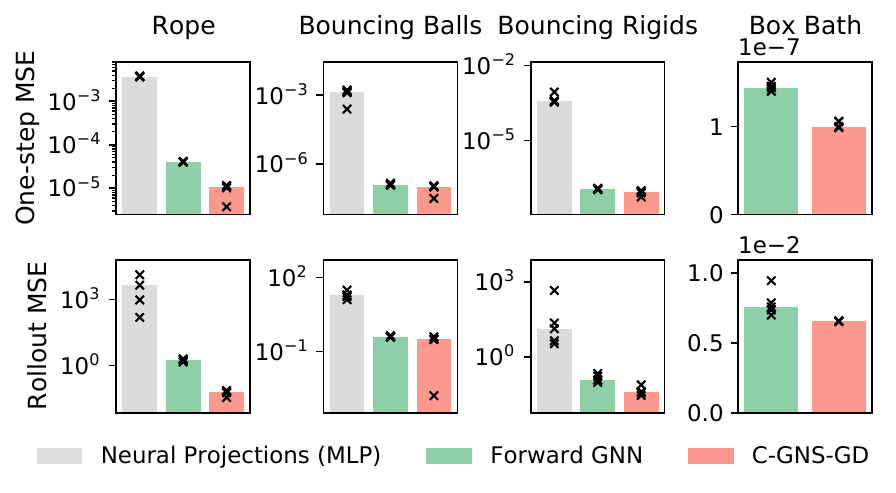}
\caption{\textbf{Comparison to the existing baselines.} Top row: 1-step test MSE on node positions. Bottom row: full-rollout test MSE (160-step). The bar height represents the median MSEs over random seeds. The black crosses show the MSE metric for each random seed.
We found Neural Projections could not effectively scale to \dataset{BoxBath} (see Section \ref{sec:advances_over_neural_proj}). All the plots except \dataset{BoxBath} are on log scale.
}
\label{fig:model_comparison}
\end{figure}

\subsection{Neural Projections and related ablations}
\label{sec:advances_over_neural_proj}

The only related work involving learned constraint-based simulation that we are aware of is \textbf{Neural Projections} (\textbf{NP}, \citet{neural_projections}). While it inspired this work, Neural Projections has key differences from our approach, and is fundamentally limited in ways that make it insufficient as a~general-purpose learned simulator.

Neural Projections operates directly on the absolute positions of the particles.
First, the model uses an Euler step to propose a future position of the particles based on the previous position, estimated velocity and known external forces. Then the model refines the proposal by iteratively projecting it onto a learned constraint manifold, implemented as a multilayer perceptron~(MLP).
In our framework, this would be equivalent to (1) setting the optimized update $\update$ to be the future positions of the system $\update := P_{t+1}$ (the \updater{} becomes the identity function), (2) making the constraint function depend on the update only: $\constr(\update)$ and (3) initializing~$\update^{(0)}$ to the output of the Euler step.

\begin{figure}[t]
\centering
\includegraphics[width=0.96\linewidth]{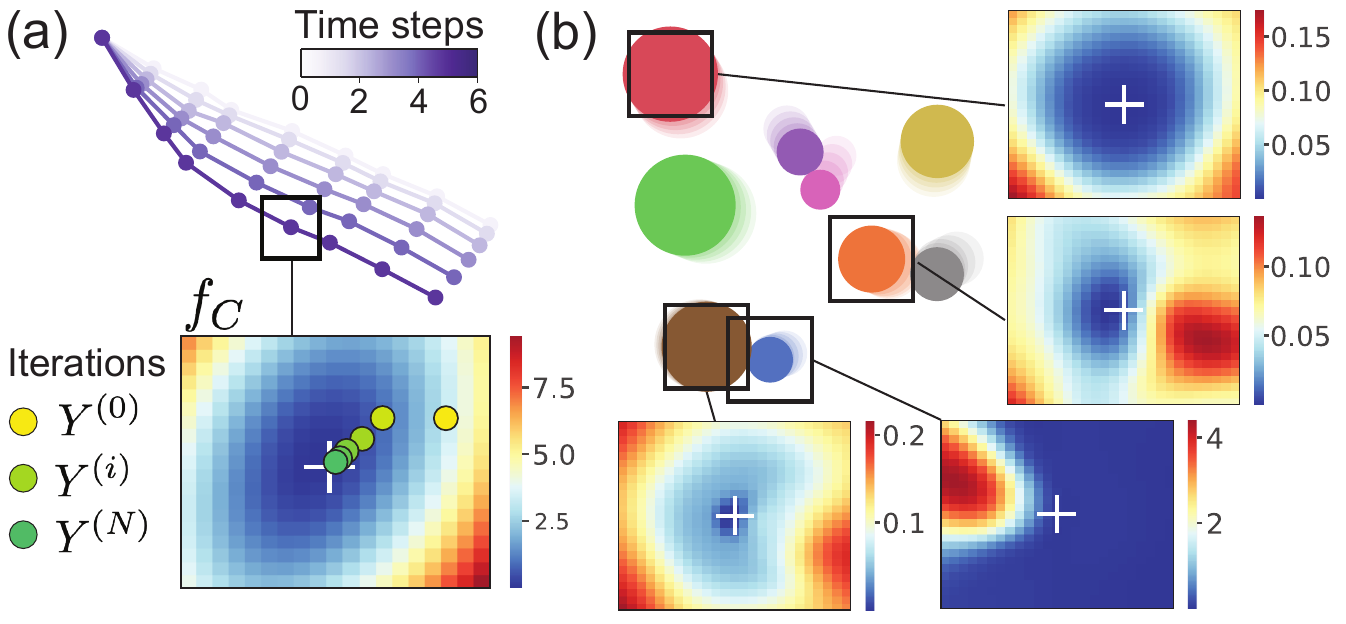}
\caption{\textbf{Visualization of the learned constraint}.
The~heatmaps show the values of the learned constraint $\constr$ as a function of the update $\update$ for one of the nodes, keeping other nodes fixed.
\\ \textbf{(a) An example from Rope simulation.} The learned constraint function has a "funnel" shape, where the minimum coincides with the only valid next state of the simulation (ground-truth, shown as the white cross). The colored points represent the iterations of the gradient descent solver as it minimizes the learned constraint: from the initial $\update^{(0)}$ (yellow) to final $\update^{(5)}$ (green). 
\\\textbf{(b) An example from Bouncing Balls.}  The red ball is far from other balls, and its constraint $\constr$ represents a smooth funnel centered around the ground-truth (white cross). For orange and blue balls, the constraint $\constr$ has high values in the areas occupied by another ball, indicating that the current ball cannot overlap with it.}
\label{fig:rope_viz}
\end{figure}

Crucially, the Neural Projections constraint function only measures how much the \textit{current} set of proposed positions of the particles violates the learned constraints without context about past states $\context$. Thus, the model cannot correctly resolve scenarios such as the elastic collisions, as the constraint function does not have access to the dynamics (i.e. how the proposed state relates to the previous state, illustrated in Supplementary Figure~\ref{fig:neural_proj_failure_cases}).
This weakness renders Neural Projections insufficient for general-purpose physical simulations. 
Our model, on the contrary, does not have this issue, because the constraint function is always conditioned on the past context  $\constr(\context,\update)$, which allows to model time dynamics as a part of the learned constraint. 
To study the effect of this difference, we provide an ablation to our model \textbf{\constraintmodel-GD-$\bm{\constr(\update)}$} that uses only the positional information of the future state and does not have access to the context $\context$ (details in the Supplementary Section \ref{sec:suppl_model_implementation})

Next, Neural Projections defines the constraint solution as $\constr(\update)=0$, and uses the zero-finding ``Fast Projection''~algorithm~(FP, \citet{goldenthal2007efficient}) to find a solution.
In contrast, our model defines the solution as a minimization problem, i.e., $\updatehat = \argmin_\update \constr(\context, \update)$, solved by a gradient descent. To explore these choices, we also tested an FP-based version of our model: \textbf{\constraintmodel-FP}.

Finally, Neural Projections uses an MLP as a constraint function that takes the concatenated features for all of the particles and outputs a constraint value. 
Compared to GNNs, MLP-based simulators have been shown to be sub-optimal to model particle systems \citep{Battaglia2016InteractionNF,sanchez2018graph}. Neural Projection paper \citep{neural_projections} includes a heuristic parameter-sharing scheme to allow variable number of particles, but it requires manually grouping subsets of the state. It is not clear how this heuristic would scale to large systems with dynamically changing interactions. We also created ablated versions of our model that uses an MLP-based constraint function instead of a~GNN: \textbf{C-MLP-GD} and \textbf{C-MLP-FP}.

\section{Results}

\subsection{Comparison to existing baselines}

Our experimental results show that the performance  of \constraintmodel-GD is generally better than the existing baselines on the datasets we tested.\footnote{Videos of the model rollouts are available at \videopagewithlink{}} Figure~\ref{fig:model_comparison} demonstrates that \constraintmodel-GD has the lowest 1-step and rollout MSE across all datasets, compared to Neural Projections \citep{neural_projections} and \forwardbaselinename \citep{pfaff2021learning} with a comparable number of parameters\footnote{To make \forwardbaselinename comparable to \constraintmodel-GD, we use the same number of MP layers in both models (2 MP layers for \dataset{Rope}, 1 MP for other datasets)}. See Supplementary Table~\ref{tab:suppl_tables} for numerical results. Qualitatively, we observed that for \forwardbaselinename, the box in \dataset{BoxBath} ``melts'' over time, as the forward model cannot preserve its rigid shape (see \videos).
By contrast, the comparable \constraintmodel-GD effectively maintains the rigid shape of the cube. We further explore the comparison to a larger \forwardbaselinename with up to 5x more parameters in Sections \ref{sec:generalization} and \ref{sec:iterations_and_layers}. These results suggest that constraint-based learned simulators are a competitive alternative to explicit forward simulators.

\subsection{Interpreting the learned constraints}

To better understand the learned constraint $\constr$ in the \constraintmodel-GD, we visualized how the output of $\constr$ changes as a function of $\update$. We varied the proposed update $\update$ (velocity in 2D space) for a particular node while holding the updates~$\update$ for other nodes fixed.
Figure \ref{fig:rope_viz}(a) shows the learned constraint for a node from the \dataset{Rope} dataset. The network learns a ``funnel" shape of the constraint that is easy to minimize with gradient descent. The constraint has a single minimum that is near the ground-truth point (the white cross). It is expected, as there is only one valid next state of the system. The sequence of points represents the proposed updates $\update^{(i)}$ from the solver, demonstrating that the solver reaches the ground-truth in five iterations, as expected. Note that we did not enforce the ``funnel" shape of the constraint.

Figure \ref{fig:rope_viz}(b) shows the learned constraint $\constr$ for several nodes in \dataset{Bouncing Balls}. For the red ball, which is far from other balls, the constraint has a ``funnel" shape, similarly to the \dataset{Rope} example. For the balls that have another object nearby, the constraint value is high in the area occupied by that object, indicating that overlapping with another object would result in an invalid physical state. 

\subsection{Constraint convergence to the minimum point}

A natural question is: does the gradient descent (GD) solver converge to the minimum? Supplementary Figure~\ref{fig:constraint_convergence} demonstrates that the constraint value in \constraintmodel-GD approaches a constant in five or more solver iterations. It is expected, as the model was trained in conjunction with the GD solver with five iterations. Using a loss on the intermediate updates with $\alpha$=0.25 further improves the convergence. We found that this loss is crucial for the generalization (Section~\ref{sec:generalization}), but does not affect the MSE otherwise (Figure~\ref{fig:model_ablations_yang}).

We also investigate whether other gradient-based solvers are able to optimize the constraint function learned by \constraintmodel-GD ($\alpha$=0.25) model at test time (Figure~\ref{fig:rope_generalisation_optimizer}). Other solvers, such as quasi-Newton BFGS method, find the solution with a similar constraint value and a similar MSE error to the GD solver. This finding suggests that it is sufficient to train the \constraintmodel model with GD solver with a fixed number of iterations in order to learn a well-behaved constraint function with the minimum near the ground-truth.

\begin{figure}[t]
\centering
\includegraphics[width=\linewidth]{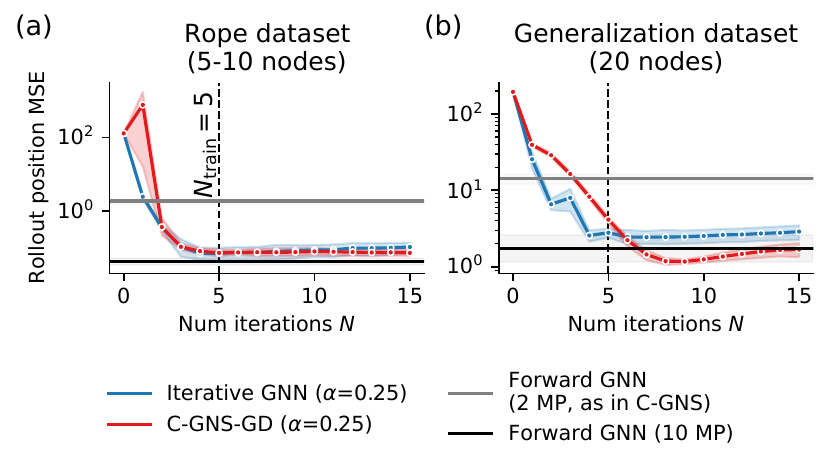}
\includegraphics[width=\linewidth]{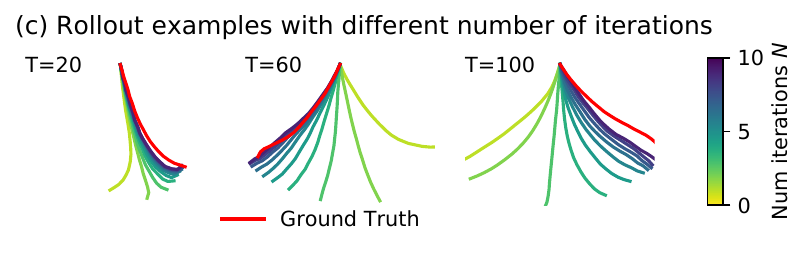}
\caption{\textbf{Generalization to more solver iterations and larger \dataset{Rope} systems at test time.} \textbf{(a)} Test rollout MSE for ropes with the same lengths as those during training (5-10 nodes) \textbf{(b)} Test rollout MSE for larger ropes (20 nodes). The x-axis indicate the number of solver iterations at test time. Vertical dashed line marks 5 iterations used at training. The y-axis represents MSE values. The horizontal black and grey lines show the performance of the Forward GNN models, which do not have an equivalent of iterations.
\textbf{(c)} Example of the generalization rollouts from \constraintmodel-GD with a different number of solver iterations used at test time at every time step. The rope examples shown at time points $T=\{20, 60, 100\}$ of the rollout. These rollouts are from a single \constraintmodel-GD model trained with 5 iterations.}
\label{fig:rope_generalisation}
\end{figure}

\subsection{Generalizing to larger systems with more iterations}
\label{sec:generalization}

A unique feature of our C-GNS-GD model is that the number of solver iterations can be increased to potentially improve the quality of the model's predictions. We explore this property on the rope simulation in two settings: on the same dataset used during the training and on a generalization to the ropes with twice as many nodes. We compare the generalization of \constraintmodel-GD model to the \iterativemodel (Section~\ref{sec:iterative}) which also iteratively refines the solution, but does not use the constraint gradients.
For this section, we use an additional loss $(\alpha=0.25)$ on intermediate updates $\update^{(i)}$ to further incentivize the convergence to the ground-truth point. See Supplementary Figure~\ref{fig:suppl_rope_generalisation} for the similar results without the additional loss.

\paragraph{\dataset{Rope} dataset}
We pre-train the \constraintmodel-GD and \iterativemodel models on the \dataset{Rope} with $N_\textrm{train}=5$ iterations. Then we investigate how the test error changes as we vary the number of solver iterations at test time in $N_\textrm{test} \in [0, 15]$.
Figure~\ref{fig:rope_generalisation}(a) shows that for \constraintmodel-GD the rollout MSE on the \dataset{Rope} remains relatively constant: error decreases by 1.2\% from iteration 5 to 15, while for \iterativemodel the error increases by 47\%.

\paragraph{Generalization to a larger rope}
We test if the models can generalize to a rope with twice as many nodes (20 nodes versus 5-10 during training) (Figure~\ref{fig:rope_generalisation}b). Crucially, for \constraintmodel-GD ($\alpha=0.25$, red), increasing the solver iterations systematically improves the rollout accuracy on the generalization task (3-fold decrease on the rollout error between iterations 5 and 9 in Figure~\ref{fig:rope_generalisation}(b), 26.2\% decrease on 1-step error in Figure~\ref{fig:suppl_rope_generalisation}(c)). We emphasize that this experiment was performed on a new dataset and with more solver iterations on each of 160 steps of the rollout -- none of these conditions were observed at training time.

Note also that this result is achieved with a shallow \constraintmodel-GD model with 2 message-passing layers that spans 1/10 of the rope length on generalization task. By contrast, the performance of the \iterativemodel (blue) with the same number of MP layers stays the same for $N_\textrm{test} > 4$. It demonstrates that \constraintmodel-GD can leverage extra computational resources at test time, because of the inductive bias that the solver should (approximately) converge to a solution.

In Figure~\ref{fig:rope_generalisation}a-b we also compared \constraintmodel-GD to the Forward GNN with the~same~number of parameters (2 MP, grey line). We find that the \constraintmodel-GD has an order of magnitude better performance, on both the \dataset{Rope} and the generalization dataset.
Next, we compared \constraintmodel-GD to a deeper Forward GNN with 10 MP (black line). Even though the state-of-the-art Forward GNN (10 MP) is slightly better on the \dataset{Rope} dataset (Figure~\ref{fig:rope_generalisation}a), \constraintmodel-GD achieves about 50\% lower error when generalizing to the larger system by leveraging additional optimization iterations (Figure~\ref{fig:rope_generalisation}b).

In this section we showed that by increasing the number of iterations $N$ at test time,  \constraintmodel-GD can achieve more accurate solutions without re-training or fine-tuning the model. To our knowledge this is the first demonstration of leveraging additional resources to improve generalization to a larger system in the domain of learned physical simulations.

\subsection{Incorporating novel constraints at test time}
\label{sec:custom_constraints}
A unique advantage of the constraint-based model is that we can incorporate additional, hand-designed constraints at test time without any fine-tuning on the model. To do so, we simply take a weighted sum of the hand-designed constraints and the learned constraint $\constr$ and run the forward evaluation to find the solution of the joint constraint.

\begin{figure}
\centering
  \includegraphics[width=\linewidth]{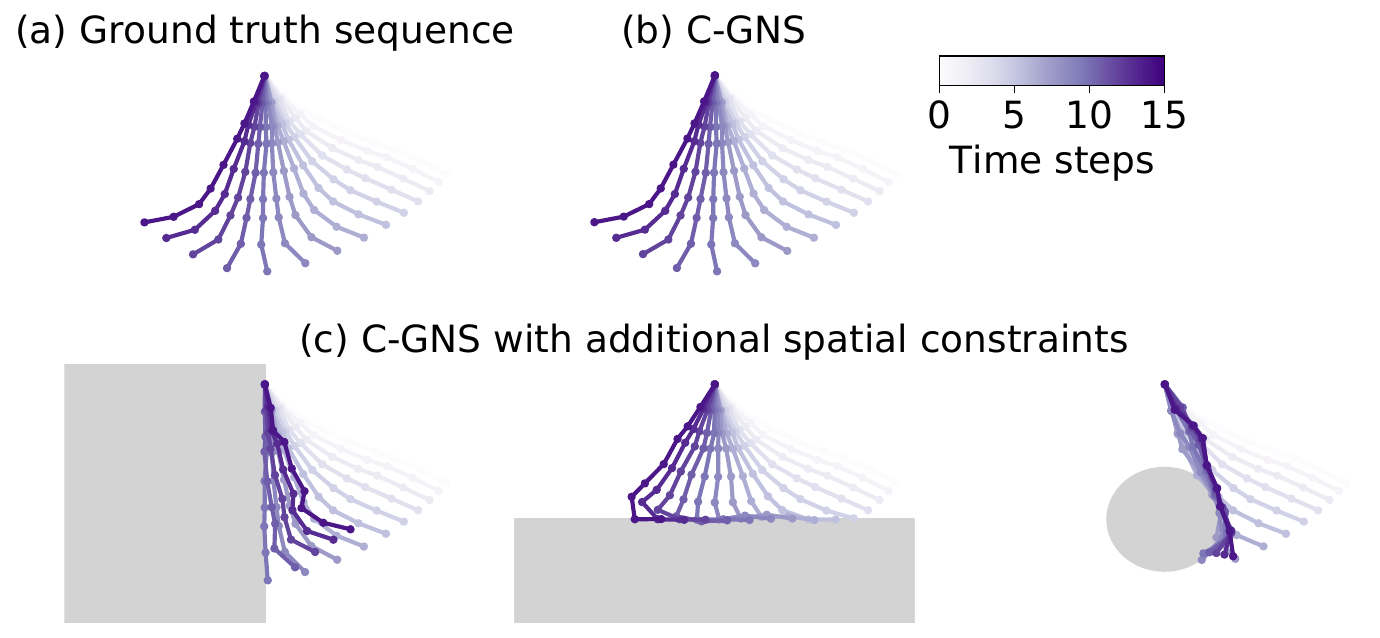}
\caption{\textbf{Adding hand-designed constraints.} \textbf{(a)} The ground truth sequence of rope simulation 14 time steps. \textbf{(b)} The rollout from \constraintmodel-GD trained on the \dataset{Rope} dataset, without added constraints. \textbf{(c)} The \constraintmodel-GD's rollout, with wall, floor, and disk-shaped obstacles, imposed at test time via hand-designed constraint functions.
}
    \label{fig:hand_designed_constr}
\end{figure}

We designed three constraint functions for the \dataset{Rope} dataset that represent the ``forbidden'' regions of the space: a vertical wall, a horizontal floor, and a disk-shaped region (Figure~\ref{fig:hand_designed_constr}). The hand-designed constraints are non-negative and increase quadratically
as the rope nodes enter the ``forbidden'' region. Figure~\ref{fig:hand_designed_constr} shows that the model resolves the collisions between the rope and the obstacle. This behavior is new: there are no examples of the rope interacting with other objects in the training data.
Note that satisfying the additional constraint may require to slightly violate the learned constraint, which is trained on the ropes moving solely under gravity. In some rollouts, the model finds the solution where the rope links change in length to avoid the obstacle. To prevent this, we add a second hand-designed constraint to preserve the lengths of the rope links (see \videos{}).

More broadly, this is a powerful example of how constraint-based models can generalize to behaviors outside of their training data, and solve both for the learned dynamics and arbitrary desired constraints.

\subsection{Examining differences from Neural Projections}

\begin{figure*}
\centering
\includegraphics[width=0.9\linewidth]{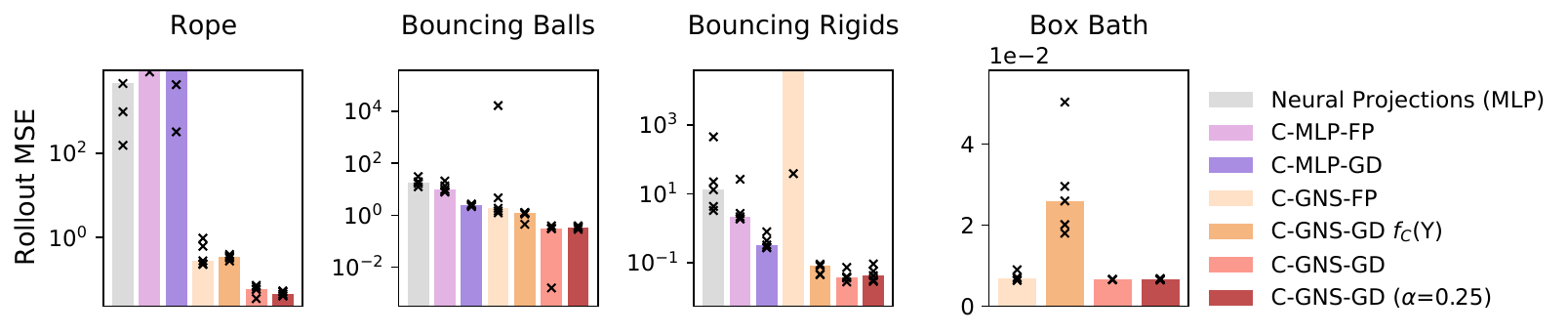}
\caption{\textbf{Ablations towards Neural Projections model}. Y axis shows full-rollout test MSE on the node positions. The bars represent the median MSE over random seeds. The black crosses show the MSE metric for each random seed. All plots except \dataset{BoxBath} are on log scale.
We crop the Y axis if~it~ exceeds the~median~MSE of \constraintmodel-GD by 5 fold. The results are not shown for MLP-based models on \dataset{BoxBath}, as these models could not effectively scale to a large system (see Section~\ref{sec:advances_over_neural_proj}). See Supplement for 1-step MSE errors.}
\label{fig:model_ablations_yang}
\end{figure*}

We demonstrate that our model's key differences from Neural Projections~\citep{neural_projections} provide substantial improvements in performance. We provide ablations of our \constraintmodel-GD for each of these differences (conditioning on past states, using GNNs, using gradient descent), as summarized in Table~\ref{tab:model_definitions}. Figure~\ref{fig:model_ablations_yang} shows that \constraintmodel-GD has several orders of magnitude lower rollout error compared to Neural Projection. Each of our ablations towards Neural Projection had higher error than \constraintmodel-GD. Parameterising the constraint with a GNN instead of an MLP yields the largest improvement on all datasets, particularly on \dataset{Rope}.

\begin{table}
    \centering
    \small
    \begin{tabular}{p{2.4cm}>{\centering\arraybackslash}p{0.7cm}>{\centering\arraybackslash}p{1.9cm}p{1.55cm}}
    \hline
    \textbf{Model variant} & \multirow{ 2}{*}{\textbf{$\constr$}} & \textbf{Solution} & \multirow{ 2}{*}{\textbf{Use $\context$?}}\\
     & & \textbf{specification} & \\
    \hline
    Neural Projections &  MLP & $\constr=0$ & $\constr(\update)$ \\
    C-MLP-FP & MLP & $\constr=0$ & $\constr(\context, \update)$ \\
    C-MLP-GD & MLP & $\argmin \constr$ & $\constr(\context, \update)$ \\
    \constraintmodel-FP & GNN  & $\constr=0$ & $\constr(\context, \update)$\\
    \constraintmodel-GD~$\constr(\update)$ & GNN & $\argmin \constr$ & $\constr(\update)$ \\
    \textbf{\constraintmodel-GD} & GNN & $\argmin \constr$ & $\constr(\context, \update)$\\
    \hline
    \end{tabular}
    \caption{\textbf{Model ablations.} The ``Model variant'' column lists the names of Neural Projections, our model, and the ablated models. The ``$\constr$'' column indicates whether the constraint function was an MLP or GNN. The ``Solution specification'' column indicates how the solution was defined, i.e., as the zero point or minimum of the constraint function. The ``Use $\context$?'' column indicates whether or not the constraint function operated over the previous states.}
    \label{tab:model_definitions}
\end{table}

We found that FP-based models were difficult to train. Notice that the FP models (\textbf{\constraintmodel-FP} and \textbf{C-MLP-FP}) suffer from instability across seeds.
We speculate that the FP algorithm makes the training challenging because the step size $\lambda$ is proportional to $\constr$. This may cause poor zero-finding early in training when the $\constr$ is not yet informative. Additionally, we find that \constraintmodel-FP algorithm becomes unstable in the areas with shallow constraint gradients, perhaps because its~$\lambda$ depends on the inverse of the gradient's norm.

We report comparisons between \constraintmodel-GD and \iterativemodel in Supplementary Figures~\ref{fig:model_ablations_yang_suppl}~and~\ref{fig:model_ablations_forward_gnn}. The \iterativemodel has higher 1-step error than \constraintmodel-GD on all our datasets and is competitive in terms of the~rollout~error.

\subsection{Further Comparison to \forwardbaselinename}
\label{sec:iterations_and_layers}

We explored how varying the number of MP layers and solver iterations $N$ at training time influenced the performance of \constraintmodel-GD compared to \forwardbaselinename in our \dataset{Rope} dataset (Supplementary Figure~\ref{fig:suppl_barplots}). Even when training with one iteration of the solver, \constraintmodel-GD outperforms the \forwardbaselinename with the same number of MP layers. The performance further improves if we train \constraintmodel-GD with more iterations (from 1 to 5), while using exactly the same number of model parameters. In comparison to deeper \forwardbaselinename (right-most facet), \constraintmodel-GD with 4 MP layers and 5 iterations has similar 1-step and full~rollout~MSE to a \forwardbaselinename with 10 MP layers, demonstrating that \constraintmodel-GD generally requires 2.5 times fewer MP layers than \forwardbaselinename to achieve comparable performance.

\section{Discussion}

We presented a general-purpose framework for constraint-based learned simulation, where a learned constraint function implicitly represents the time dynamics of the physical system and future predictions are generated via a constraint solver. We implemented our constraint function as a graph network and used gradient descent as the constraint solver. Our results showed that our \constraintmodel has competitive or better performance compared to previous learned simulators in a variety of challenging physical simulation problems. We demonstrated unique abilities of \constraintmodel to generalize to novel, hand-designed constraints and improve the simulation accuracy on larger systems at test time by increasing solver iterations. These properties have not been previously demonstrated in the space of learned physical simulations.

Implicit constraint-based models have stronger inductive biases, compared to explicit forward simulators, offering trade-offs between expressivity, adaptive computation and allowing to incorporate manual constraint terms. One inductive bias is parameter-sharing: the gradient $\nabla_\update \constr$ in \constraintmodel effectively ties the parameters across $N$ solver iterations. In principle, a deep forward simulator can be more expressive than \constraintmodel: each layer in the unshared forward model could take values of the shared parameters of \constraintmodel. In practice, \constraintmodel requires 2.5 times fewer MP layers to achieve a comparable test performance to the forward simulator with 10 MP layers (Section \ref{sec:iterations_and_layers}). Another inductive bias is that \constraintmodel searches for a solution that converges to the fixed point. This property makes it easy and natural to incorporate novel hand-designed constraints at test time, and generalize to more solver iterations and larger systems.

One key area for further improvements in constraint-based models is the runtime. Our model applies the gradient descent solver in the forward pass, requiring $2N$-times longer computation time ($N$ is the number of solver iterations) compared to the forward model with the same number of parameters. The multiplier $2$ is due to the computation of the constraint gradient via vector-Jacobian product (VJP). Similar iterative models, such as Deep Equilibrium models (DEQ, \citet{bai2019deep}), suffer from similar issues. Different techniques may help reduce the runtime and the memory cost: using more efficient, adaptive solvers or alternative ways to compute the gradient of the solution (e.g., implicit differentiation~\citep{Liao2018RevivingAI}, as used in DEQs).

One area where constraint-based simulation may be especially effective is in systems with hard constraints that require finding the equilibrium state of many local constraints and might leverage the adaptive computation. Domains with global constraints might also benefit from using constraint-based simulators, as it is easier to compute the constraint value than to directly propose the state that satisfies it.

Overall, the performance, generality and unique advantages of constraint-based learned simulation make it an important new direction of machine learning methods for complex simulation problems in science and engineering.

\bibliography{references}

\begin{thebibliography}{47}
\providecommand{\natexlab}[1]{#1}
\providecommand{\url}[1]{\texttt{#1}}
\expandafter\ifx\csname urlstyle\endcsname\relax
  \providecommand{\doi}[1]{doi: #1}\else
  \providecommand{\doi}{doi: \begingroup \urlstyle{rm}\Url}\fi

\bibitem[Amos \& Kolter(2017)Amos and Kolter]{amos2017optnet}
Amos, B. and Kolter, J.~Z.
\newblock Optnet: Differentiable optimization as a layer in neural networks.
\newblock In \emph{International Conference on Machine Learning}, pp.\
  136--145. PMLR, 2017.

\bibitem[Ba et~al.(2016)Ba, Kiros, and Hinton]{ba2016layer}
Ba, J.~L., Kiros, J.~R., and Hinton, G.~E.
\newblock Layer normalization, 2016.

\bibitem[Bai et~al.(2019)Bai, Kolter, and Koltun]{bai2019deep}
Bai, S., Kolter, J.~Z., and Koltun, V.
\newblock Deep equilibrium models.
\newblock \emph{arXiv preprint arXiv:1909.01377}, 2019.

\bibitem[Bai et~al.(2020)Bai, Koltun, and Kolter]{bai2020multiscale}
Bai, S., Koltun, V., and Kolter, J.~Z.
\newblock Multiscale deep equilibrium models.
\newblock \emph{arXiv preprint arXiv:2006.08656}, 2020.

\bibitem[Baraff(1994)]{baraff1994fast}
Baraff, D.
\newblock Fast contact force computation for nonpenetrating rigid bodies.
\newblock In \emph{Proceedings of the 21st annual conference on Computer
  graphics and interactive techniques}, pp.\  23--34, 1994.

\bibitem[Bartunov et~al.(2020)Bartunov, Nair, Battaglia, and
  Lillicrap]{bartunov2020continuous}
Bartunov, S., Nair, V., Battaglia, P., and Lillicrap, T.
\newblock Continuous latent search for combinatorial optimization.
\newblock In \emph{Learning Meets Combinatorial Algorithms at NeurIPS2020},
  2020.

\bibitem[Battaglia et~al.(2016)Battaglia, Pascanu, Lai, Rezende, and
  Kavukcuoglu]{Battaglia2016InteractionNF}
Battaglia, P., Pascanu, R., Lai, M., Rezende, D.~J., and Kavukcuoglu, K.
\newblock Interaction networks for learning about objects, relations and
  physics.
\newblock \emph{ArXiv}, abs/1612.00222, 2016.

\bibitem[Battaglia et~al.(2018)Battaglia, Hamrick, Bapst, Sanchez-Gonzalez,
  Zambaldi, Malinowski, Tacchetti, Raposo, Santoro, Faulkner,
  et~al.]{battaglia2018relational}
Battaglia, P.~W., Hamrick, J.~B., Bapst, V., Sanchez-Gonzalez, A., Zambaldi,
  V., Malinowski, M., Tacchetti, A., Raposo, D., Santoro, A., Faulkner, R.,
  et~al.
\newblock Relational inductive biases, deep learning, and graph networks.
\newblock \emph{arXiv preprint arXiv:1806.01261}, 2018.

\bibitem[Bouaziz et~al.(2014)Bouaziz, Martin, Liu, Kavan, and
  Pauly]{bouaziz2014projective}
Bouaziz, S., Martin, S., Liu, T., Kavan, L., and Pauly, M.
\newblock Projective dynamics: Fusing constraint projections for fast
  simulation.
\newblock \emph{ACM transactions on graphics (TOG)}, 33\penalty0 (4):\penalty0
  1--11, 2014.

\bibitem[Bronstein et~al.(2017)Bronstein, Bruna, LeCun, Szlam, and
  Vandergheynst]{bronstein2017geometric}
Bronstein, M.~M., Bruna, J., LeCun, Y., Szlam, A., and Vandergheynst, P.
\newblock Geometric deep learning: going beyond euclidean data.
\newblock \emph{IEEE Signal Processing Magazine}, 34\penalty0 (4):\penalty0
  18--42, 2017.

\bibitem[Chen et~al.(2018)Chen, Rubanova, Bettencourt, and
  Duvenaud]{chen2018neural}
Chen, R.~T., Rubanova, Y., Bettencourt, J., and Duvenaud, D.
\newblock Neural ordinary differential equations.
\newblock \emph{arXiv preprint arXiv:1806.07366}, 2018.

\bibitem[Chen et~al.(2019)Chen, Zhang, Arjovsky, and
  Bottou]{chen2019symplectic}
Chen, Z., Zhang, J., Arjovsky, M., and Bottou, L.
\newblock Symplectic recurrent neural networks.
\newblock \emph{arXiv preprint arXiv:1909.13334}, 2019.

\bibitem[Cranmer et~al.(2020)Cranmer, Greydanus, Hoyer, Battaglia, Spergel, and
  Ho]{cranmer2020lagrangian}
Cranmer, M., Greydanus, S., Hoyer, S., Battaglia, P., Spergel, D., and Ho, S.
\newblock Lagrangian neural networks.
\newblock \emph{arXiv preprint arXiv:2003.04630}, 2020.

\bibitem[De~Avila Belbute-Peres et~al.(2020)De~Avila Belbute-Peres, Economon,
  and Kolter]{belbuteperes20a}
De~Avila Belbute-Peres, F., Economon, T., and Kolter, Z.
\newblock Combining differentiable {PDE} solvers and graph neural networks for
  fluid flow prediction.
\newblock In III, H.~D. and Singh, A. (eds.), \emph{Proceedings of the 37th
  International Conference on Machine Learning}, volume 119 of
  \emph{Proceedings of Machine Learning Research}, pp.\  2402--2411. PMLR,
  13--18 Jul 2020.

\bibitem[Donti et~al.(2021)Donti, Rolnick, and Kolter]{donti2021dc3}
Donti, P.~L., Rolnick, D., and Kolter, J.~Z.
\newblock Dc3: A learning method for optimization with hard constraints.
\newblock \emph{arXiv preprint arXiv:2104.12225}, 2021.

\bibitem[Duvenaud et~al.(2020)Duvenaud, Kolter, and
  Johnson]{duvenaud_kolter_johnson_2020}
Duvenaud, D., Kolter, Z., and Johnson, M.
\newblock Deep implicit layers - neural odes, deep equilibrium models, and
  beyond, 2020.
\newblock URL \url{http://implicit-layers-tutorial.org/}.

\bibitem[Finzi et~al.(2020)Finzi, Wang, and Wilson]{finzi2020simplifying}
Finzi, M., Wang, K.~A., and Wilson, A.~G.
\newblock Simplifying hamiltonian and lagrangian neural networks via explicit
  constraints.
\newblock \emph{arXiv preprint arXiv:2010.13581}, 2020.

\bibitem[Goldenthal et~al.(2007)Goldenthal, Harmon, Fattal, Bercovier, and
  Grinspun]{goldenthal2007efficient}
Goldenthal, R., Harmon, D., Fattal, R., Bercovier, M., and Grinspun, E.
\newblock Efficient simulation of inextensible cloth.
\newblock In \emph{ACM SIGGRAPH 2007 papers}, pp.\  49--es. ACM New York, NY,
  USA, 2007.

\bibitem[Greydanus et~al.(2019)Greydanus, Dzamba, and
  Yosinski]{greydanus2019hamiltonian}
Greydanus, S., Dzamba, M., and Yosinski, J.
\newblock Hamiltonian neural networks.
\newblock \emph{Advances in Neural Information Processing Systems},
  32:\penalty0 15379--15389, 2019.

\bibitem[Karniadakis et~al.(2021)Karniadakis, Kevrekidis, Lu, Perdikaris, Wang,
  and Yang]{karniadakis2021physics}
Karniadakis, G.~E., Kevrekidis, I.~G., Lu, L., Perdikaris, P., Wang, S., and
  Yang, L.
\newblock Physics-informed machine learning.
\newblock \emph{Nature Reviews Physics}, 3\penalty0 (6):\penalty0 422--440,
  2021.

\bibitem[Kochkov et~al.(2021)Kochkov, Smith, Alieva, Wang, Brenner, and
  Hoyer]{kochkov2021machine}
Kochkov, D., Smith, J.~A., Alieva, A., Wang, Q., Brenner, M.~P., and Hoyer, S.
\newblock Machine learning--accelerated computational fluid dynamics.
\newblock \emph{Proceedings of the National Academy of Sciences}, 118\penalty0
  (21), 2021.

\bibitem[Li et~al.(2019)Li, Wu, Tedrake, Tenenbaum, and
  Torralba]{li2019learning}
Li, Y., Wu, J., Tedrake, R., Tenenbaum, J.~B., and Torralba, A.
\newblock Learning particle dynamics for manipulating rigid bodies, deformable
  objects, and fluids.
\newblock In \emph{ICLR}, 2019.

\bibitem[Liao et~al.(2018)Liao, Xiong, Fetaya, Zhang, Yoon, Pitkow, Urtasun,
  and Zemel]{Liao2018RevivingAI}
Liao, R., Xiong, Y., Fetaya, E., Zhang, L., Yoon, K., Pitkow, X., Urtasun, R.,
  and Zemel, R.~S.
\newblock Reviving and improving recurrent back-propagation.
\newblock In \emph{ICML}, 2018.

\bibitem[Loula et~al.(2020)Loula, Allen, Silver, and
  Tenenbaum]{loula2020learning}
Loula, J., Allen, K., Silver, T., and Tenenbaum, J.
\newblock Learning constraint-based planning models from demonstrations.
\newblock In \emph{2020 IEEE/RSJ International Conference on Intellitgent
  Robots and Systems (IROS)}, pp.\  5410--5416. IEEE, 2020.

\bibitem[Lutter et~al.(2019)Lutter, Ritter, and Peters]{lutter2019deep}
Lutter, M., Ritter, C., and Peters, J.
\newblock Deep lagrangian networks: Using physics as model prior for deep
  learning.
\newblock \emph{arXiv preprint arXiv:1907.04490}, 2019.

\bibitem[Macklin et~al.(2014{\natexlab{a}})Macklin, M{\"u}ller, Chentanez, and
  Kim]{macklin2014unified}
Macklin, M., M{\"u}ller, M., Chentanez, N., and Kim, T.-Y.
\newblock Unified particle physics for real-time applications.
\newblock \emph{ACM Transactions on Graphics (TOG)}, 33\penalty0 (4):\penalty0
  1--12, 2014{\natexlab{a}}.

\bibitem[Macklin et~al.(2014{\natexlab{b}})Macklin, M\"{u}ller, Chentanez, and
  Kim]{nvidia_flex}
Macklin, M., M\"{u}ller, M., Chentanez, N., and Kim, T.-Y.
\newblock Unified particle physics for real-time applications.
\newblock \emph{ACM Trans. Graph.}, 33\penalty0 (4), jul 2014{\natexlab{b}}.
\newblock ISSN 0730-0301.
\newblock \doi{10.1145/2601097.2601152}.
\newblock URL \url{https://doi.org/10.1145/2601097.2601152}.

\bibitem[Mirtich \& Canny(1995)Mirtich and Canny]{Impulse_Based_Simulation}
Mirtich, B. and Canny, J.
\newblock Impulse-based simulation of rigid bodies.
\newblock In \emph{Proceedings of the 1995 Symposium on Interactive 3D
  Graphics}, I3D '95, pp.\  181–ff., New York, NY, USA, 1995. Association for
  Computing Machinery.
\newblock ISBN 0897917367.
\newblock \doi{10.1145/199404.199436}.
\newblock URL \url{https://doi.org/10.1145/199404.199436}.

\bibitem[Monaghan(2005)]{Smoothed_Particle_Hydrodynamics}
Monaghan, J.
\newblock Smoothed particle hydrodynamics.
\newblock \emph{Reports on Progress in Physics}, 68:\penalty0 1703, 07 2005.
\newblock \doi{10.1088/0034-4885/68/8/R01}.

\bibitem[Mrowca et~al.(2018)Mrowca, Zhuang, Wang, Haber, Fei-Fei, Tenenbaum,
  and Yamins]{mrowca2018flexible}
Mrowca, D., Zhuang, C., Wang, E., Haber, N., Fei-Fei, L., Tenenbaum, J.~B., and
  Yamins, D.~L.
\newblock Flexible neural representation for physics prediction.
\newblock \emph{arXiv preprint arXiv:1806.08047}, 2018.

\bibitem[M{\"u}ller et~al.(2007)M{\"u}ller, Heidelberger, Hennix, and
  Ratcliff]{muller2007position}
M{\"u}ller, M., Heidelberger, B., Hennix, M., and Ratcliff, J.
\newblock Position based dynamics.
\newblock \emph{Journal of Visual Communication and Image Representation},
  18\penalty0 (2):\penalty0 109--118, 2007.

\bibitem[Nocedal \& Wright(2006)Nocedal and Wright]{NoceWrig06}
Nocedal, J. and Wright, S.~J.
\newblock \emph{Numerical Optimization}.
\newblock Springer, New York, NY, USA, 2e edition, 2006.

\bibitem[Pfaff et~al.(2021)Pfaff, Fortunato, Sanchez-Gonzalez, and
  Battaglia]{pfaff2021learning}
Pfaff, T., Fortunato, M., Sanchez-Gonzalez, A., and Battaglia, P.
\newblock Learning mesh-based simulation with graph networks.
\newblock In \emph{International Conference on Learning Representations}, 2021.
\newblock URL \url{https://openreview.net/forum?id=roNqYL0_XP}.

\bibitem[Rubanova et~al.(2019)Rubanova, Chen, and Duvenaud]{latent_ode}
Rubanova, Y., Chen, R.~T., and Duvenaud, D.
\newblock Latent odes for irregularly-sampled time series.
\newblock In \emph{Proceedings of the 33rd International Conference on Neural
  Information Processing Systems}, pp.\  5320--5330, 2019.

\bibitem[Sanchez-Gonzalez et~al.(2018)Sanchez-Gonzalez, Heess, Springenberg,
  Merel, Riedmiller, Hadsell, and Battaglia]{sanchez2018graph}
Sanchez-Gonzalez, A., Heess, N., Springenberg, J.~T., Merel, J., Riedmiller,
  M., Hadsell, R., and Battaglia, P.
\newblock Graph networks as learnable physics engines for inference and
  control.
\newblock In \emph{International Conference on Machine Learning}, pp.\
  4470--4479. PMLR, 2018.

\bibitem[Sanchez-Gonzalez et~al.(2019)Sanchez-Gonzalez, Bapst, Cranmer, and
  Battaglia]{sanchez2019hamiltonian}
Sanchez-Gonzalez, A., Bapst, V., Cranmer, K., and Battaglia, P.
\newblock Hamiltonian graph networks with ode integrators.
\newblock \emph{arXiv preprint arXiv:1909.12790}, 2019.

\bibitem[Sanchez-Gonzalez et~al.(2020)Sanchez-Gonzalez, Godwin, Pfaff, Ying,
  Leskovec, and Battaglia]{sanchezgonzalez20a}
Sanchez-Gonzalez, A., Godwin, J., Pfaff, T., Ying, R., Leskovec, J., and
  Battaglia, P.
\newblock Learning to simulate complex physics with graph networks.
\newblock In III, H.~D. and Singh, A. (eds.), \emph{Proceedings of the 37th
  International Conference on Machine Learning}, volume 119 of
  \emph{Proceedings of Machine Learning Research}, pp.\  8459--8468. PMLR,
  13--18 Jul 2020.
\newblock URL
  \url{https://proceedings.mlr.press/v119/sanchez-gonzalez20a.html}.

\bibitem[Stachenfeld et~al.(2021)Stachenfeld, Fielding, Kochkov, Cranmer,
  Pfaff, Godwin, Cui, Ho, Battaglia, and
  Sanchez-Gonzalez]{stachenfeld2021learned}
Stachenfeld, K., Fielding, D.~B., Kochkov, D., Cranmer, M., Pfaff, T., Godwin,
  J., Cui, C., Ho, S., Battaglia, P., and Sanchez-Gonzalez, A.
\newblock Learned coarse models for efficient turbulence simulation.
\newblock \emph{arXiv preprint arXiv:2112.15275}, 2021.

\bibitem[Thomaszewski et~al.(2009)Thomaszewski, Pabst, and
  Strasser]{thomaszewski2009continuum}
Thomaszewski, B., Pabst, S., and Strasser, W.
\newblock Continuum-based strain limiting.
\newblock In \emph{Computer Graphics Forum}, volume~28, pp.\  569--576. Wiley
  Online Library, 2009.

\bibitem[Thuerey et~al.(2020)Thuerey, Wei{\ss}enow, Prantl, and
  Hu]{thuerey2020deep}
Thuerey, N., Wei{\ss}enow, K., Prantl, L., and Hu, X.
\newblock Deep learning methods for reynolds-averaged navier--stokes
  simulations of airfoil flows.
\newblock \emph{AIAA Journal}, 58\penalty0 (1):\penalty0 25--36, 2020.

\bibitem[Todorov et~al.(2012)Todorov, Erez, and Tassa]{todorov2012mujoco}
Todorov, E., Erez, T., and Tassa, Y.
\newblock Mujoco: A physics engine for model-based control.
\newblock In \emph{2012 IEEE/RSJ International Conference on Intelligent Robots
  and Systems}, pp.\  5026--5033. IEEE, 2012.

\bibitem[Virtanen et~al.(2020)Virtanen, Gommers, Oliphant, Haberland, Reddy,
  Cournapeau, Burovski, Peterson, Weckesser, Bright, et~al.]{virtanen2020scipy}
Virtanen, P., Gommers, R., Oliphant, T.~E., Haberland, M., Reddy, T.,
  Cournapeau, D., Burovski, E., Peterson, P., Weckesser, W., Bright, J., et~al.
\newblock Scipy 1.0: fundamental algorithms for scientific computing in python.
\newblock \emph{Nature methods}, 17\penalty0 (3):\penalty0 261--272, 2020.

\bibitem[Wang et~al.(2019)Wang, Donti, Wilder, and Kolter]{wang2019satnet}
Wang, P.-W., Donti, P., Wilder, B., and Kolter, Z.
\newblock Satnet: Bridging deep learning and logical reasoning using a
  differentiable satisfiability solver.
\newblock In \emph{International Conference on Machine Learning}, pp.\
  6545--6554. PMLR, 2019.

\bibitem[Witkin et~al.(1990)Witkin, Gleicher, and Welch]{Interactive_Dynamics}
Witkin, A., Gleicher, M., and Welch, W.
\newblock Interactive dynamics.
\newblock \emph{SIGGRAPH Comput. Graph.}, 24\penalty0 (2):\penalty0 11–21,
  feb 1990.
\newblock ISSN 0097-8930.
\newblock \doi{10.1145/91394.91400}.
\newblock URL \url{https://doi.org/10.1145/91394.91400}.

\bibitem[Wu et~al.(2018)Wu, Xiao, and Paterson]{wu2018physics}
Wu, J.-L., Xiao, H., and Paterson, E.
\newblock Physics-informed machine learning approach for augmenting turbulence
  models: A comprehensive framework.
\newblock \emph{Physical Review Fluids}, 3\penalty0 (7):\penalty0 074602, 2018.

\bibitem[Yang et~al.(2020)Yang, He, and Zhu]{neural_projections}
Yang, S., He, X., and Zhu, B.
\newblock Learning physical constraints with neural projections.
\newblock In Larochelle, H., Ranzato, M., Hadsell, R., Balcan, M.~F., and Lin,
  H. (eds.), \emph{Advances in Neural Information Processing Systems},
  volume~33, pp.\  5178--5189. Curran Associates, Inc., 2020.
\newblock URL
  \url{https://proceedings.neurips.cc/paper/2020/file/37bc5e7fb6931a50b3464ec66179085f-Paper.pdf}.

\bibitem[Zhang et~al.(2018)Zhang, Sung, and Mavris]{zhang2018airfoil}
Zhang, Y., Sung, W.~J., and Mavris, D.~N.
\newblock Application of convolutional neural network to predict airfoil lift
  coefficient.
\newblock In \emph{2018 AIAA/ASCE/AHS/ASC Structures, Structural Dynamics, and
  Materials Conference}, pp.\  1903, 2018.

\end{thebibliography}
\bibliographystyle{icml2022}

\newpage
\appendix
\onecolumn
\counterwithin{figure}{section}
\counterwithin{table}{section}

\section*{Supplementary Material}

\section{Implementation}

\subsection{The datasets}

We generate the \dataset{Rope}, \dataset{Bouncing Balls} and \dataset{Bouncing Rigids} datasets using the MuJoCo physics simulator, with a timestep of 0.001, and recording every 30th time step for our datasets. Our MuJoCo datasets contain 8000/100/100 train/validation/test trajectories of 160 time points each. We show examples of the rollouts for each environment in Supplementary Figure~\ref{fig:rollout_examples} and \videos{}.

\paragraph{\dataset{Rope}} The rope is a mass-spring system, where the masses are represented by nodes, and the springs are represented by edges. We randomly sample the number of masses from the discrete interval [5, 10], and the rest length of the springs from the interval [0.6, 1.1]. The springs have effectively infinite stiffness, and thus maintain their rest lengths during the simulation. The rope is fixed in space at one end, and the rest moves under the force of gravity in 2D space. 

\paragraph{\dataset{Bouncing Balls}} The bouncing balls are a 2D particle system confined to a square box, where interactions between the balls, and between the balls and walls, are simulated as rigid collisions. The number of balls is randomly sampled from the discrete interval [5, 10], and the radii of each ball from the interval [0.11, 0.3]. The size of the box is fixed to 5x5 in MuJoCo coordinates.

\paragraph{\dataset{Bouncing Rigids}} The bouncing rigids are similar to \dataset{Bouncing Balls}, except all the balls are connected to each other with rigid bars. We randomly sample the number of balls from the discrete interval [3, 6].

\paragraph{\dataset{BoxBath}} This dataset is from \cite{li2019learning}, and simulates 3D fluid particle dynamics within a box, with a rigid cube comprised of particles floating on the surface of the fluid, using the NVIDIA FleX physics engine \cite{nvidia_flex}. Each simulation contains 960 fluid particles and 64 particles representing the cube. The dataset contains 2700/10/100 training/validation/test trajectories with 150 time steps each.

\subsection{Constructing the input graph}
\label{sec:input_graph}

We construct the input graph such that the representation is translation invariant, motivated by the idea that the laws of physics do not change based on position in space. To do so, we never provide absolute positions of the nodes in the input to the GNN. Instead, we use velocities (position differences across time) as node features and pairwise position differences between the nodes as edge features, as described below and in the main text. In preliminary work, we found that providing absolute positions to the models causes poorer generalization, especially in larger environments, such as a longer rope.

In \dataset{Bouncing Rigids} and \dataset{Bouncing Balls} we use a fully-connected graph. In \dataset{Rope} we add edges between nodes that are adjacent within the rope. In \dataset{BoxBath} we add edges between particles that are within a radius of 0.08 from within each other, and then recompute these edges at every step of a rollout according to the updated positions (as in \citet{sanchezgonzalez20a}).

\subsubsection{Node features}

To construct each input node feature, we use the concatenation of the three most recent velocities (position differences) of the node, concatenated with the static parameters (context $\context$): $[z^j, v^j_{t-2}, v^j_{t-1}, v^j_{t}]$. See Figure~\ref{fig:ablations}(a-b) for experiments with different number of time points. We use five most recent velocities for \dataset{BoxBath} to match the paradigm in \cite{sanchezgonzalez20a}. For the constraint-based models, e.g. \constraintmodel-GD, we also concatenate the optimized update $\update$ to the node features, as $\update$ represents velocity or acceleration for each node.

We provide an additional one-hot node feature indicating the node type (e.g. rigid, fluid, fixed). For \dataset{Bouncing Balls} and \dataset{Bouncing Rigids}, we provide the radius of the object as an additional node feature. Note, because the edges' relative positional displacement and distance features are computed between centers of the nodes, the model must factor in the object size feature to determine whether a collision is happening.

\paragraph{Handling walls} To handle the walls, we include the Euclidean distance between the center of the node to each of wall as additional node features, treating the wall as a plain, similarly to \cite{sanchezgonzalez20a}.

We clip the distance to the wall at a fixed maximum value so that this feature cannot be exploited by the network to infer the absolute position within the box. For \dataset{Bouncing Balls} and \dataset{Bouncing Rigids} we clip the distance at 2.0, and for \dataset{BoxBath} at 0.08. For the constraint-based and iterative models, we update the distances to the walls after every step of constraint optimization or iteration, respectively.

\subsubsection{Edge features}

To construct each input edge feature, we use the concatenated displacement vectors $e^{jk}_{t} = p^k_{t} - p^j_{t}$ between the most recent positions at time point $t$ for the nodes $j$ and $k$ connected by the edge. Through the ablation studies, we found that it is sufficient to provide the displacements between the nodes only for the most recent time point $t$.

For \dataset{BoxBath}, we also provide the vector norm of the relative distances (not just the vector itself) as an additional edge feature to match \cite{sanchezgonzalez20a}. 

Note that we do not provide the ``rest shapes'' (the ground-truth distances of the nodes) for the rigid structures or the rope. When generating a rollout, the model only observes the pairwise displacements/distances between nodes predicted in the previous steps. This makes the rollout prediction more challenging, as the rigid shape might gradually drift from true ``rest shape'' during the rollout, and there is no way to recover the original shape.

\subsubsection{Further details}
\label{sec:suppl_model_details}

\paragraph{Parameterizing the update $\update$}

For \dataset{Rope}, \dataset{Bouncing Balls}, \dataset{Bouncing Rigids} we use the velocity (defined as difference between the positions at adjacent time points) as the update $\update$. At the first iterations, $\update^{(0)}$ is initialized to the previous velocity $V_t$. For \dataset{BoxBath}, we use normalized acceleration  of the particle as the update $\update$ to better match the approach in \cite{sanchezgonzalez20a, pfaff2021learning}. The acceleration is estimated as a backward difference: $A_{t+1} = V_{t+1} - V_t = P_{t+1} - 2P_t + P_{t-1}$. In this case, the $\updater{}$ takes the form $\hat{P}_{t+1} = P_{t} + V_t + \hat{A}_{t+1} = 2P_{t} - P_{t-1} +  \hat{A}_{t+1}$, where $\hat{A}_{t+1}$ is produced by the \predictor{}. The update $\update^{(0)}$ is initialized to a zero vector on the first iteration. In both cases, the proposed $\update$ is concatenated as an extra node feature at each optimization iteration.

\paragraph{Normalization} For \dataset{BoxBath} we found it was important to normalize inputs and targets to zero-mean unit-variance (as in \citet{sanchezgonzalez20a}). In the other datasets, the scale of the features was already close to zero-mean unit-variance, except for the input/target velocities in \dataset{Bouncing Balls} and \dataset{Bouncing Rigids}, so we scaled them by a factor of 100.

\paragraph{Noise} To stabilize rollouts in \dataset{BoxBath}, we added noise to the input sequences in the same manner, and with the same magnitude, as in \cite{li2019learning,sanchezgonzalez20a}.

\paragraph{Fixed particles} Some of the datasets contain fixed nodes that do not change the position, such as the ``pinned'' node in the \dataset{Rope}. As our GNN models are translation invariant, they do not observe absolute positions of the nodes and cannot correct the position of the fixed nodes. Therefore, we prevent the update for the fixed nodes by using \textit{stop\_gradient} for gradient-based constraint models, similarly to \cite{neural_projections}. For non-constraint-based models, we override fixed particles positions to remain static during a rollout for all models. We also mask out fixed particles from the loss computation. Note that excluding the fixed particles from the predicted output is a standard practice \citep{sanchezgonzalez20a, pfaff2021learning}.

\subsection{Model Implementation}
\label{sec:suppl_model_implementation}

\paragraph{Computing the constraint gradients}

To compute the gradients of the constraint scalars for the batch of graphs, we use the vector-Jacobian product (VJP) function using JAX. VJP does not explicitly construct a Jacobian, and its asymptotic computational cost is the same as the forward evaluation of the constraint function.

\paragraph{Constraint function} To construct $\constr$, we first encode the nodes and edges of the graph using MLP encoders. Then, we process the graph using a GNN model from \citep{sanchezgonzalez20a, pfaff2021learning}. The GNN model has residual connections on each message-passing layer and does not use global updates. Next, we decode the node outputs of the graph network using an MLP decoder with a scalar output to compute the per-node values $\{c^j \vert j=1\dots J\}$. Finally, to obtain the scalar constraint value for the entire graph, we use the mean aggregation for the per-node values  $c = \constr(\context, \updatehat) = \frac{1}{J} \sum_{j=1}^{J} (c^j)^2$. For gradient descent solver, we take a square of per-node outputs before aggregating them. For fast projections, we simply take the sum of per-node outputs.

We use a fixed learning rate of 0.001 for gradient descent-based constraint solvers. We did not find the model to be very sensitive to this value of the learning rate. We speculate this is because the model can indirectly control the learning rate by learning an arbitrary scaling factor for the constraint function. We use five iterations of the solver for both gradient descent and fast projection solvers during the training.

\paragraph{Loss} In most experiments we used the MSE loss between the output of the last iteration $\updatehat^{(N)}$ and the corresponding ground-truth state updates on node positions. In section \ref{sec:generalization} we used the additional MSE loss between intermediate states $\updatehat^{(i)}$ and the ground-truth point, with exponentially decaying weights $ \mathcal{L}_{interm} = \alpha^{N-i}  \text{MSE}(\updatehat^{(i)}, T)$, where  $T$ is the ground-truth, $N$ is the number of solver iterations, $\alpha$ is a parameter in (0,1]. The goal of the weighted loss is to encourage the solver to reach the solution in fewer iterations. Earlier iterations have a smaller weight and are penalized less for being farther from the ground-truth. The loss on the last iteration is the same as our standard MSE loss between the last iteration and the ground-truth, as the weight on $N$-th iterations is $\alpha^{N-N}=1$.

\begin{figure}[H]
\centering
\includegraphics[width=\linewidth]{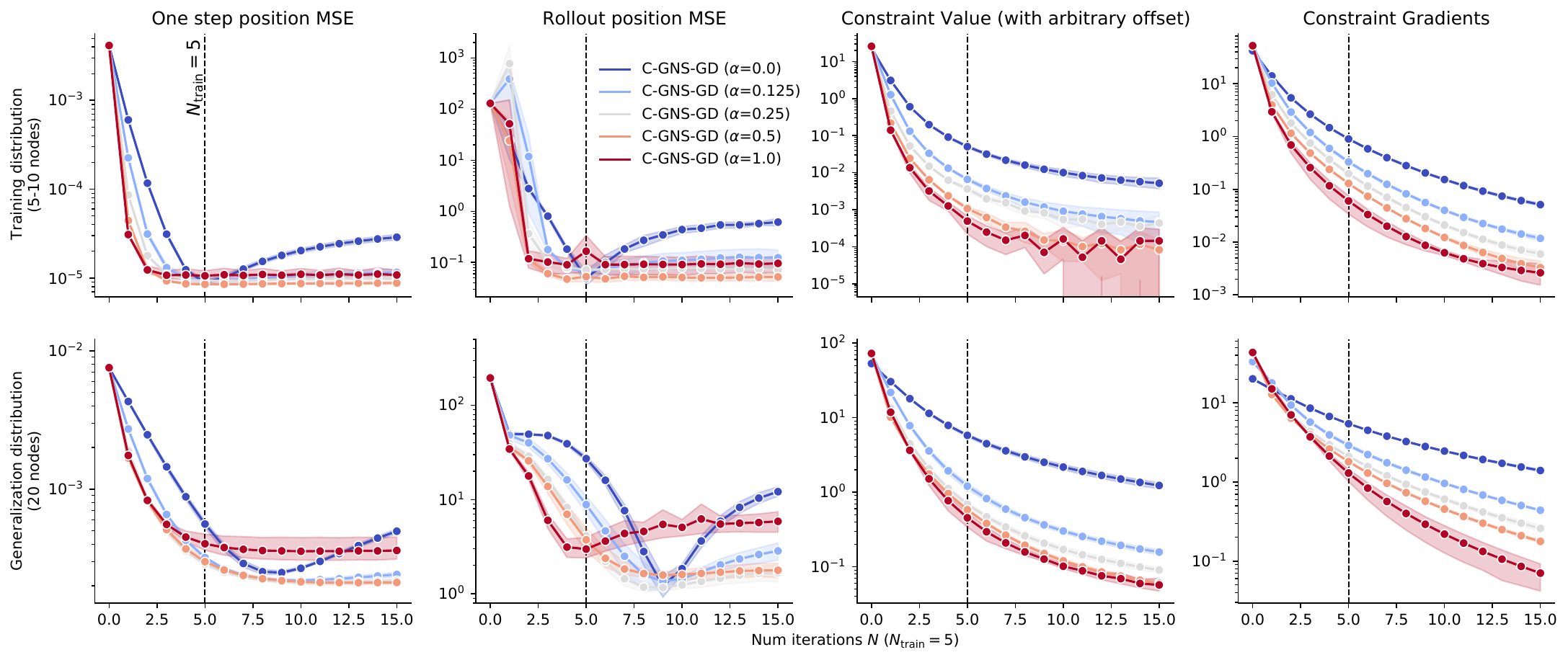}

\caption{\textbf{Using more solver iterations at test time $N$ as function of $\alpha$ on the \dataset{Rope} dataset}. Here we loosely label \constraintmodel ($\alpha=0.0$) to be the model with the loss on the last iteration only. Increasing $\alpha$ leads to faster convergence of the solver. With $\alpha=1$ the model has worse one-step and rollout MSE on the generalization task. Therefore, we chose to use $\alpha=0.25$ for our generalization experiments.}
\label{fig:rope_generalisation_alpha}
\end{figure}

The results on the MSE error and constraint values and gradients with different choices of $\alpha$ are provided on Supplementary Figure~\ref{fig:rope_generalisation_alpha}. We used $\alpha=0.25$ for both \constraintmodel-GD and \iterativemodel in Section \ref{sec:generalization}.

\paragraph{Fast Projections}

Fast Projection~(FP)~algorithm~\cite{goldenthal2007efficient} is a zero-finding algorithm, for constraint functions whose solutions are defined as, $\constr(\context, \update)=0$. FP uses an adaptive step
\begin{align*}
\lambda = -\frac{\constr(\context, \update^{(i)}) }{ \left\lVert\left. \nabla_\update \constr(\context, \update) \right\vert_{\update=\update^{(i)}} \right\rVert^2} .
\end{align*}
Then FP updates the proposed state analogous to our \constraintmodel-GD model,
\begin{align*}
\delta\update = -\lambda \left. \nabla_\update \constr(\context, \update) \right\rvert_{\update=\update^{(i)}}
\end{align*}
\begin{align*}
\update^{(i+1)} = \delta\update + \update^{(i)}
\end{align*}
For our experiments with Fast Projection, we use $N=5$ iterations during training, same as for the gradient descent solver.

\paragraph{\forwardbaselinename}

For the \forwardbaselinename, we use the Graph Network Simulator (GNS) model \citep{sanchezgonzalez20a, pfaff2021learning}.

The $\predictor{}$ takes only the context $\context$ and directly outputs the update $\updatehat$. For \forwardbaselinename, the update $\update$ is set to the acceleration $A_{t+1} = V_{t+1} - V_t = P_{t+1} - 2 P_t + P_{t-1}$, for consistency with the previous work. Then the update rule in the \updater{} becomes $\hat{P}_{t+1} = P_{t} + V_t + \hat{A}_{t+1} = 2 P_t - P_{t-1} + \hat{A}_{t+1}$.

The graph with the context  $\context$  is built similarly to the one for \constraintmodel-GD. After the graph is processed by the graph network, the model uses a per-node MLP decoder to output the update values for each node $(\hat{y}^j)^{j=1\dots J}$

\paragraph{\iterativemodel} In the \iterativemodel, the function $\iterativesim$ takes both the context $\context$ and the proposed update $\update^{(i)}$ and outputs a change to the proposed update $\delta \update$. Then, the model computes the update variable for the next iteration as $\update^{(i+1)} = \update^{(i)} + \delta \update$. The \predictor{} outputs the update variable from the last iteration $\update^{(N)}$.

The input to the $\iterativesim$ at each iteration $i$ is constructed the same way as in \constraintmodel. We take the graph representing the context $\context$ and concatenate the proposed update $Y^{(i)}$ to each node vector. We use a GNS model with a per-node decoder from \citep{sanchezgonzalez20a, pfaff2021learning}  to process the input graph and output $\delta \update$ for each node.

We set meaning of the update $\update$ similarly to the \constraintmodel: $\update$ represents the future velocity $V_{t+1}$ on \dataset{Rope}, \dataset{Bouncing Balls} and \dataset{Bouncing Rigids}; and acceleration $A_{t+1}$ for \dataset{BoxBath}. The corresponding \updater{} is also the same as in \constraintmodel (see Section \ref{sec:implementation_and_notation}). For the first iteration, we initialize $\update^{(0)}$ to the most recent velocity $V_t$, or to a zero vector if $\update$ represents the acceleration (\dataset{BoxBath}).

\paragraph{\textbf{\constraintmodel-GD-$\bm{\constr(\update)}$}} This model is similar to \constraintmodel-GD, except the past states $\context$ are not provided as part of the input. For this ablation, the model directly optimizes the positional information of the future state, similarly to \cite{neural_projections}, rather than velocity or acceleration. Thus, the update $\update$ is set to the positions $P_{t+1}$, as in the Neural Projections model. The corresponding \updater{} becomes simply an identity function.

To construct the input graph, we use only the positional information of the proposed future state $P_{t+1}$ and static properties $Z$. Thus, the node features do not include any information about the past states, nor the proposed approximate future velocity $V_{t+1} = P_{t+1} - P_{t}$. For the edge features, we use the relative displacement vector between the positions of the nodes $j$ and $k$ for the \textit{future} state $e^{jk}_{t+1} = p^k_{t+1} - p^j_{t+1}$.

\paragraph{Neural Projections, C-MLP-FP and C-MLP-GD}

For the models with MLP-based constraint function we use a similar setup to \cite{neural_projections}. We concatenate the features for each node into a single vector and run an MLP to produce a scalar constraint output. For Neural Projections, we include only absolute positions of the nodes into the input. For C-MLP-FP and C-MLP-GD we additionally use the context of the past states, including absolute positions, velocities and distances to the walls for each node. MLP-based models are not provided with explicit pairwise position displacements between the particles, which for the GNN-based models would be in the edge features of the graph. Therefore, we include absolute positions as input to the MLP-based models instead. Next, MLP-based models take a fixed-sized inputs by construction and cannot handle the scenes with variable number of nodes without additional state segmentation schemes. We adapted the MLP-based models to handle scenes with variable number of nodes by padding with zeros up to the maximum state size.

We did not report results for MLP-based models on \dataset{BoxBath}: MLP models take the concatenated input of node embeddings in order, and on the datasets with 1024 nodes like \dataset{BoxBath} the model is likely to overfit to the specific ordering of the particles and would be unlikely to yield competitive performance. Additionally, the input vector optimized by the solver would have $1024 \times 32$ elements resulting in very small gradients for each element.

\paragraph{Hand-designed constraints} We make the hand-designed constraints to be non-negative and such that the minimum of the constraint results in the desired behavior. We parameterize these constraints with a $c_{hand}^j = (\text{ReLU}(-D(p_j)))^2$, where $D(p_j)$ is the signed distance between the position of the node $j$ and the boundary of the obstacle. If $D(p_j) > 0$, the node is outside of the obstacle, and the constraint is zero. If $D(p_j) < 0$, the node overlaps with an obstacle, and the constraint for this node becomes positive. We compute the total constraint for the entire graph as the average of per-node values:  $c_{hand} = \frac{1}{J} \sum_{j=1}^{J} c_{hand}^j$. 
We optimize the weighted sum of the learned and hand-designed constraints: $c_{hand} + W c$, where $W$ is a constant weight, selected for each hand-designed constraint separately. Note that the hand-designed constraints and the constraints learned by the network can have different scales, and we use the weight $W$ to bring the two constraints roughly on the same scale. We ran a grid search to find the appropriate weight $W$.

In some cases, optimizing the joint constraint resulted in an unexpected behavior: the rope shrinks or expands to avoid the obstacle instead of moving around it or stopping at the obstacle. This effect can be explained as follows. Recall that in the \dataset{Rope} dataset used for training, the ropes do not collide with any obstacles, and the only valid next step is to continue moving under the force of gravity. When we optimize the the hand-designed obstacle constraints together with the learned constraint, the solution would inevitably violate the learned constraint, under which the rope would continue moving under gravity. In some cases, the model chooses to violate the learned constraint by changing the relative distances between the rope nodes, instead of changing the dynamics.

To incentivize the model to preserve the node distances, we add the second hand-designed constraint on the relative distances between the nodes. First, we compute the norm of the distances between each pair of adjacent nodes at the current time point and at the first time point of the simulation. Then we set the constraint to the squared difference between the distance norms at the two time steps, and average for all pairs of adjacent nodes. This constraint is minimized when the distances between each pair of nodes remain the same. See \videos{} for the rollouts with obstacle constraint only and the rollouts with both obstacle and distance constraints.

\subsection{Hyperparameters}
\label{sec:hyperparameters}

\begin{figure}[H]
\centering
\includegraphics[width=\textwidth]{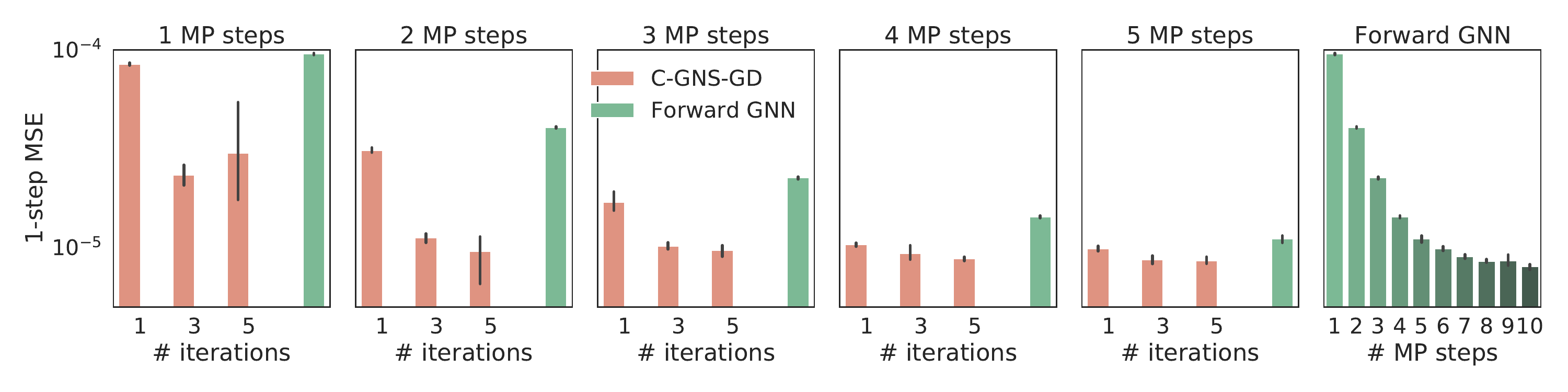}
\includegraphics[width=\textwidth]{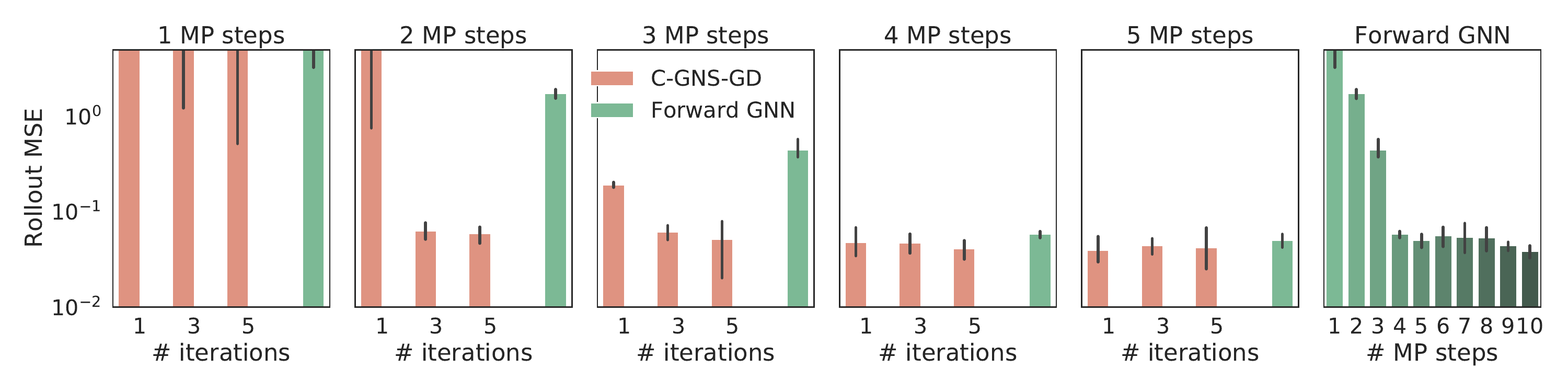}
\caption{\textbf{Test MSE error on \dataset{Rope} as a function of message-passing (MP) steps and solver iterations.} Top row: 1-step test MSE error. Bottom row: full rollout test MSE error. The left five subplots shows performance of \constraintmodel-GD for different numbers of message-passing steps and different number of solver iterations during training. The green bars show the \forwardbaselinename (it does not use solver iterations). The rightmost subplot shows the \forwardbaselinename with 1 to 10 MP steps.}
\label{fig:suppl_barplots}
\end{figure}

\paragraph{Choice of the number of message-passing layers}
We chose the smallest possible number of message passing (MP) steps that would allow the C-GNS family of models to solve the task with 5 optimization iterations (2 MP for \dataset{Rope}, 1MP steps for all other datasets). We then compared this model to a Forward GNN (GNS) with the same architecture and number of parameters for the main result. A full comparison of Forward GNN and C-GNS across multiple values of message passing steps and optimization iterations on \dataset{Rope} is available in Figure \ref{fig:suppl_barplots}.

For all GNNs, we used a residual connection for the nodes and edges on each message-passing layer. GNNs have only node and edge updates and do not use global updates.

\paragraph{Choice of the activation function}
We noticed that the choice of the activation function affected the Fast Projection method more than Gradient Descent. For each dataset, we chose the activation function for which Fast Projection algorithm (C-GNS-FP) was more stable (more random seeds converged). Then we used the same activation function for other models, including Gradient Descent (C-CNS-GD), as the choice of the activation function had little impact on the performance for other models.

\paragraph{Rope} For GNN-based models, we used 2 message-passing steps. We use the latent size of 32 for nodes and edges. The MLPs for processing nodes and edges, as well as node encoder and decoder MLPs, have 3 hidden layers with 256 hidden units each. We used  \texttt{softplus} activation and a LayerNorm \citep{ba2016layer}.

\paragraph{Bouncing Balls} 

For GNN-based models, we used 1 message-passing step. We use the latent size of 32 for nodes and edges. The MLPs for processing nodes and edges, as well as node encoder and decoder MLPs, have 3 hidden layers with 256 hidden units each. We use \texttt{softplus} activation and LayerNorm after every MLP, except the final decoder.

\paragraph{Bouncing Rigids}

For GNN-based models, we used 1 message-passing step. We use the latent size of 32 for nodes and edges. The MLPs for processing nodes and edges, as well as node encoder and decoder MLPs, have 3 hidden layers with 256 hidden units each. We use \texttt{tanh} activation and LayerNorm after every MLP, except the final decoder.

\paragraph{Box Bath}

For GNN-based models, we used 1 message-passing step. All other hyperparameters are as in \cite{sanchezgonzalez20a}. The GNNs' node and edge function MLPs each had 2 hidden layers with 128 hidden units, and hidden node and edge latent sizes of 128 each. We use \texttt{softplus} activation and LayerNorm after every MLP, except the final decoder.

\paragraph{Models with MLP constraints}

For MLP-based models (Neural Projecitons, C-MLP-FP and C-MLP-GD), we used an MLP with 5 hidden layers and 256 hidden units, following the architecture in \cite{neural_projections}. We used \texttt{softplus} activation with no LayerNorm.

\paragraph{Training}

We train the models for 1M steps on \dataset{Rope}, \dataset{Bouncing Balls} and \dataset{Bouncing Rigids}. We used the Adam optimizer with an initial learning rate of 0.0001, and a decay factor of 0.7 applied with a schedule at steps (1e5, 2e5, 4e5, 8e5). We use a batch size of 64. We trained for 2.5M steps for the experiments studying the number of solver iterations. On \dataset{Box Bath} we trained for 2.5M steps with a batch size of 2, and a learning rate starting at 0.001 and decaying continuously at a rate of 0.1 every 1M steps, as in \citet{sanchezgonzalez20a}.

\subsection[Limitations of ``Neural Projections'']{Limitations of ``Neural Projections''~\citep{neural_projections}}
\label{sec:limitations_yang_et_al}

The Neural Projections (NP) by \cite{neural_projections} is another approach involving learned constraint-based simulation. However, it has several fundamental limitations that make it insufficient as a general-purpose learned simulator. Here we elaborate on these limitations, briefly described in the main text.

\begin{figure}[H]
\centering
\includegraphics[width=0.5\textwidth]{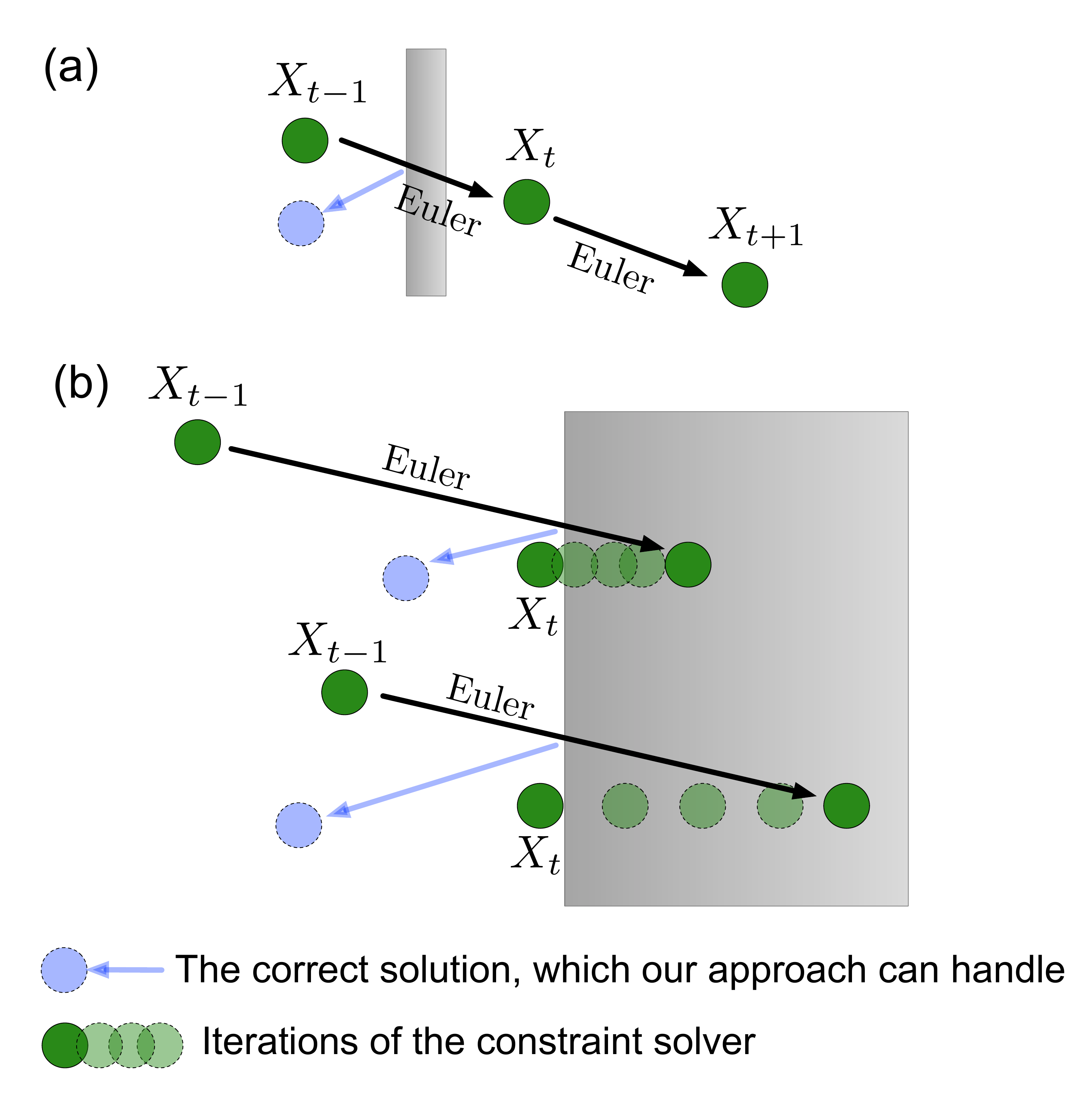}
\caption{Failure cases of Neural Projections~\citep{neural_projections}\\ \textbf{(a) Collision with a thin wall.} The Euler step in NP would propose that the ball moves through the wall. Because NP's constraint $\constr(\text{Euler}(X_{t-1}))$ depends only on the state proposed by the Euler step, it cannot determine that there was a collision between time points $t-1$ and $t$. The ball will remain on the other side of the wall and will continue moving forward in later time steps.\\\textbf{(b) Collision with a thick wall}.
The Euler step in NP would propose that the ball moves into the wall, which should violate the learned constraint. The constraint-solver in NP would then move the ball to the nearest point where the constraint is not violated -- the position where the ball touches the wall.  As the constraint operates only on the position of the ball, but not on the previous positions or velocities, the ball would always be predicted as just touching the wall, rather than bouncing off the wall.}
\label{fig:neural_proj_failure_cases}
\end{figure}

\textbf{Limitation of static constraints} The learned constraint in NP model depends only on the proposed future positions (i.e. the ``static'' state). Thus, NP cannot represent constraints that depend on two or more states across time by construction. For example, NP's constraint function on its own cannot model the time dynamics of a single particle with constant velocity, which is presumably why NP applies an Euler step to initialize the first proposal that is passed to the constraint solver. NP is also in principle not well suited to model elastic collisions properly, as illustrated in Figure~\ref{fig:neural_proj_failure_cases}. In the top scenario, the Euler step proposes the ball moves past the thin wall. Since NP's constraint function does not regard this as a constraint violation, the ball will continue moving ahead as if the wall does not exist. In the bottom scenario, the Euler step places the ball within the wall, and because the nearest constraint-satisfying position for the ball is at the edge of the wall, the approach would, in principle, only correct the position of the ball until the ball stops overlapping the wall as the solution, leaving the ball right next to the wall, regardless of the ball’s initial position and velocity before the wall collision. 

More generally, NP cannot enforce constraints or symmetries defined over time, such as energy preservation: once the Euler step breaks energy preservation, the proposed future state does not contain enough information about the energy of the previous state to be able to identify a constraint violation and resolve it in a way that is consistent with the true dynamics. 

Neural Projection also incorporates external forces, such as gravity, by directly updating the velocities before the Euler step, which is probably necessary because, again, NP's constraint function cannot enforce external effects which involve time (e.g., force and acceleration relate to the second \textit{time} derivative of the position). This is a strong assumption: it means NP must be provided with such temporal effects explicitly, along with the appropriate hard-coded update mechanism, outside of the learnable part of the architecture.
Similar to these examples, there are many other types of dynamics that cannot be expressed as constraint satisfaction over the predicted state from an initial Euler proposal, so overall NP cannot be considered a general-purpose learned simulator.

By contrast, because our approach's constraint function takes both the proposed future state and history as input, i.e., $\constr(X_{\leq t}, Y)$, our method can, in principle, capture any time dynamics which explicit forward simulators can.

\textbf{Hard-coded Euler step} NP relies on an Euler step to generate the initial proposed future state for the solver. Given that forward Euler is a relatively inaccurate integrator, when the Euler proposal is not accurate, the constraint function's lack of access to the previous state makes it difficult for NP to recover.

In our approach, the initial proposal to the solver is less important, because the constraint function can capture all aspects of the dynamics. For this reason, we simply initialized the proposal to the most recent velocity given as input (or zero acceleration for \dataset{BoxBath}). However it is possible to generate initial proposals accounting for external forces or a more sophisticated dynamics prediction mechanism (e.g., an explicit forward simulator).

\textbf{MLP network and the hard-coded grouping technique} NP uses an MLP network as the constraint function, and serializes and concatenates all input features into a vector before passing to the MLP. Generally this approach is not scalable to even moderately large systems (e.g., the $\geq 1000$ nodes in \dataset{BoxBath}) or systems which vary greatly in size (thus requiring significant padding), for similar reasons that serializing images and passing them to an MLP is inferior to CNN- and Transformer-based methods.

\cite{neural_projections} do present a scheme for grouping subsets of the input state and passing them to a shared constraint MLP, which is likely intended to overcome the above weakness, however this is more of a heuristic that takes a partial step toward a more mainstream, full-fledged sharing approach, such as our GNN constraint function.

\textbf{Lack of translation and permutation equivariance}. One of the fundamental properties for modeling physical dynamics is translation equivariance, as the laws of physics do not change based on position in space. Similarly, bodies in a physical system are equivariant to permutations: re-indexing them does not affect the dynamics. \cite{neural_projections}'s implementation of NP lack inductive biases for translation and permutation equivariance, which may lead to poor sample complexity of learning and overfitting.

\begin{figure}[h!]
\begin{center}
\includegraphics[width=\textwidth]{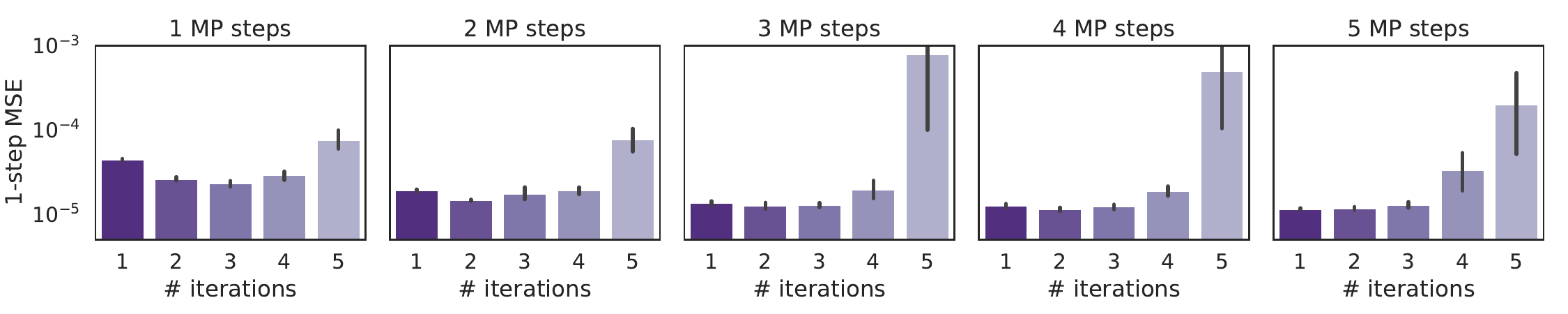}
\includegraphics[width=\textwidth]{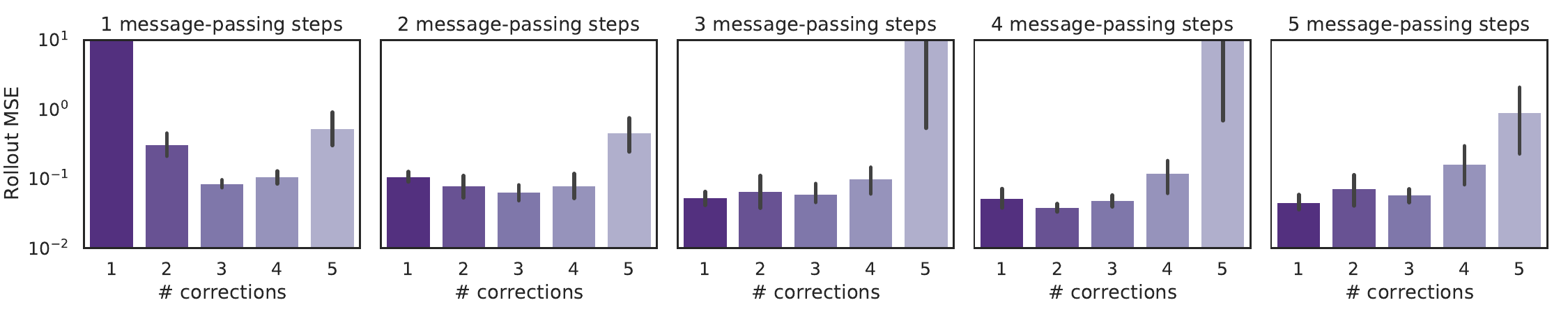}
\end{center}
\caption{Test 1-step MSE and Full Rollout MSE of the \constraintmodel with Neural Projection with different number of message-passing layers and number of constraint solver iterations on the Rope dataset. Top row: position MSE on the full rollout.
}
\label{fig:suppl_barplots_neural proj}
\end{figure}

\textbf{Zero-finding Fast Projections algorithm} NP uses the Fast Projection (FP) algorithm~\citep{goldenthal2007efficient} to find zero points in its constraint function. In practice we found that using FP, i.e., in our C-MLP-FP and C-GNS-FP model variants, to train less stably across seeds, and harder to train with deeper networks (see the variance of random seeds in Figure~\ref{fig:model_comparison}, and trends in Figure~\ref{fig:suppl_barplots_neural proj}). We speculate that because FP's step size is proportional to ratio of the constraint function's value over the squared norm of its gradient, if, early in training, the learned constraint value is large and/or the constraint gradient norm is small, the FP algorithm may take large steps which contribute to the unstable training.

\newpage

\newcommand{\rownamewidth}{0.1\linewidth}
\newcommand{\subfigwidthrollouts}{0.17\linewidth}
\newcommand{\imagewidthrollouts}{0.95\linewidth}
\newcommand{\ropeexample}{rollout_xid30047326_wid36_sample4}
\newcommand{\bouncingballexample}{rollout_xid30074862_wid4_sample4}
\newcommand{\bouncingrigidexample}{rollout_xid30075261_wid3_sample3}
\newcommand{\boxbathexample}{rollout_xid29557492_wid2_sample0}

\section{Supplementary plots and tables}

\subsection{Examples of model rollouts from \constraintmodel-GD models}
\begin{figure}[H]
\centering
    \begin{subfigure}{\rownamewidth}
    Rope\\(GT)
    \end{subfigure}
    \begin{subfigure}{\subfigwidthrollouts}
        \includegraphics[trim={100 250 100 50},clip,width=\imagewidthrollouts]{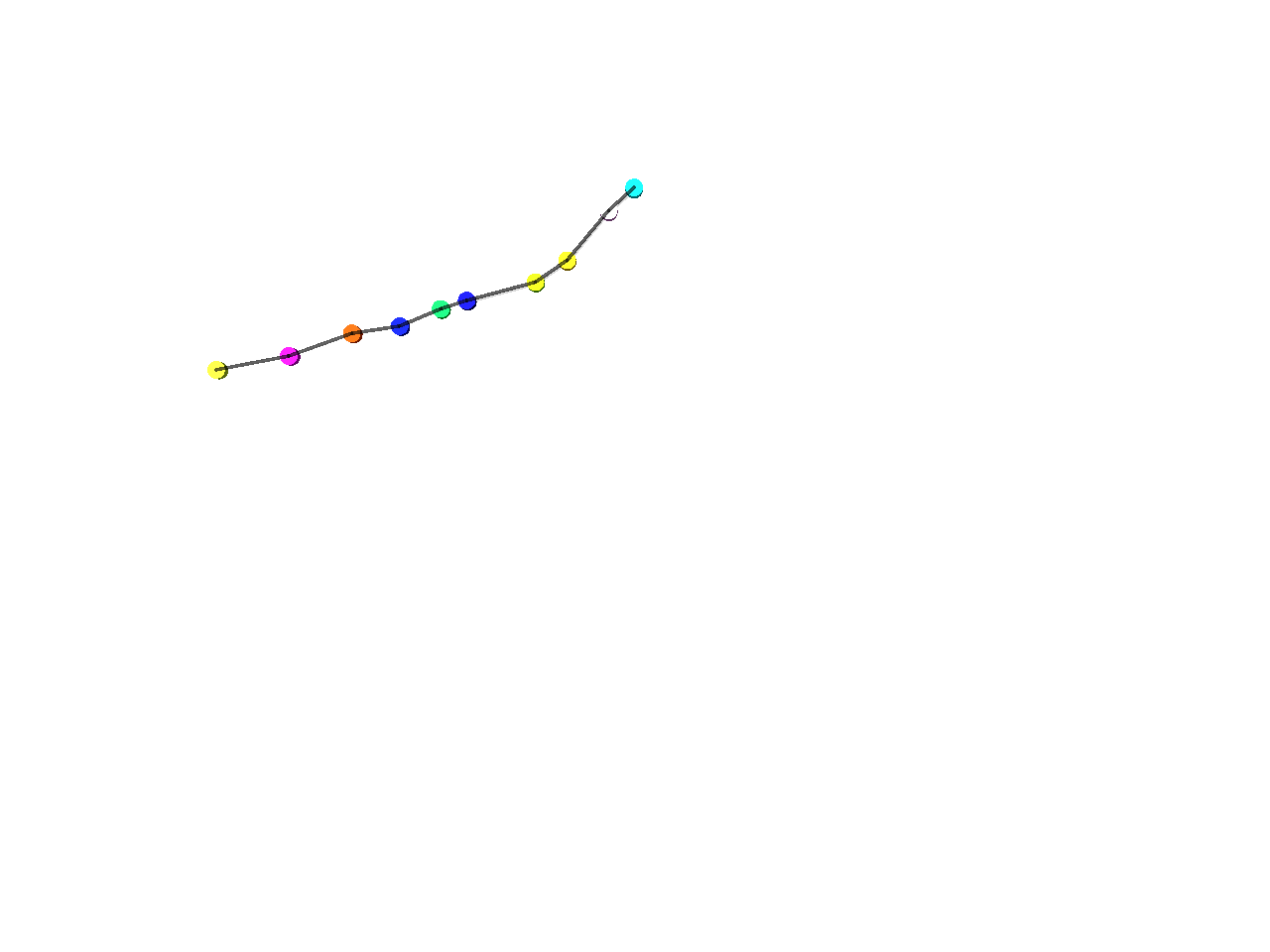}
    \end{subfigure}
    \begin{subfigure}{\subfigwidthrollouts}
        \includegraphics[trim={100 250 100 50},clip,width=\imagewidthrollouts]{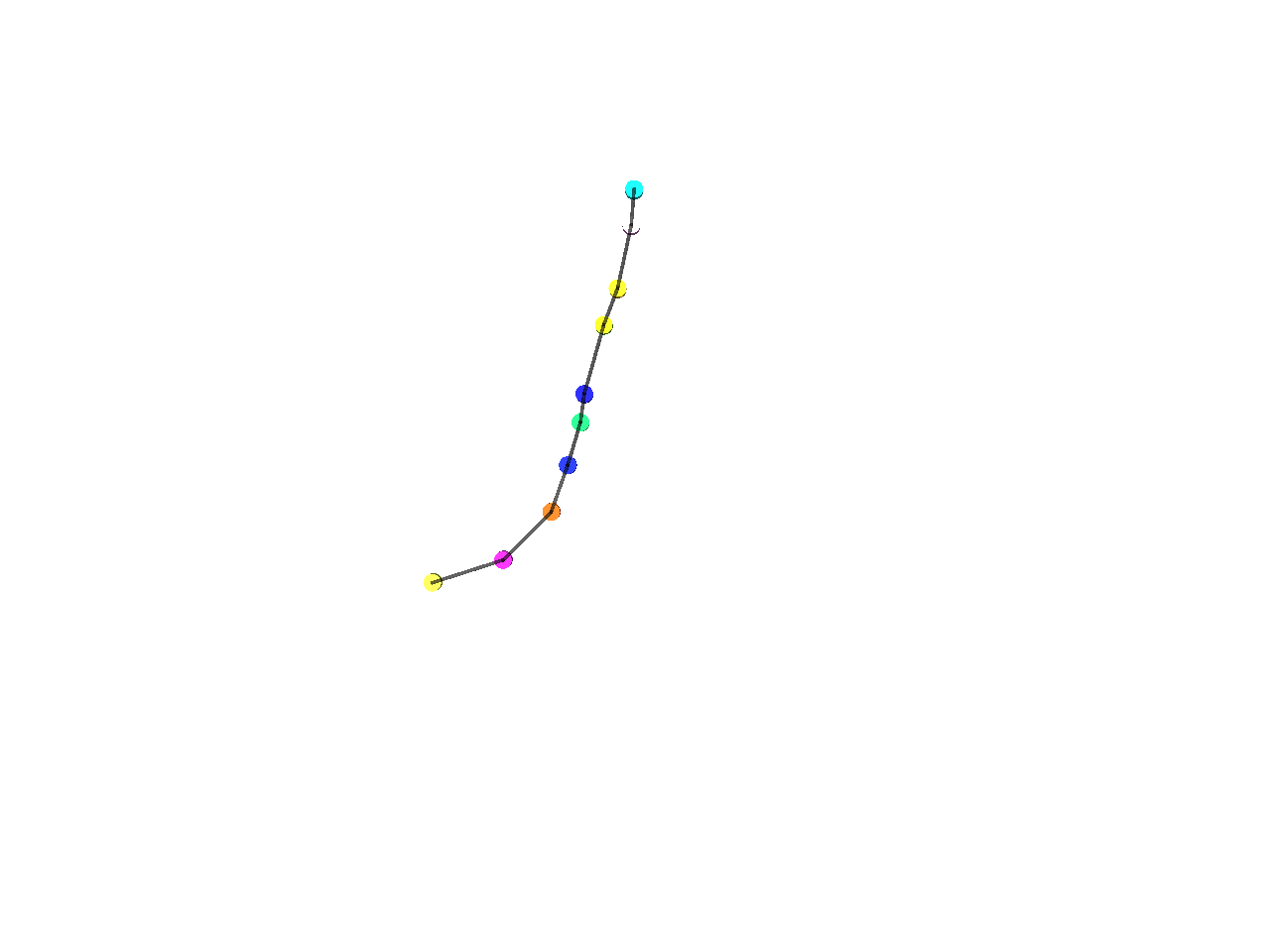}
    \end{subfigure}
    \begin{subfigure}{\subfigwidthrollouts}
       \includegraphics[trim={100 250 100 50},clip,width=\imagewidthrollouts]{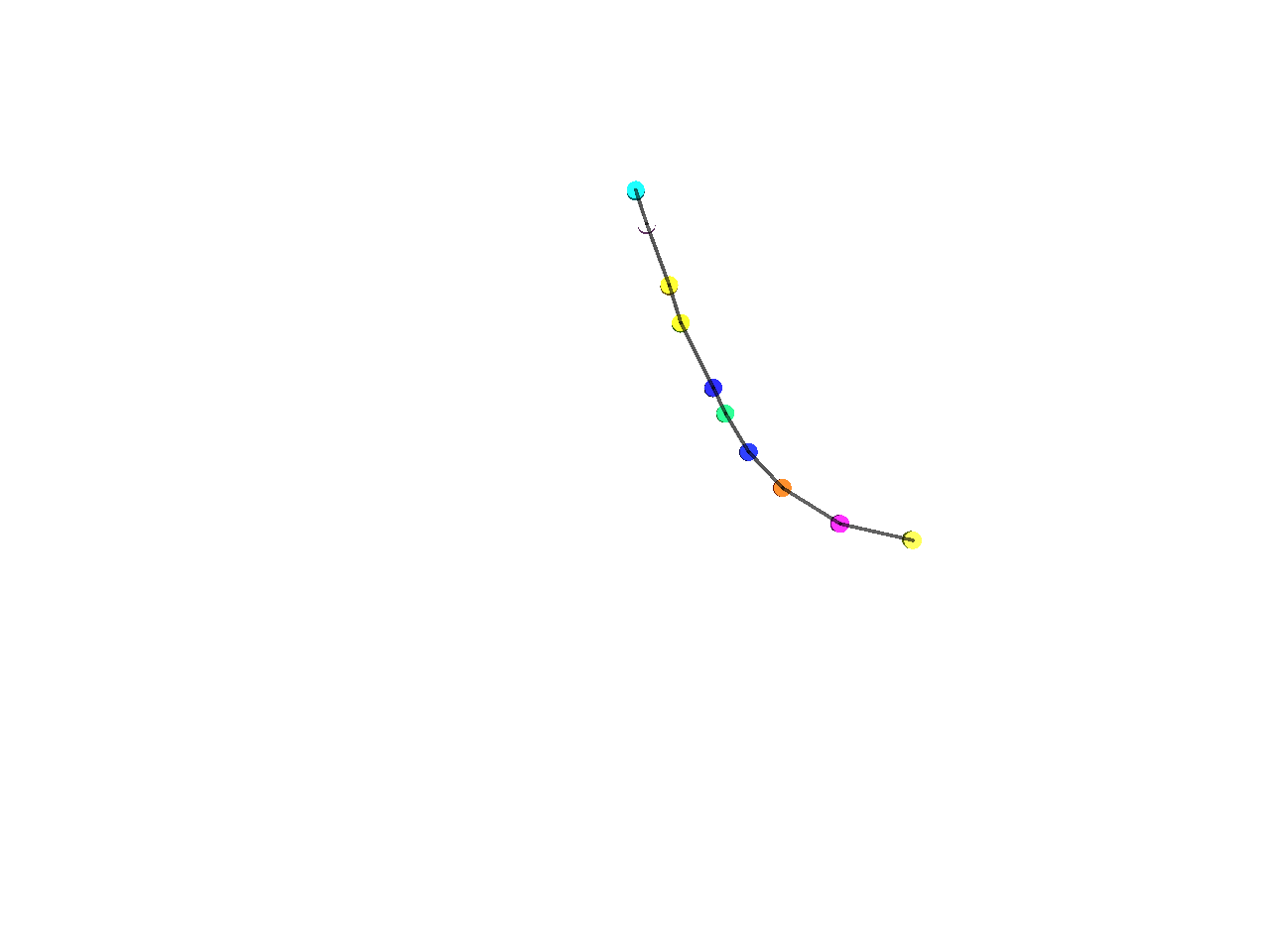}
    \end{subfigure}
    \begin{subfigure}{\subfigwidthrollouts}
       \includegraphics[trim={100 250 100 50},clip,width=\imagewidthrollouts]{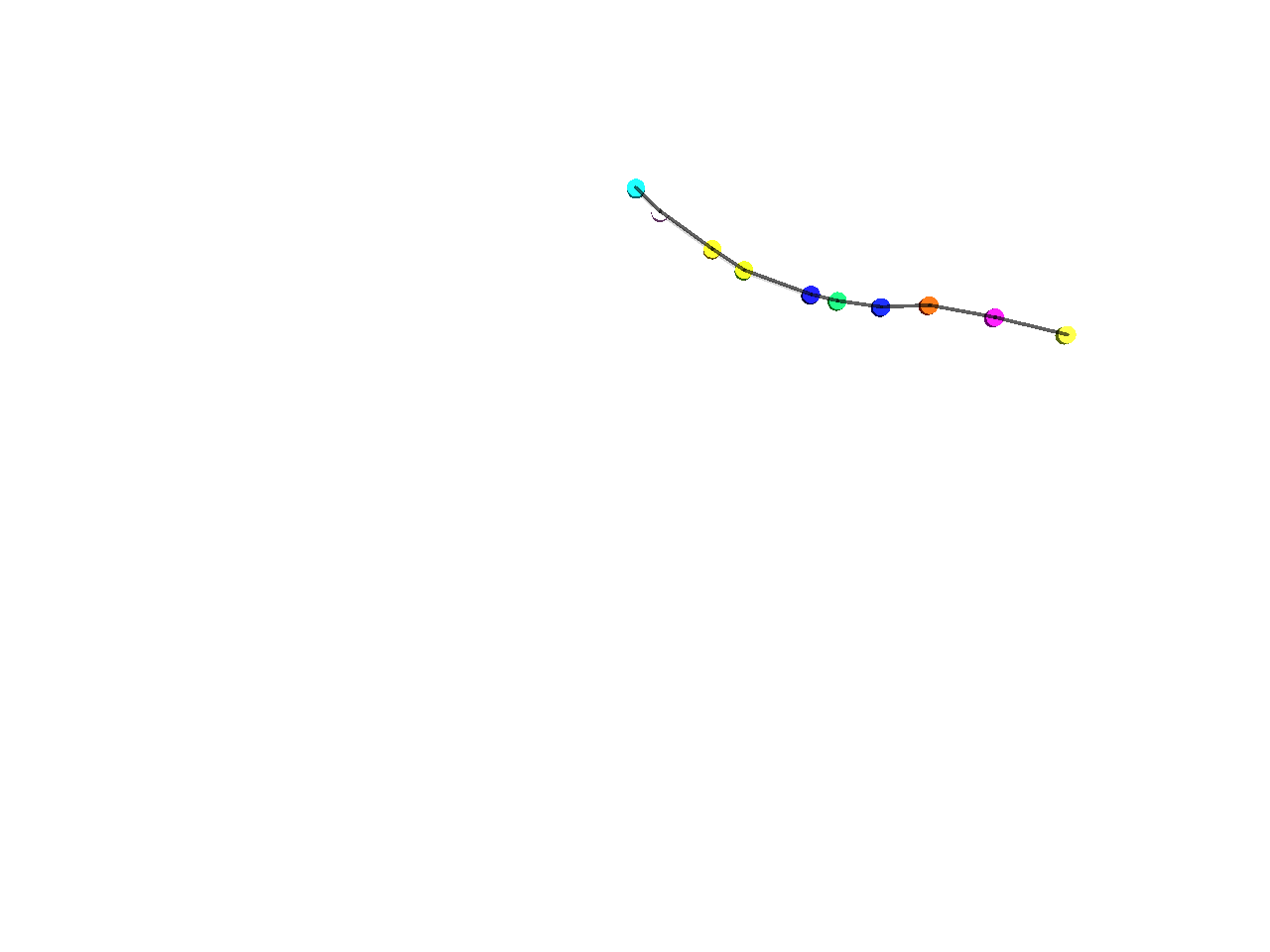}
    \end{subfigure}
    \begin{subfigure}{\subfigwidthrollouts}
       \includegraphics[trim={100 250 100 50},clip,width=\imagewidthrollouts]{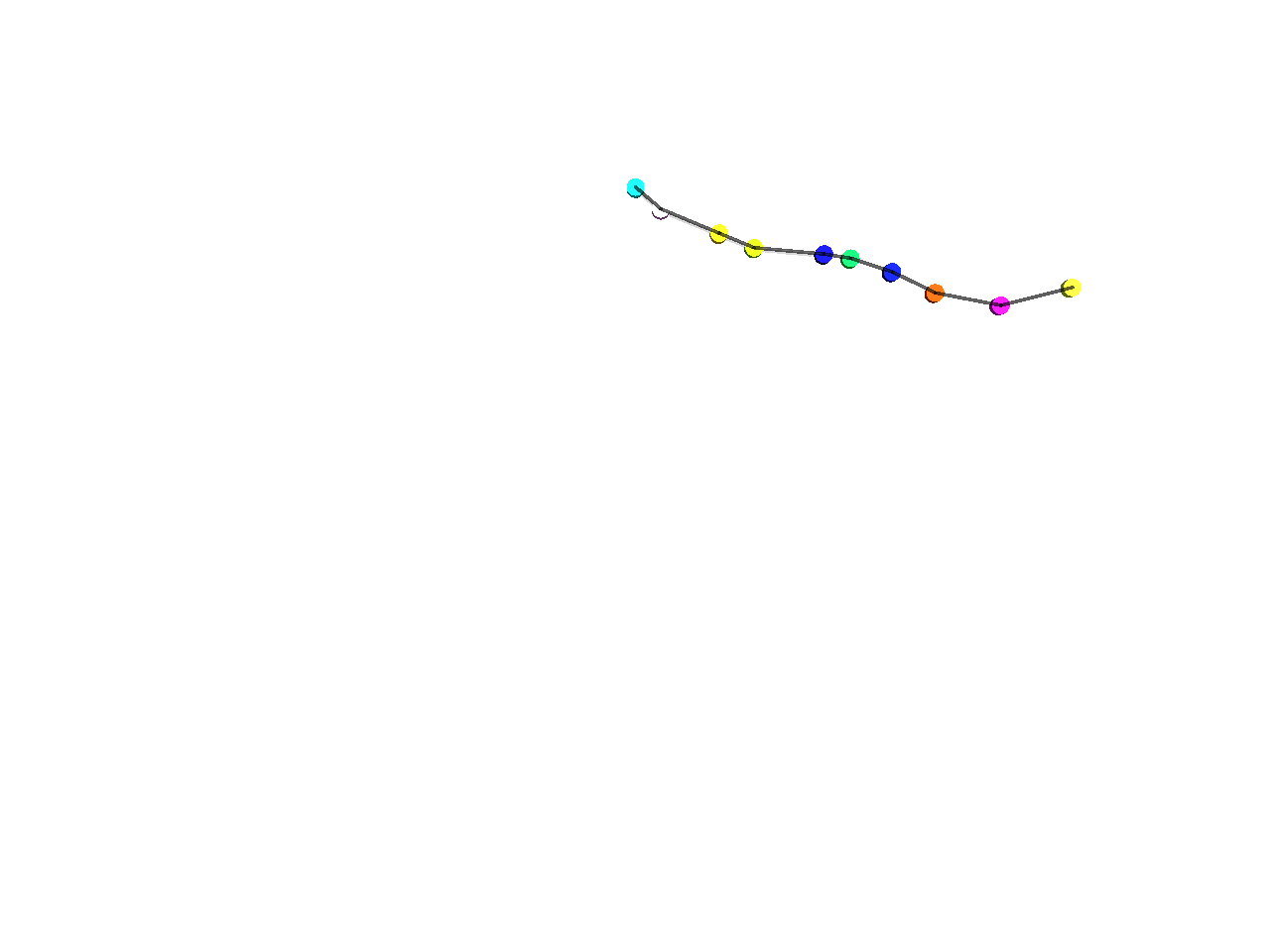}
    \end{subfigure}
    \hfill
    \begin{subfigure}{\rownamewidth}
    Rope\\(Rollout)
    \end{subfigure}
    \begin{subfigure}{\subfigwidthrollouts}
        \includegraphics[trim={120 250 120 50},clip,width=\imagewidthrollouts]{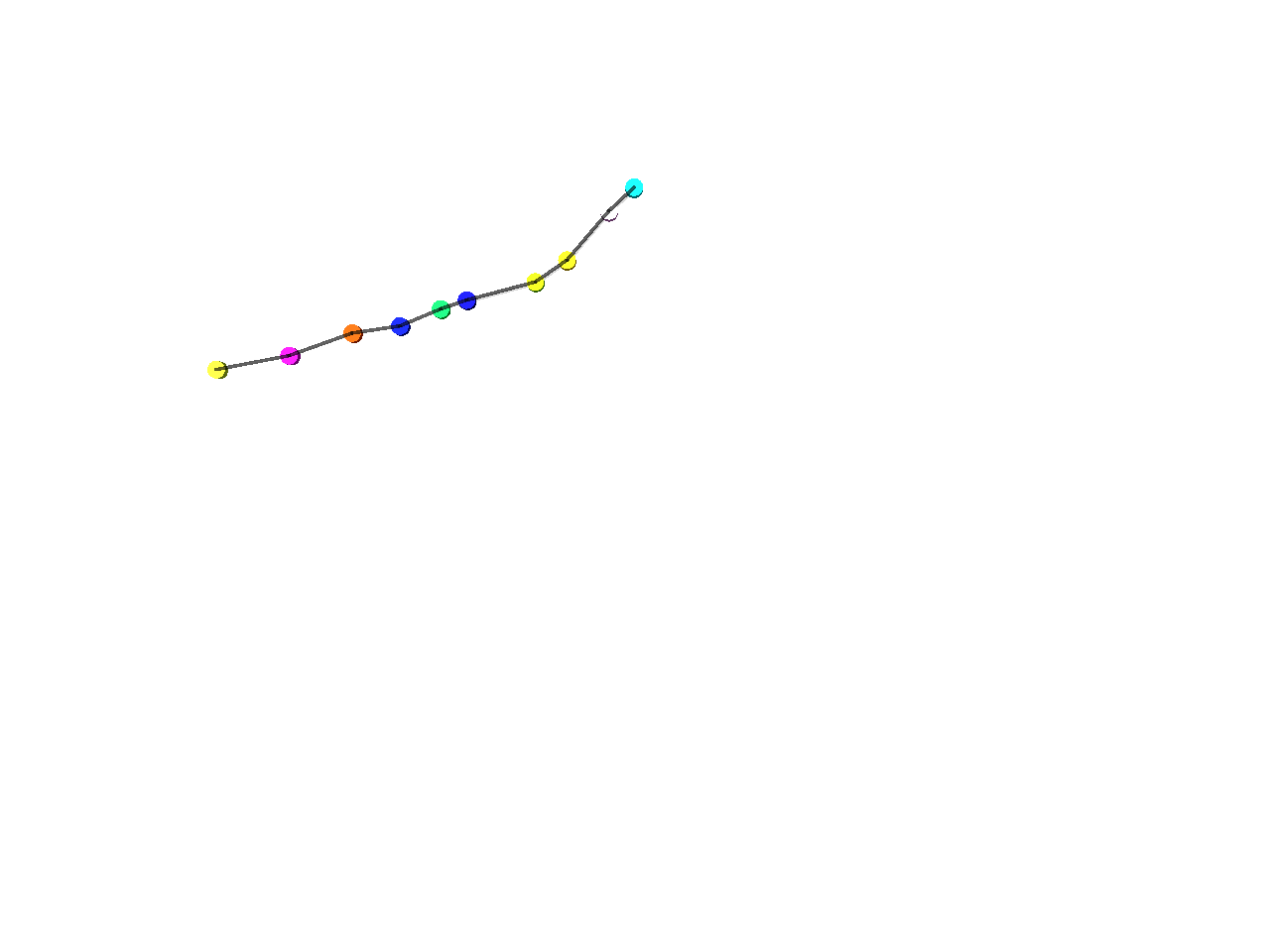}
    \end{subfigure}
    \begin{subfigure}{\subfigwidthrollouts}
        \includegraphics[trim={100 250 100 50},clip,width=\imagewidthrollouts]{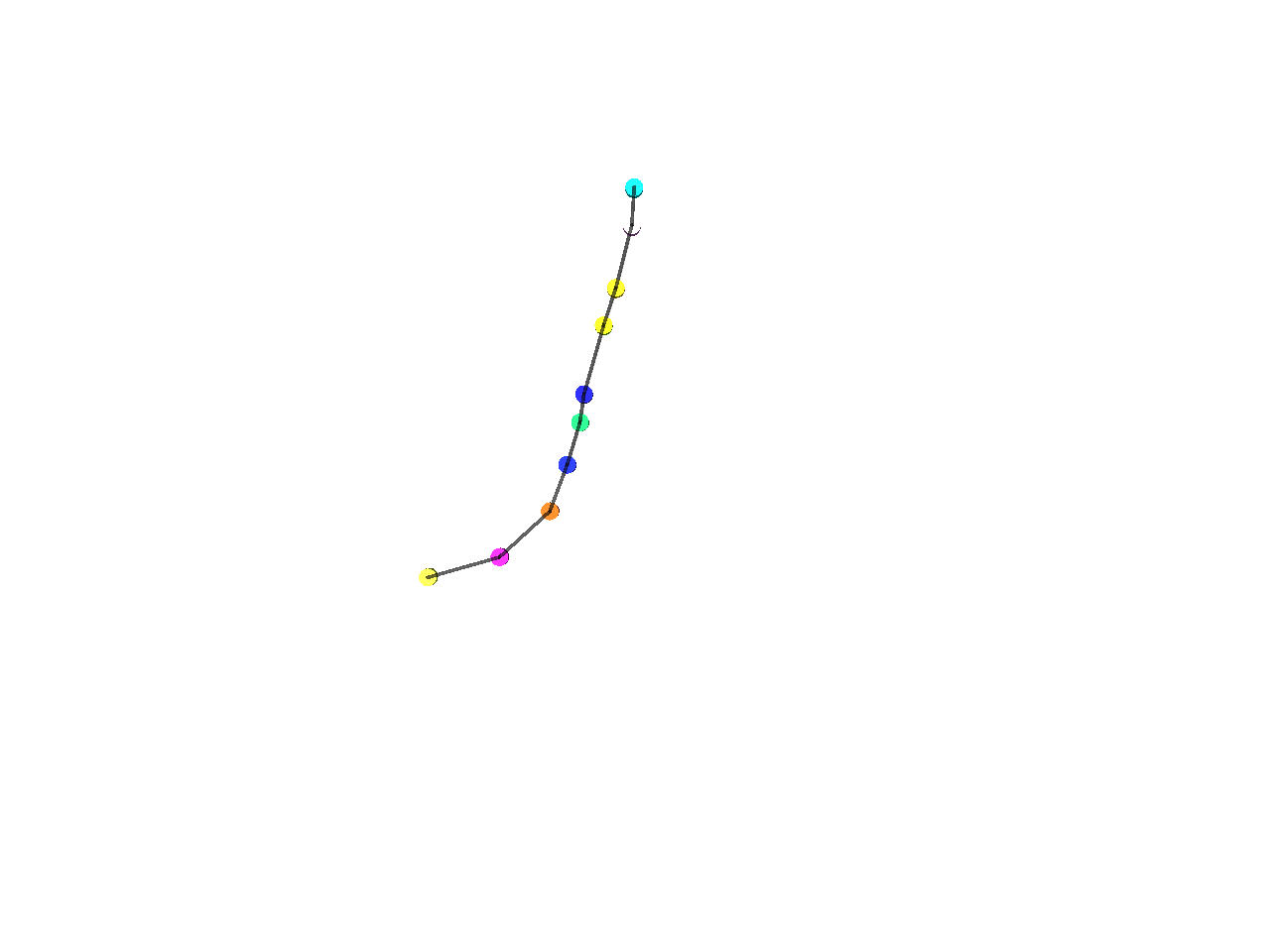}
    \end{subfigure}
    \begin{subfigure}{\subfigwidthrollouts}
       \includegraphics[trim={100 250 100 50},clip,width=\imagewidthrollouts]{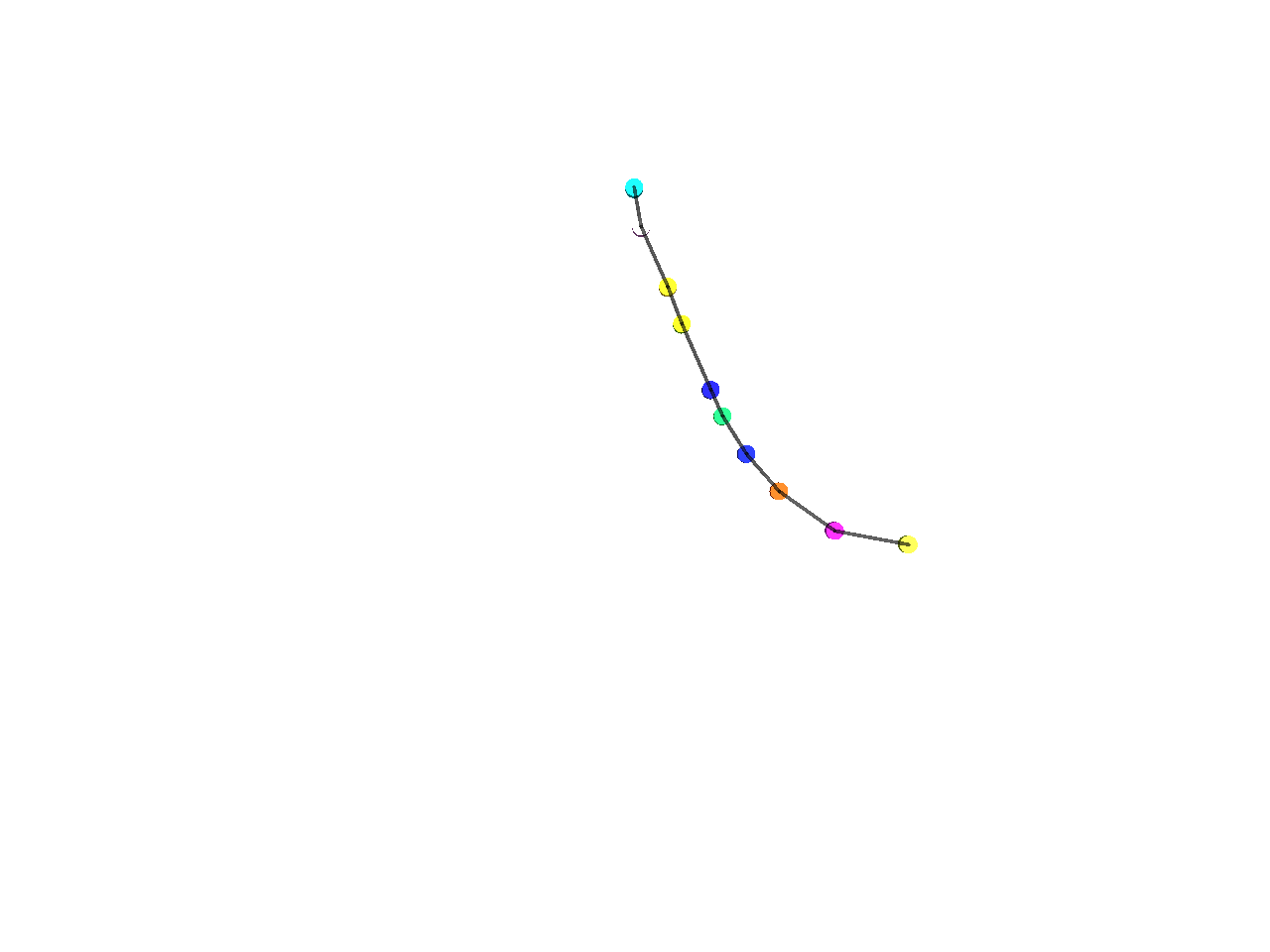}
    \end{subfigure}
    \begin{subfigure}{\subfigwidthrollouts}
       \includegraphics[trim={100 250 100 50},clip,width=\imagewidthrollouts]{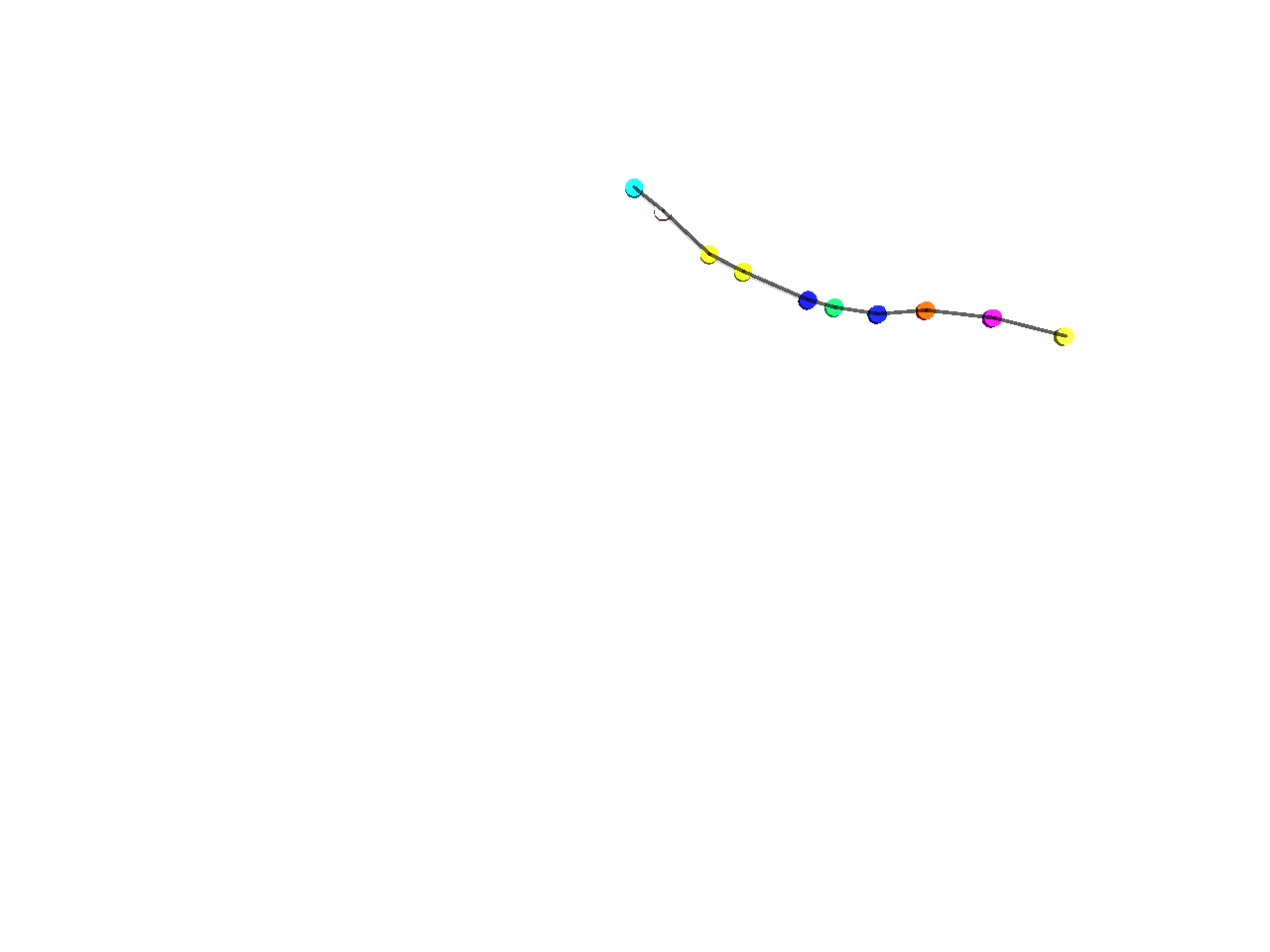}
    \end{subfigure}
    \begin{subfigure}{\subfigwidthrollouts}
       \includegraphics[trim={100 250 100 50},clip,width=\imagewidthrollouts]{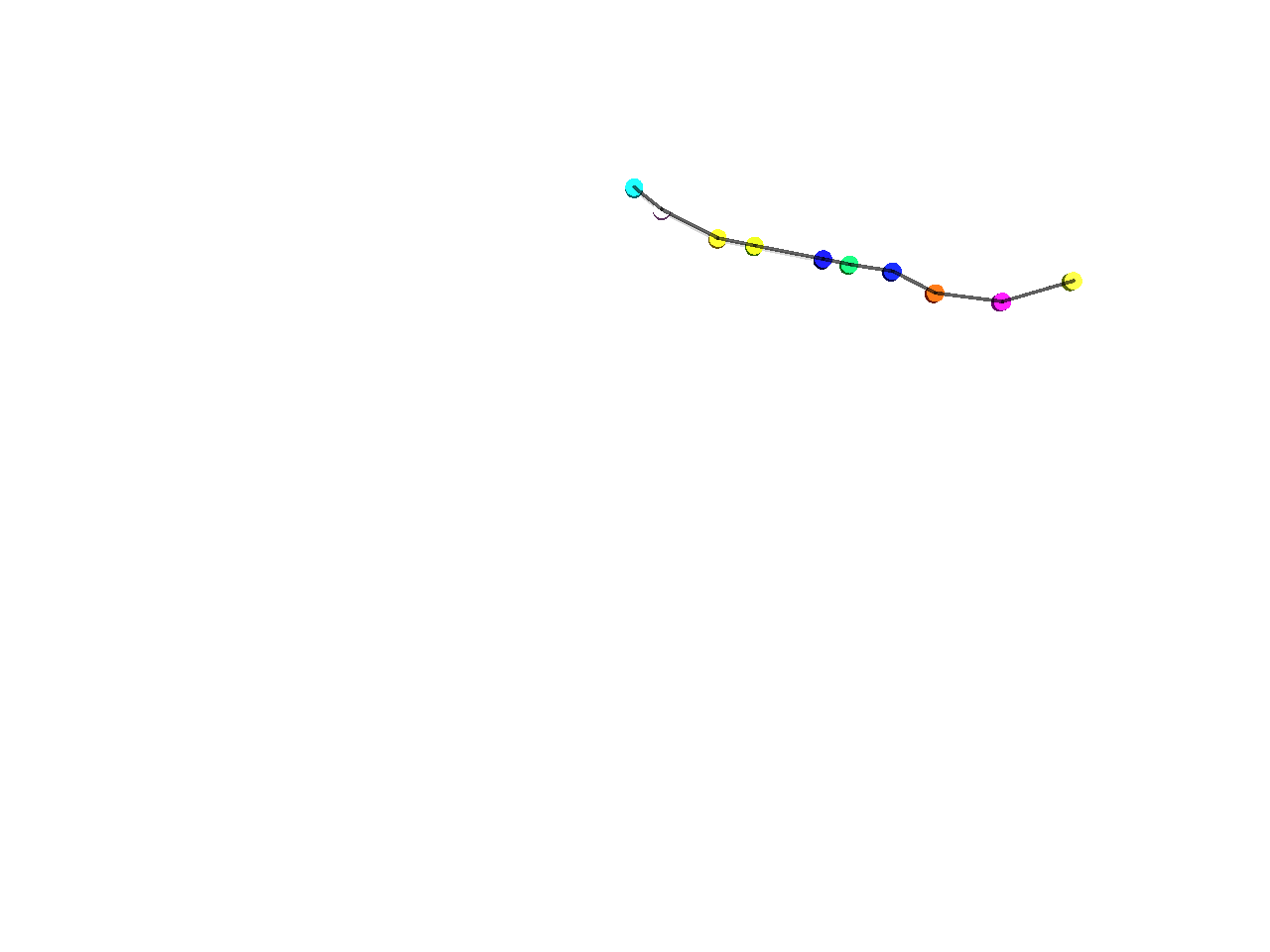}
    \end{subfigure}
    \hfill
    \begin{subfigure}{\rownamewidth}
        Bouncing\\Balls\\(GT)
    \end{subfigure}
    \begin{subfigure}{\subfigwidthrollouts}
        \includegraphics[width=\imagewidthrollouts]{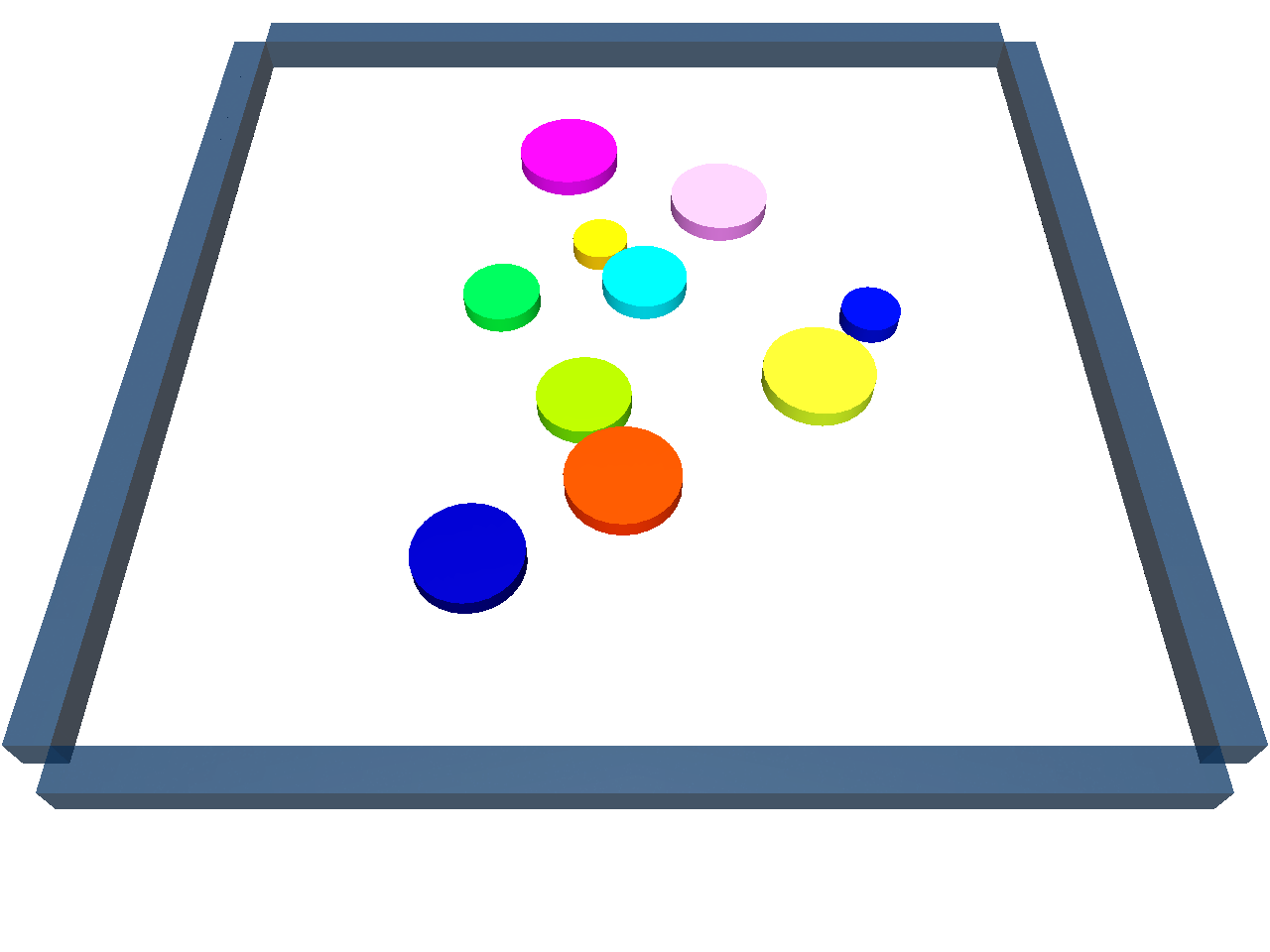}
    \end{subfigure}
    \begin{subfigure}{\subfigwidthrollouts}
         \includegraphics[width=\imagewidthrollouts]{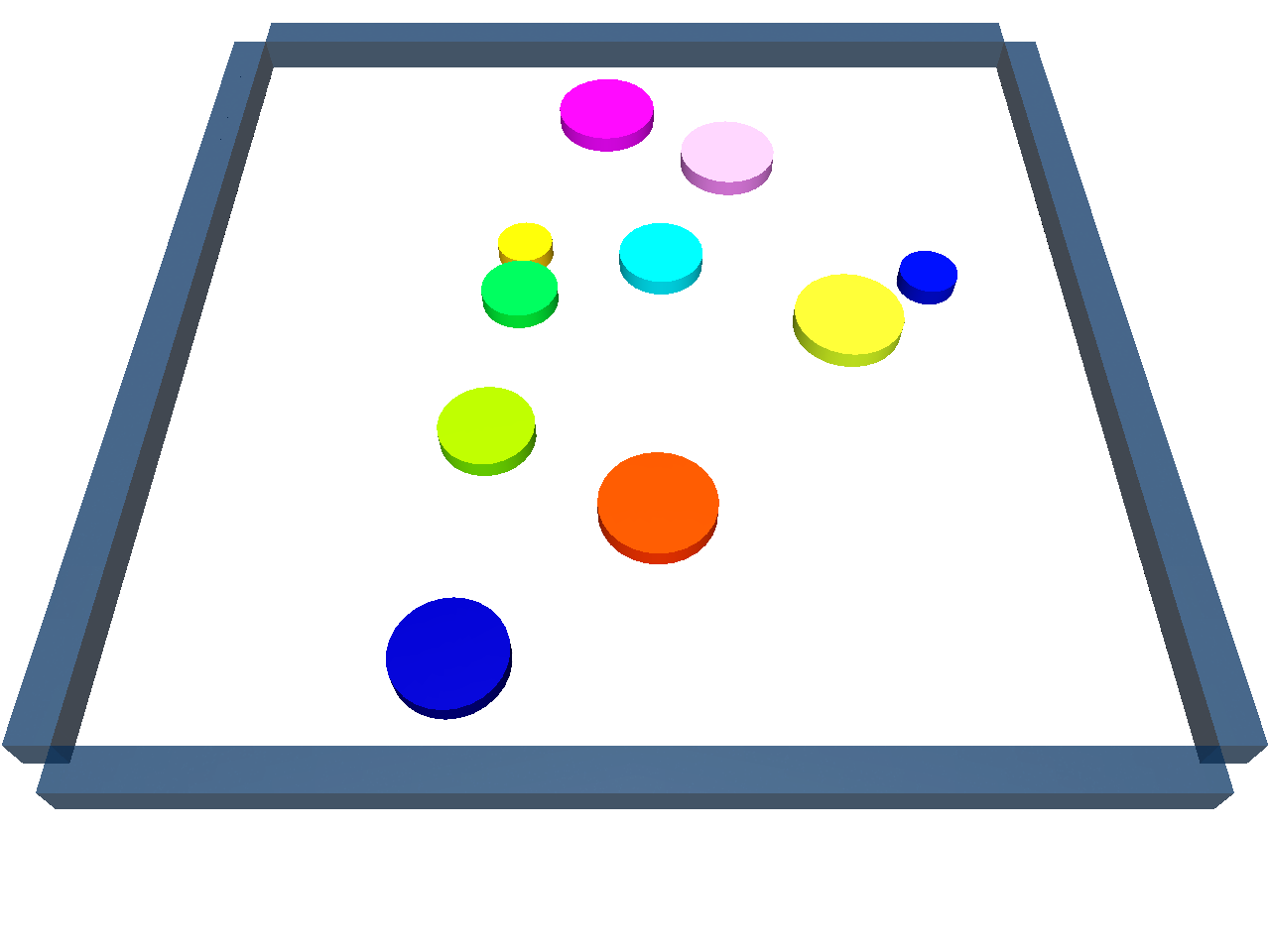}
    \end{subfigure}
    \begin{subfigure}{\subfigwidthrollouts}
         \includegraphics[width=\imagewidthrollouts]{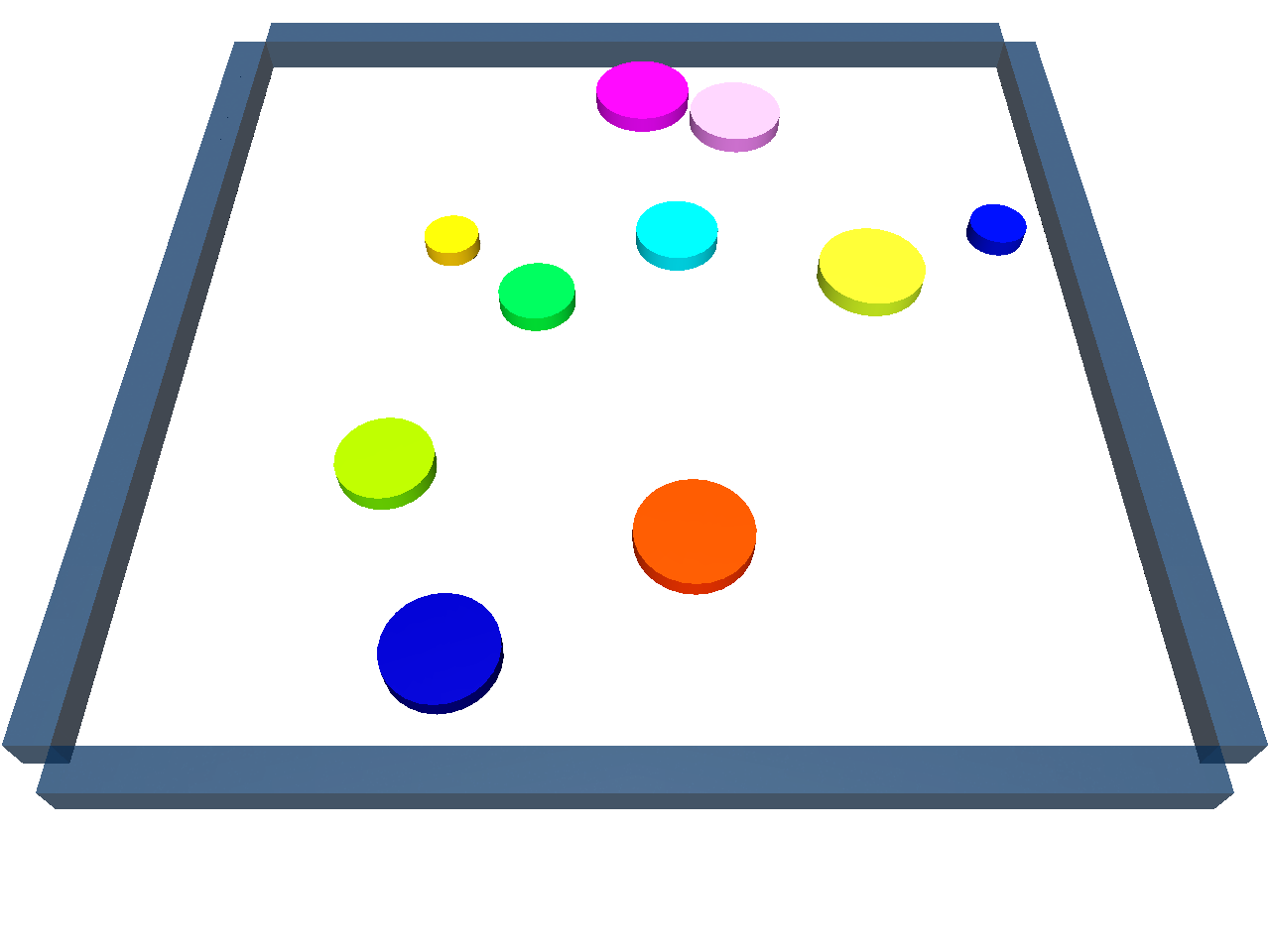}
    \end{subfigure}
    \begin{subfigure}{\subfigwidthrollouts}
         \includegraphics[width=\imagewidthrollouts]{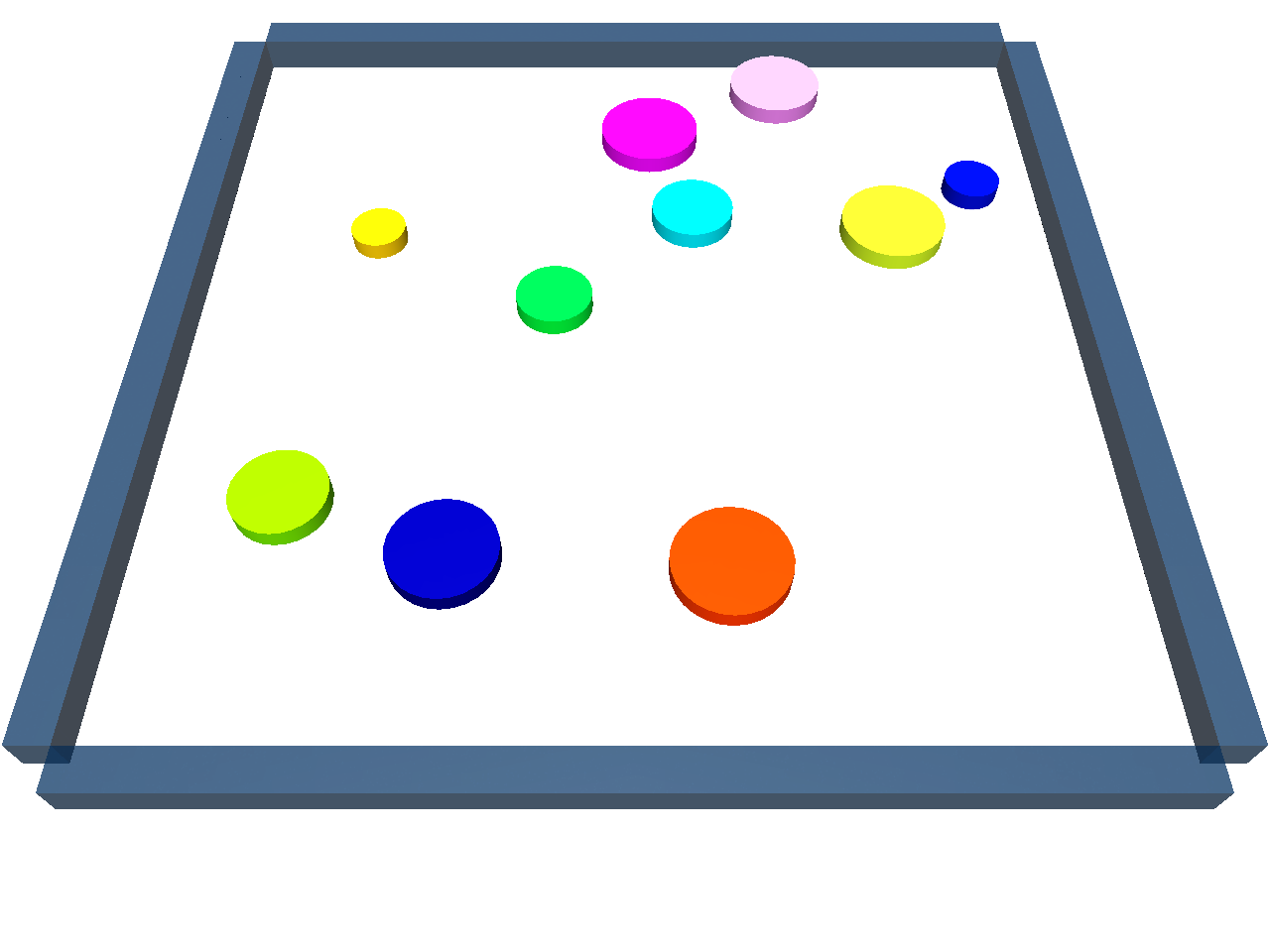}
    \end{subfigure}
    \begin{subfigure}{\subfigwidthrollouts}
         \includegraphics[width=\imagewidthrollouts]{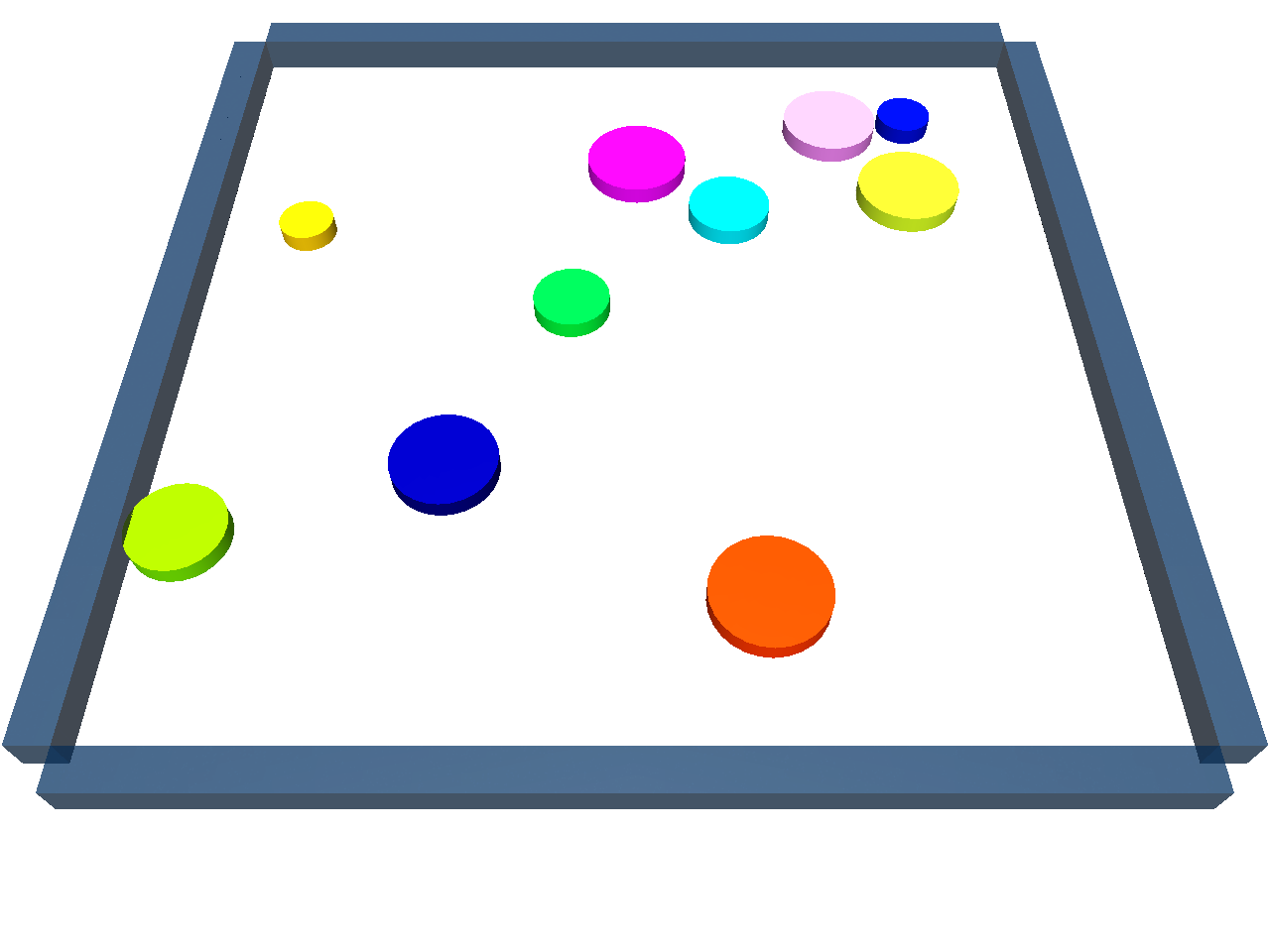}
    \end{subfigure}
    \hfill
    \begin{subfigure}{\rownamewidth}
        Bouncing\\Balls\\(Rollout)
    \end{subfigure}
    \begin{subfigure}{\subfigwidthrollouts}
        \includegraphics[width=\imagewidthrollouts]{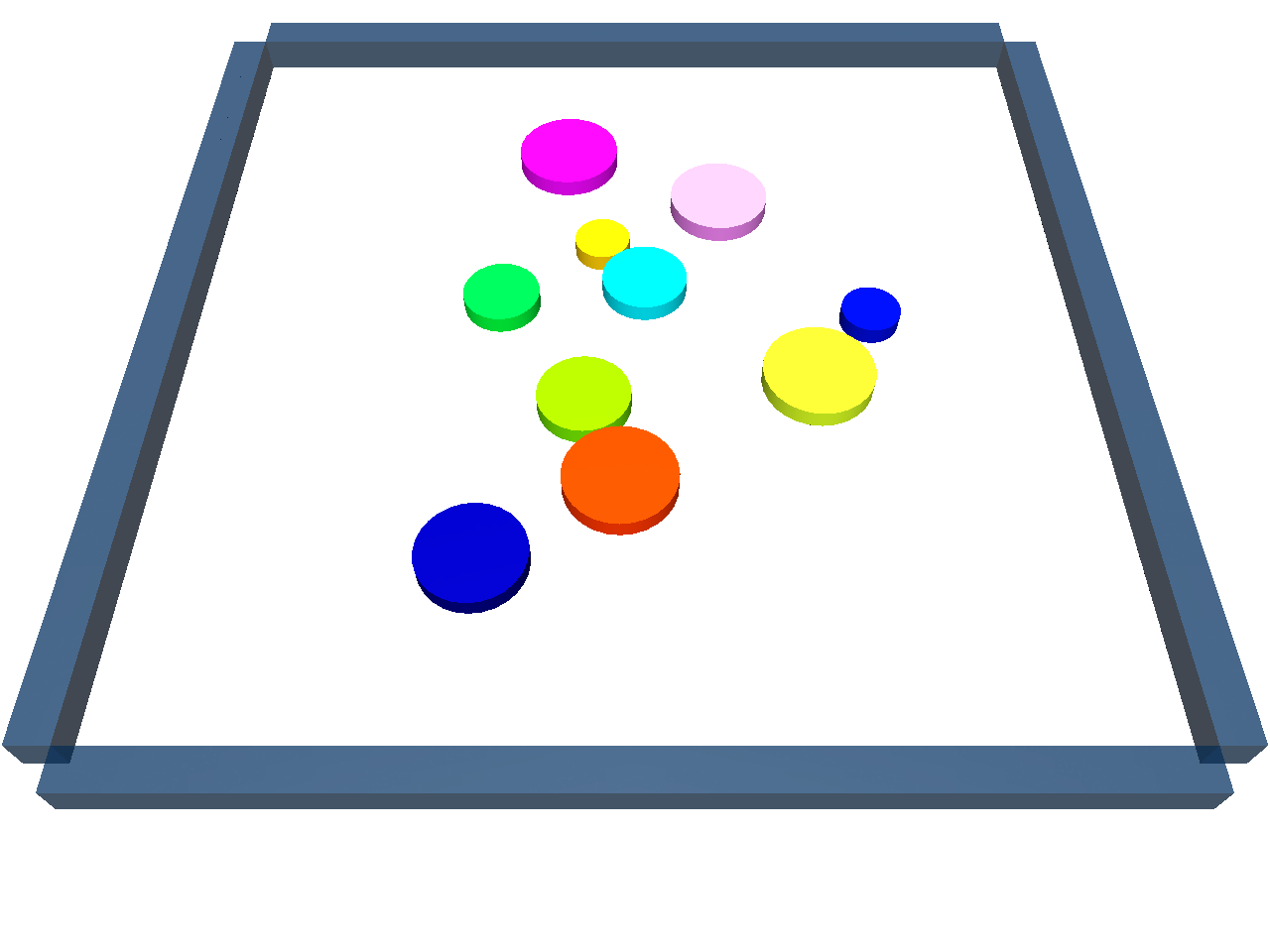}
    \end{subfigure}
    \begin{subfigure}{\subfigwidthrollouts}
         \includegraphics[width=\imagewidthrollouts]{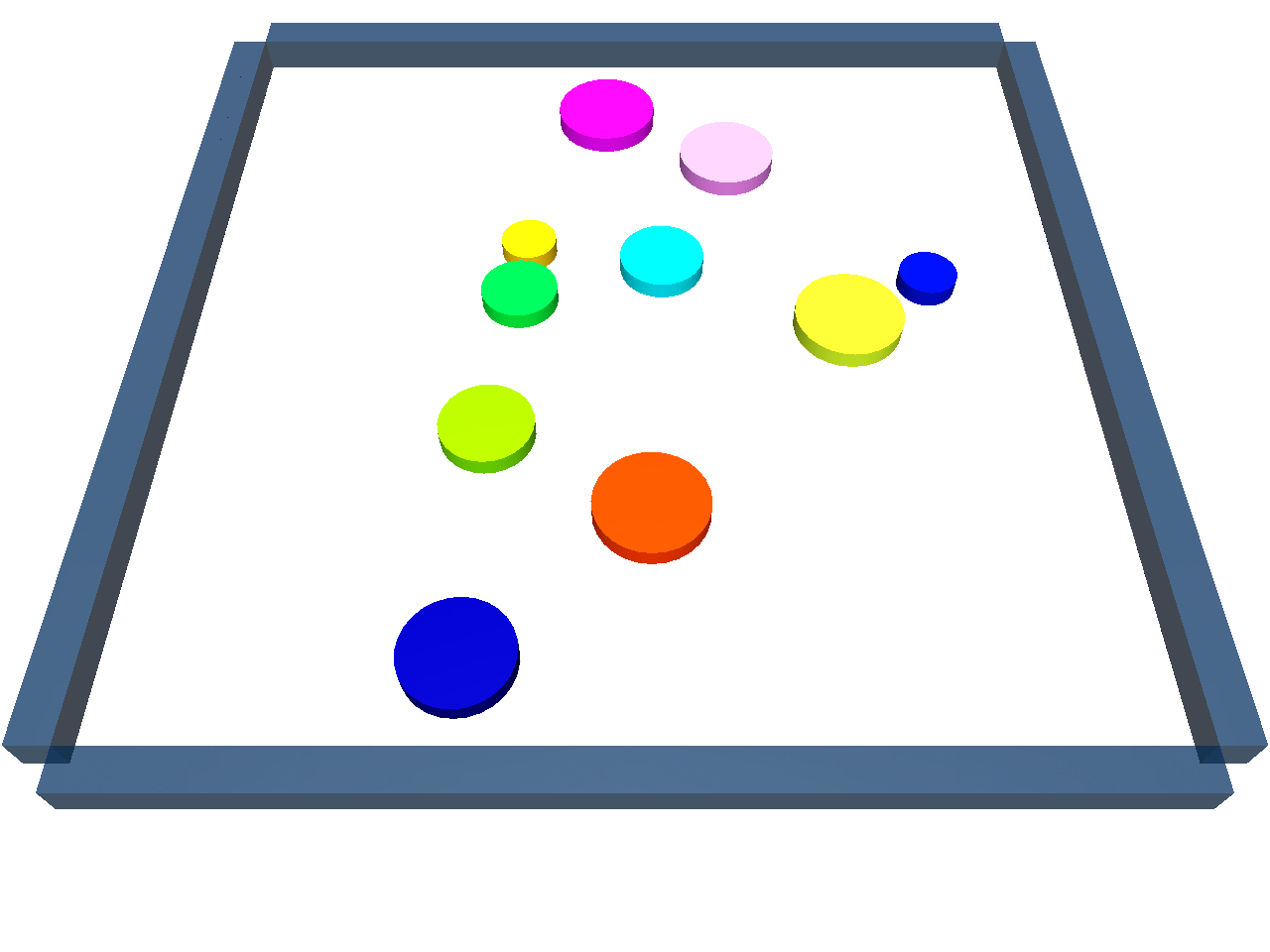}
    \end{subfigure}
    \begin{subfigure}{\subfigwidthrollouts}
         \includegraphics[width=\imagewidthrollouts]{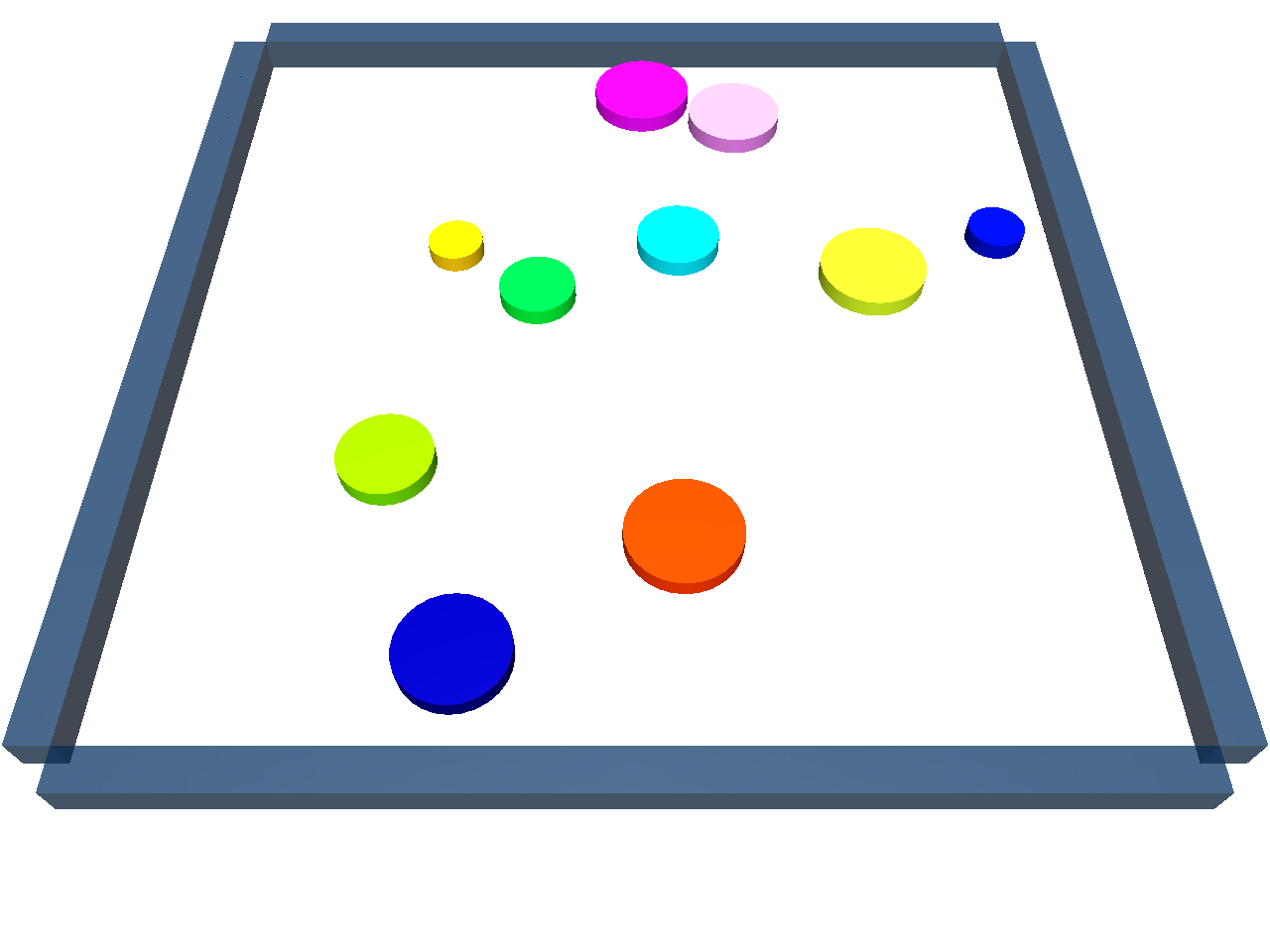}
    \end{subfigure}
    \begin{subfigure}{\subfigwidthrollouts}
         \includegraphics[width=\imagewidthrollouts]{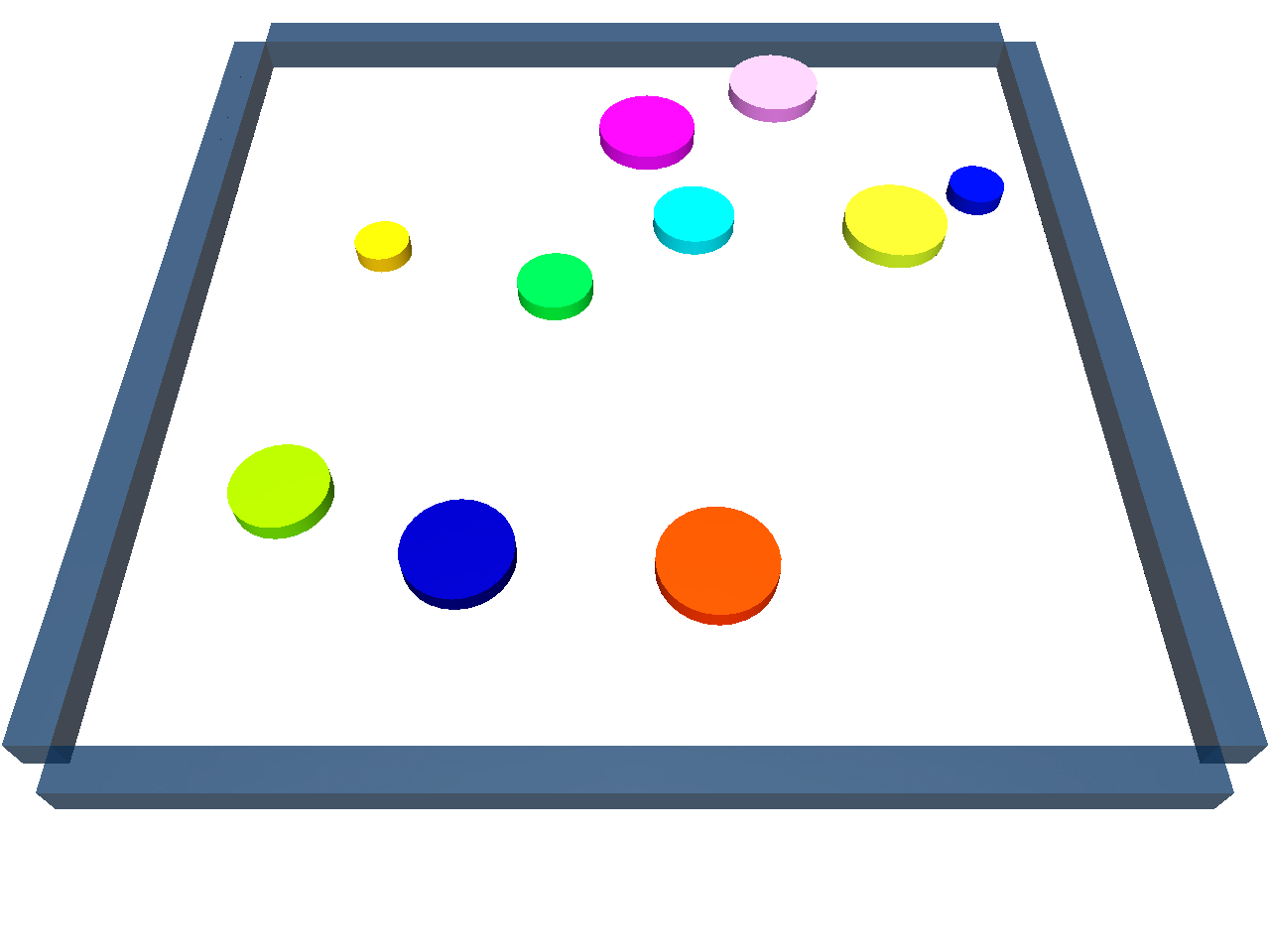}
    \end{subfigure}
    \begin{subfigure}{\subfigwidthrollouts}
         \includegraphics[width=\imagewidthrollouts]{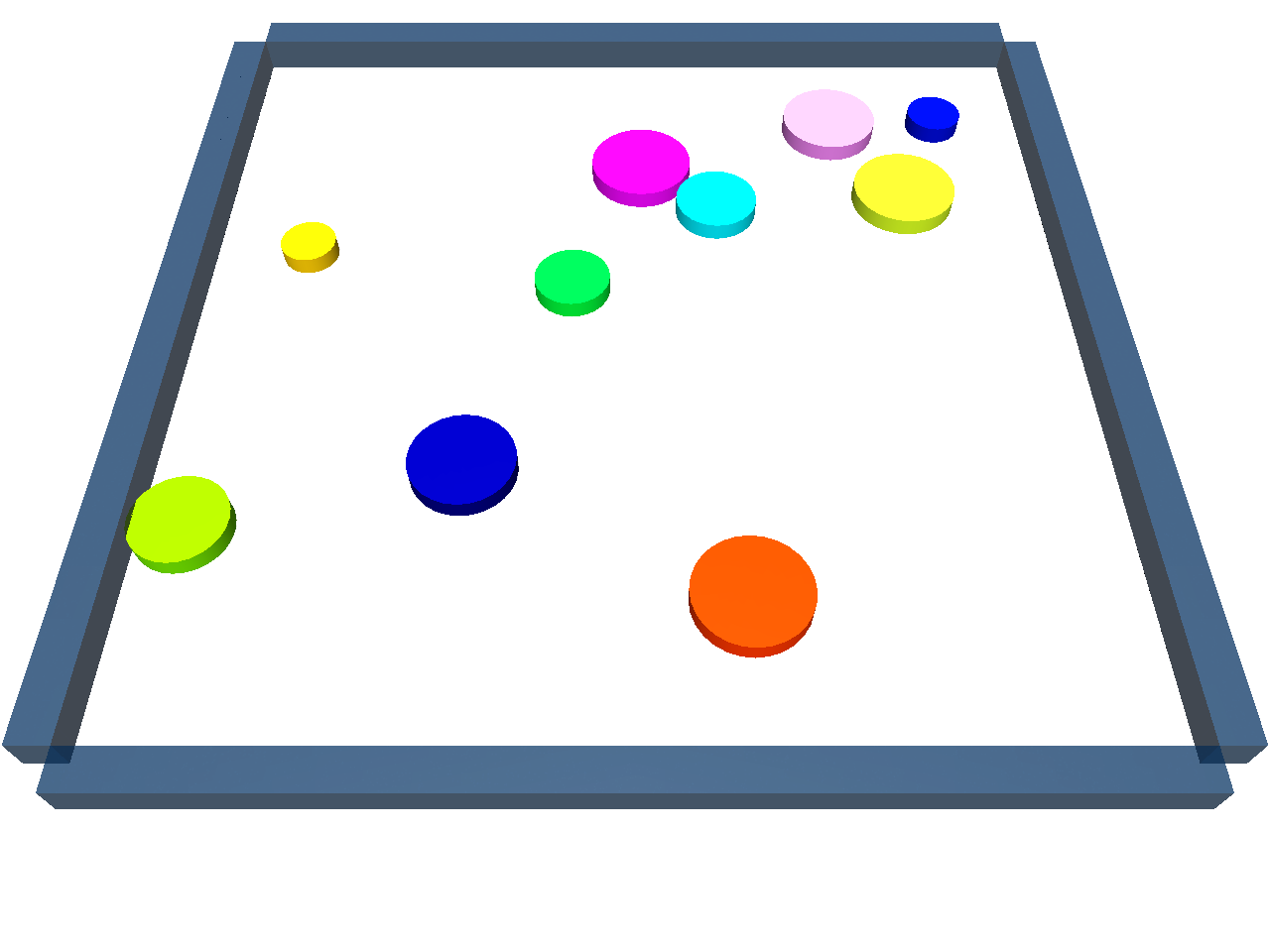}
    \end{subfigure}
    \hfill
    \begin{subfigure}{\rownamewidth}
        Bouncing\\Rigids\\(GT)
    \end{subfigure}
    \begin{subfigure}{\subfigwidthrollouts}
        \includegraphics[width=\imagewidthrollouts]{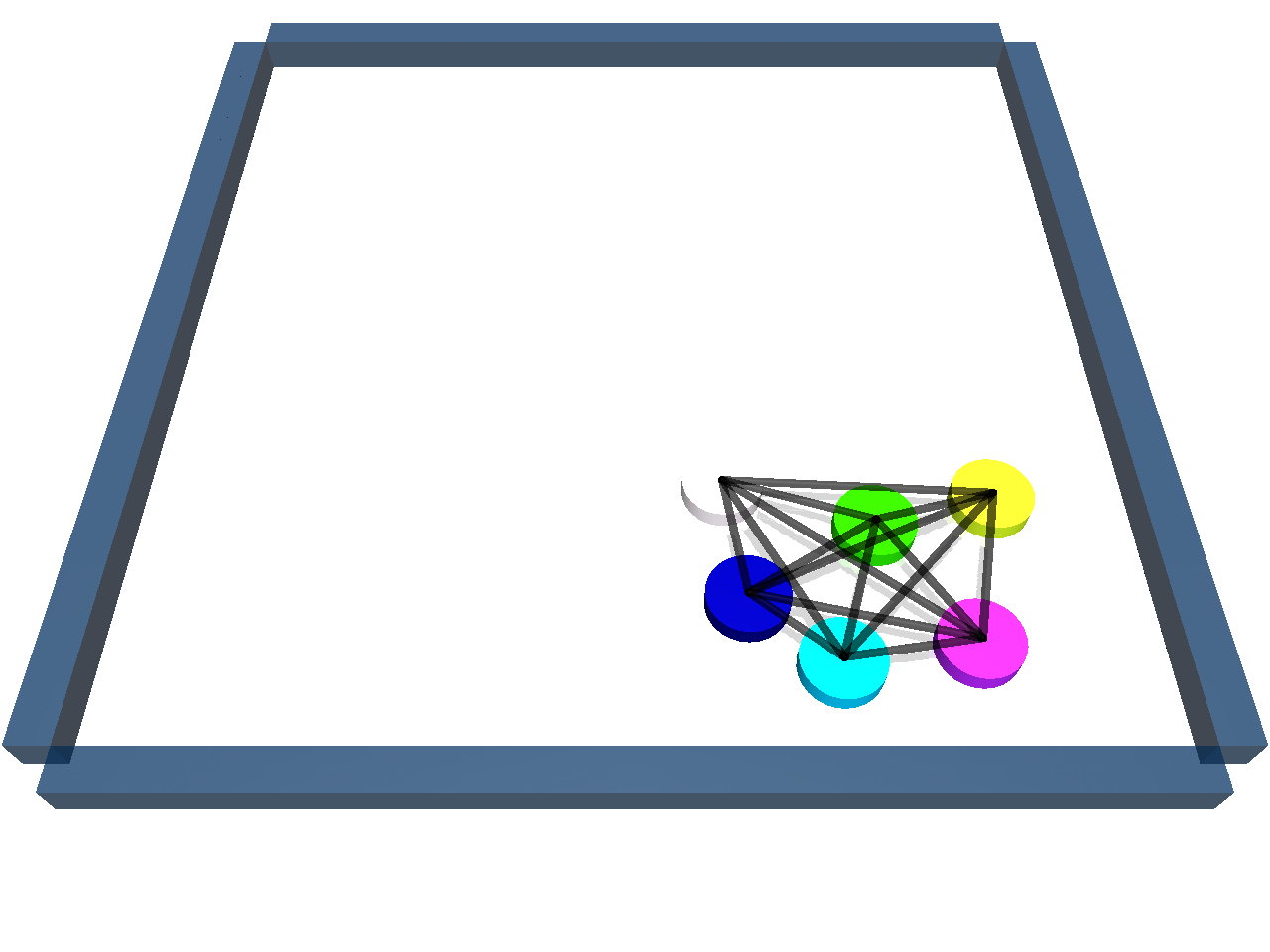}
    \end{subfigure}
    \begin{subfigure}{\subfigwidthrollouts}
        \includegraphics[width=\imagewidthrollouts]{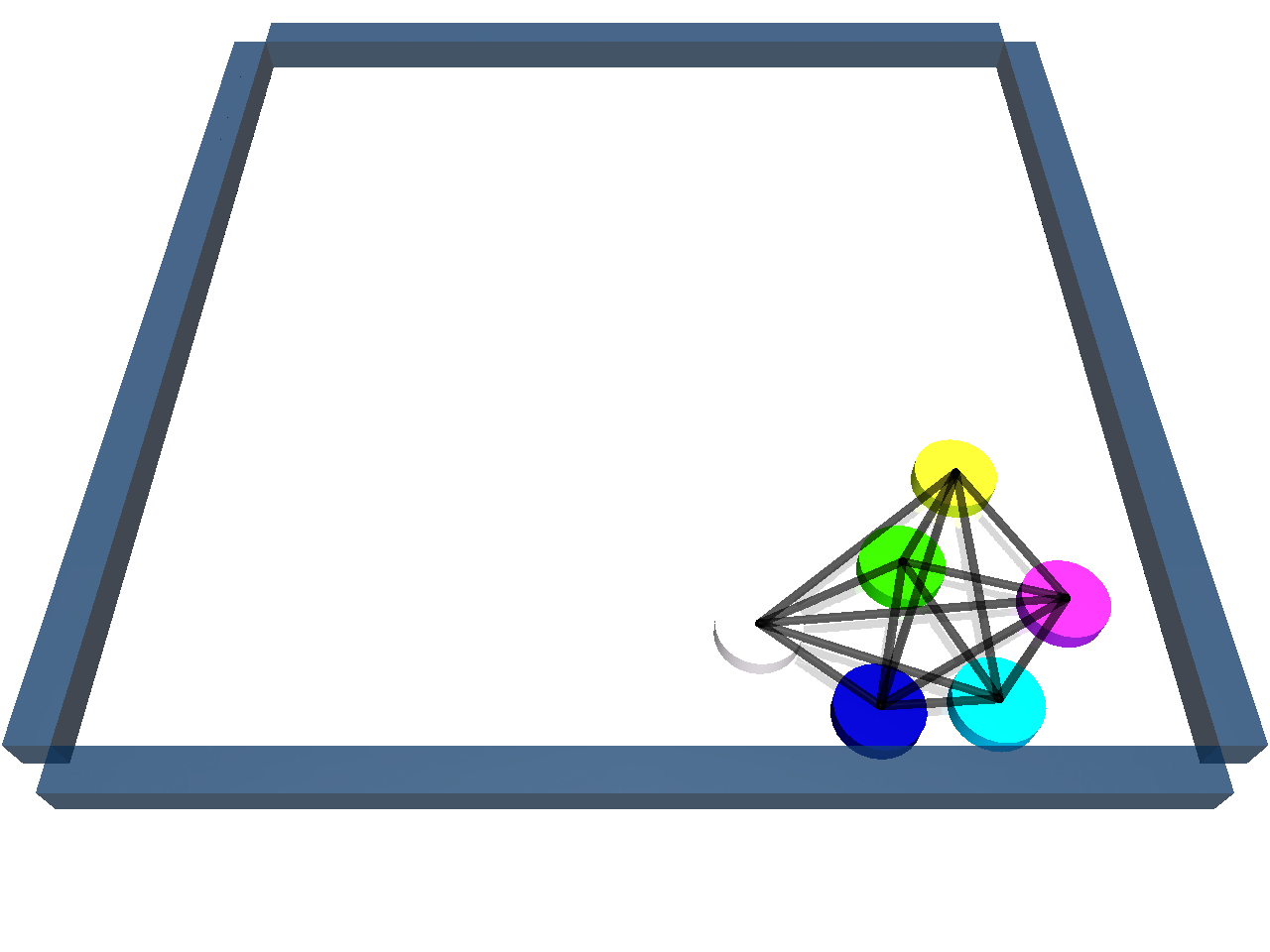}
    \end{subfigure}
    \begin{subfigure}{\subfigwidthrollouts}
        \includegraphics[width=\imagewidthrollouts]{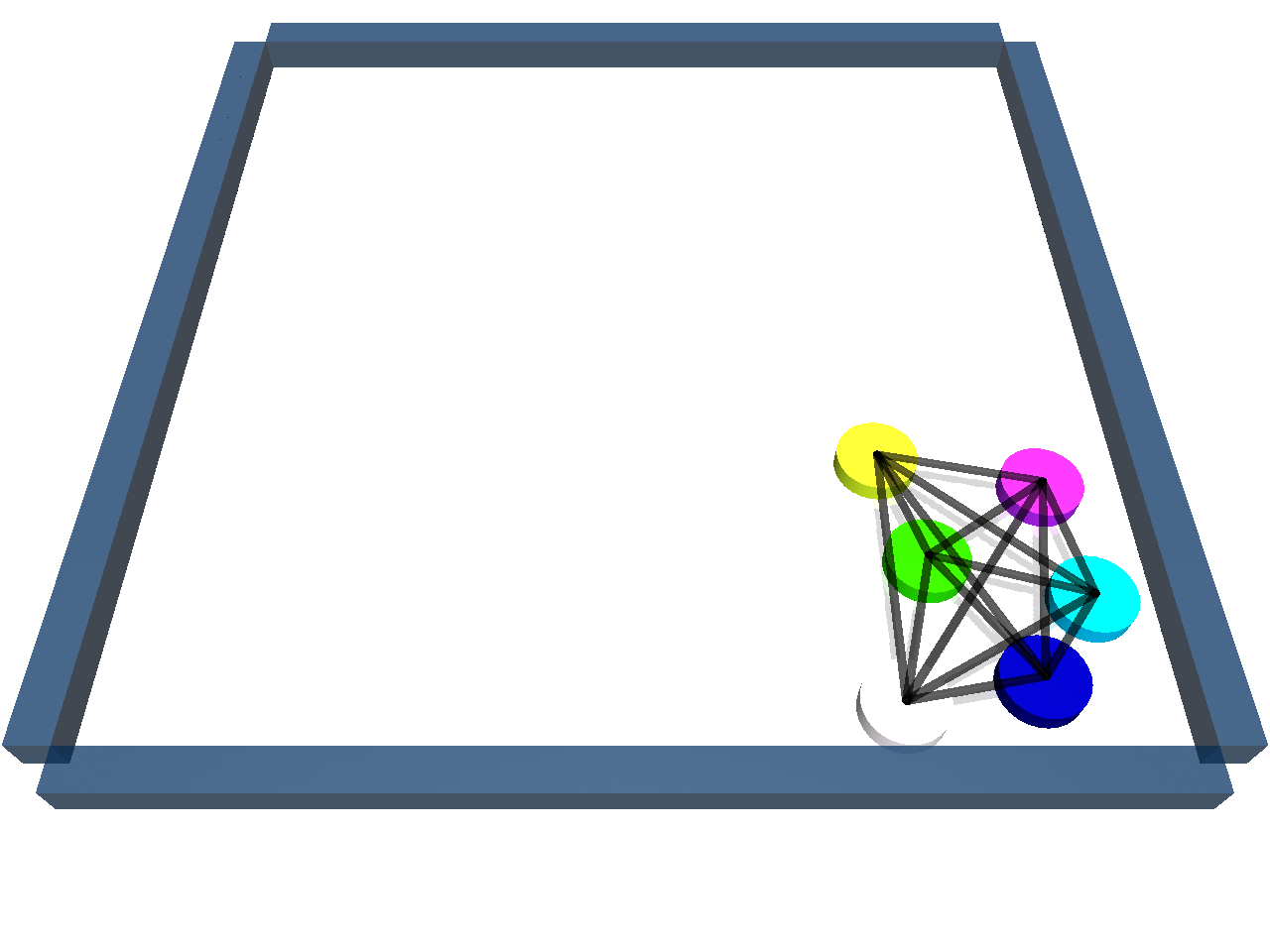}
    \end{subfigure}
    \begin{subfigure}{\subfigwidthrollouts}
       \includegraphics[width=\imagewidthrollouts]{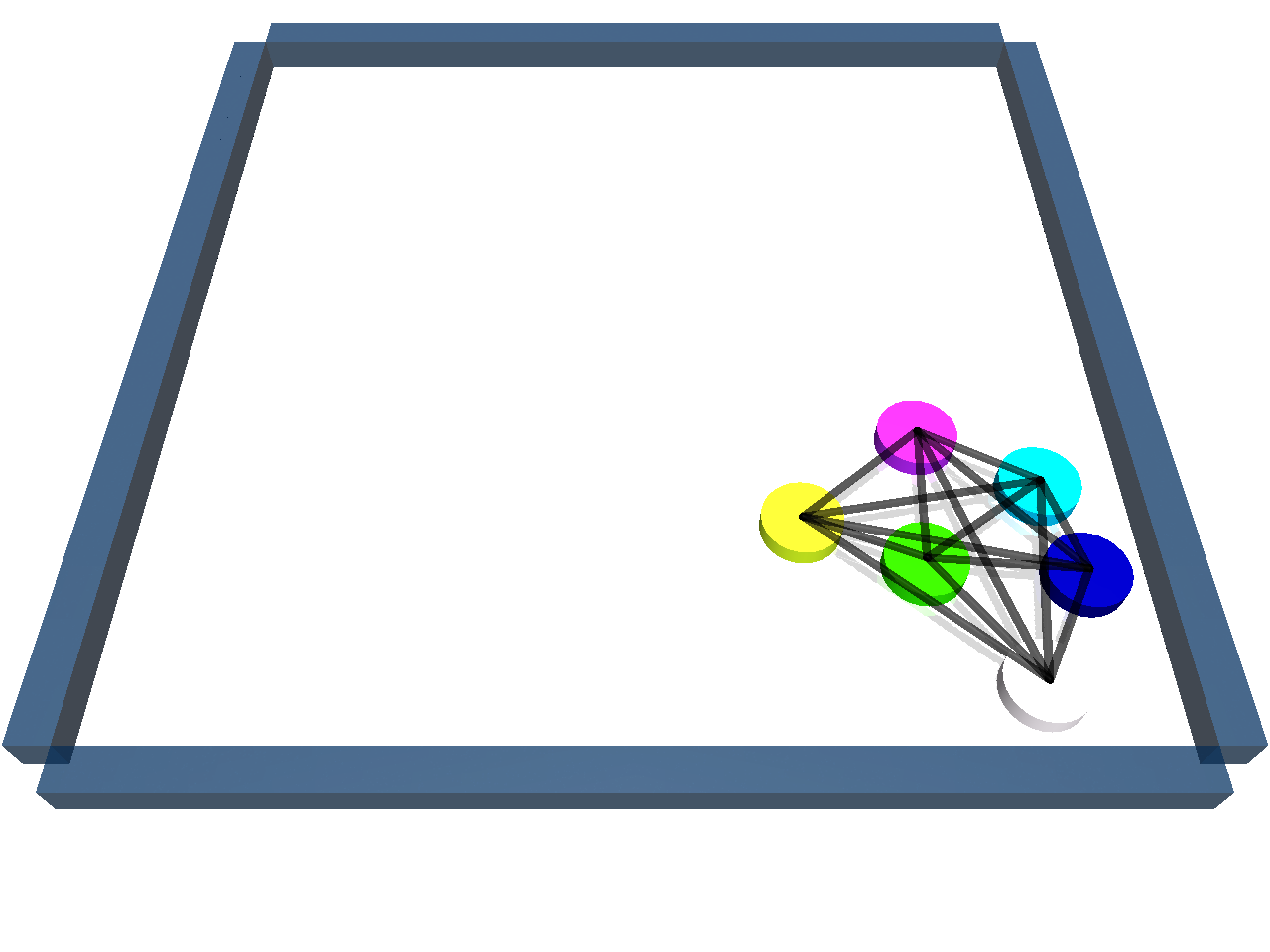}
    \end{subfigure}
    \begin{subfigure}{\subfigwidthrollouts}
        \includegraphics[width=\imagewidthrollouts]{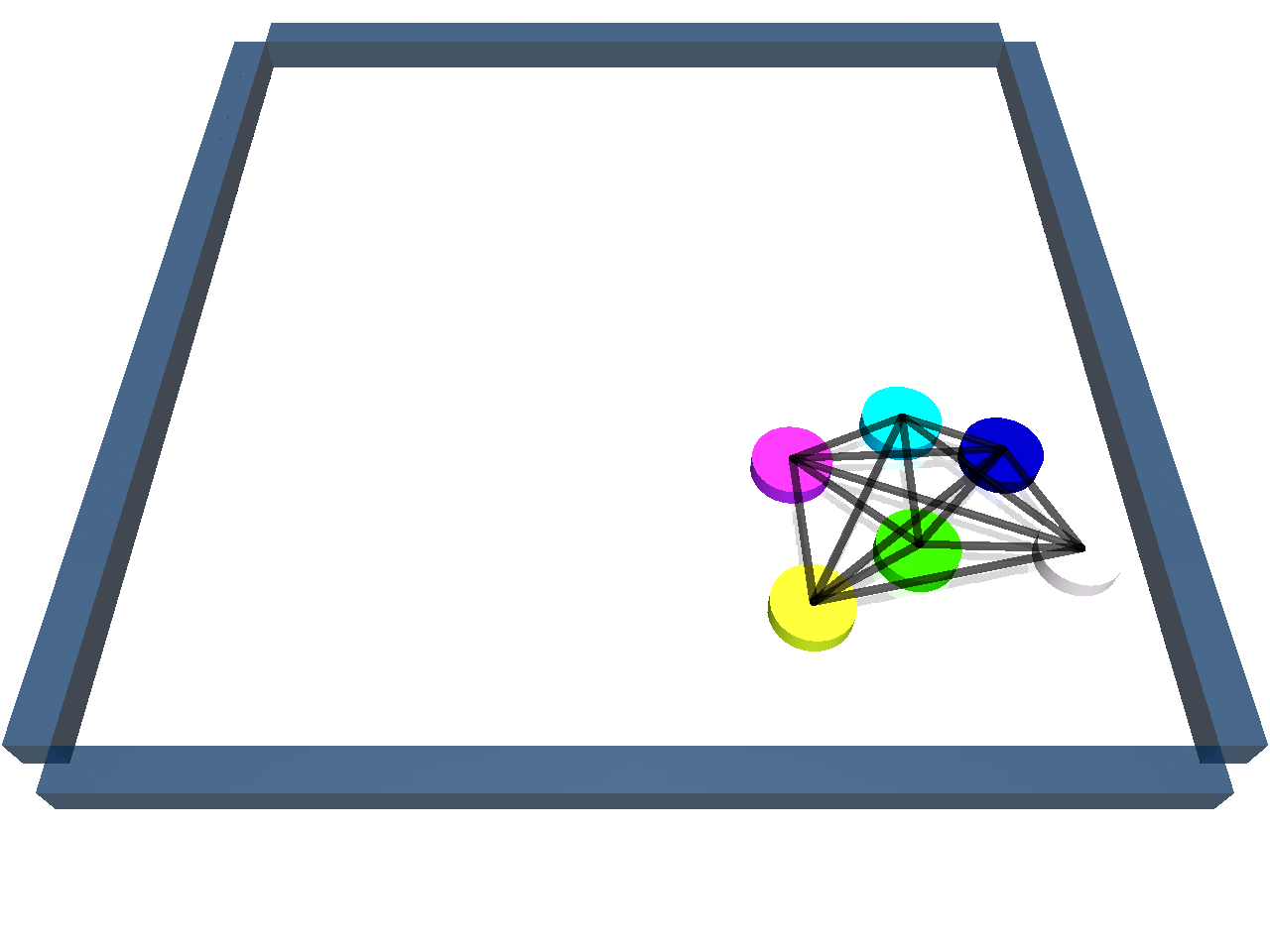}
    \end{subfigure}
    \hfill
    \begin{subfigure}{\rownamewidth}
        Bouncing\\Rigids\\(Rollout)
    \end{subfigure}
    \begin{subfigure}{\subfigwidthrollouts}
        \includegraphics[width=\imagewidthrollouts]{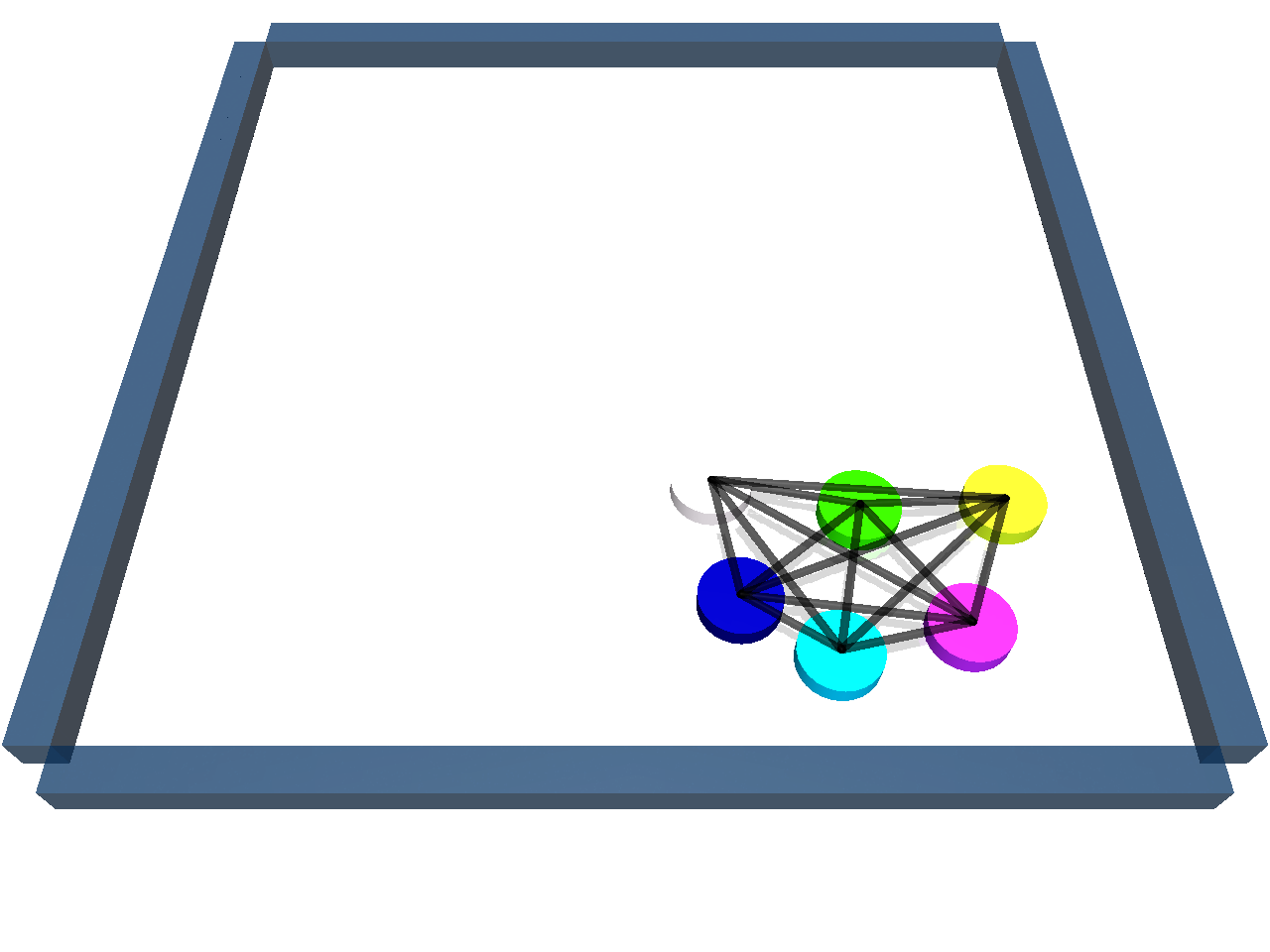}
    \end{subfigure}
    \begin{subfigure}{\subfigwidthrollouts}
        \includegraphics[width=\imagewidthrollouts]{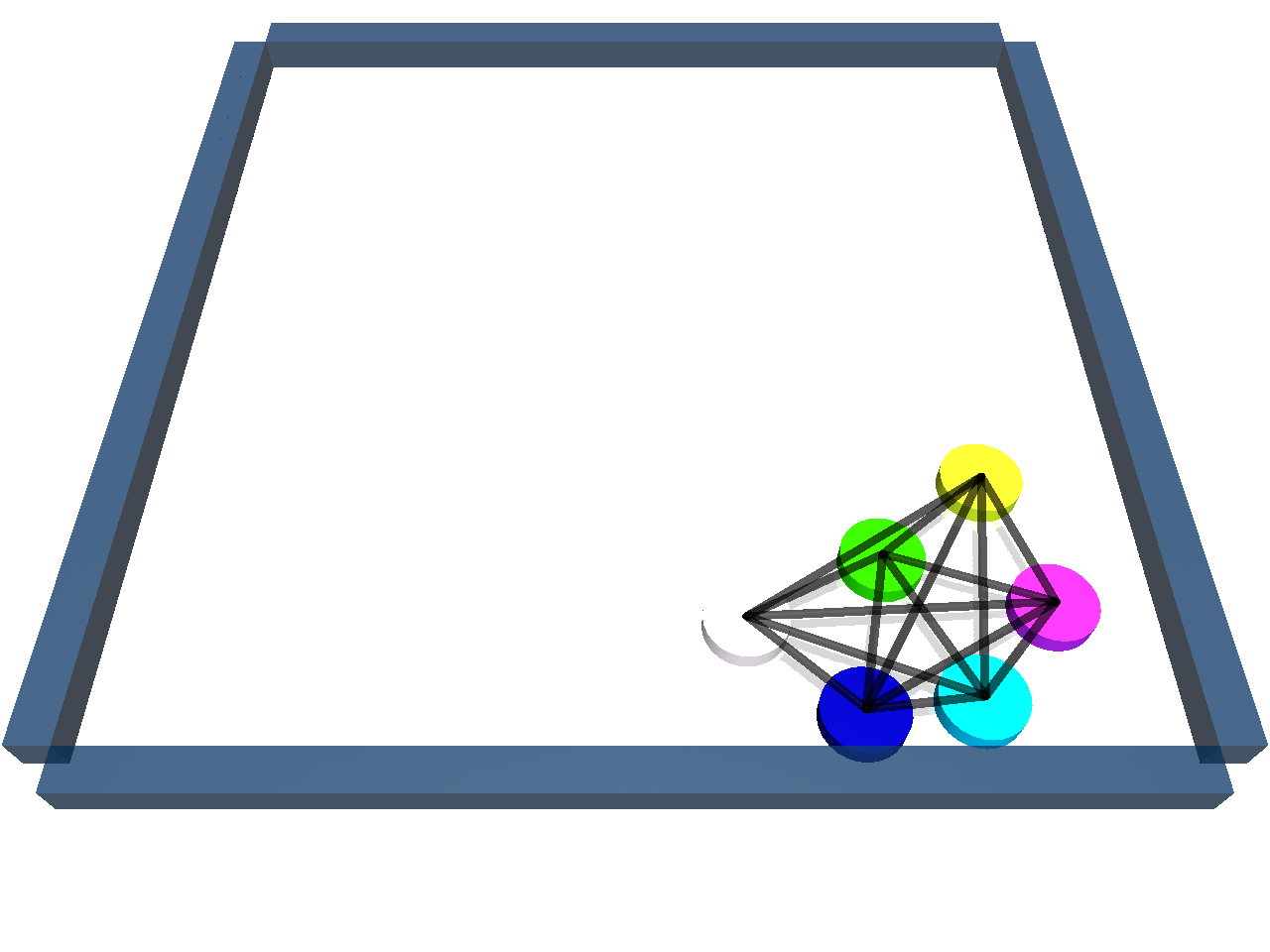}
    \end{subfigure}
    \begin{subfigure}{\subfigwidthrollouts}
        \includegraphics[width=\imagewidthrollouts]{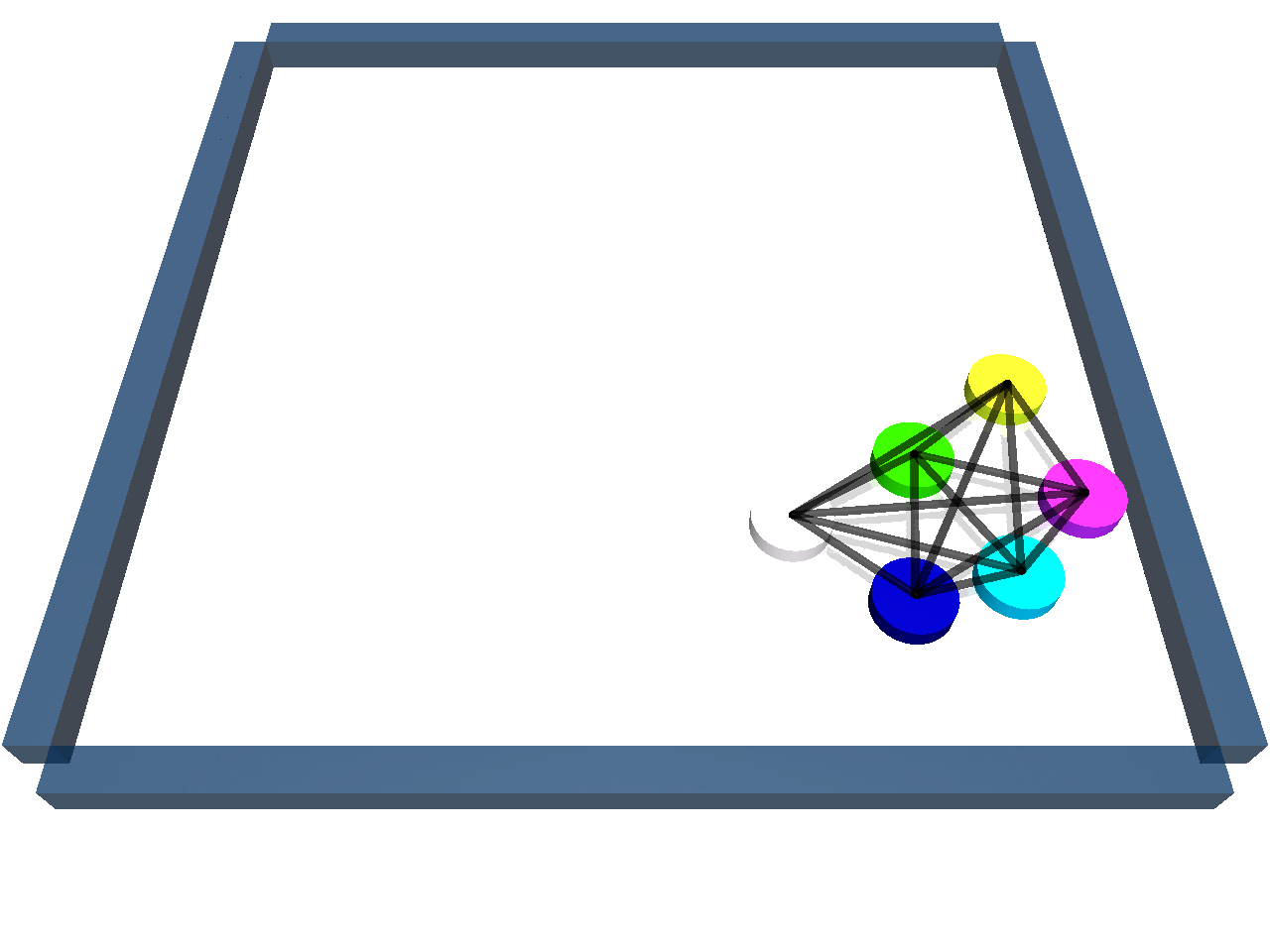}
    \end{subfigure}
    \begin{subfigure}{\subfigwidthrollouts}
       \includegraphics[width=\imagewidthrollouts]{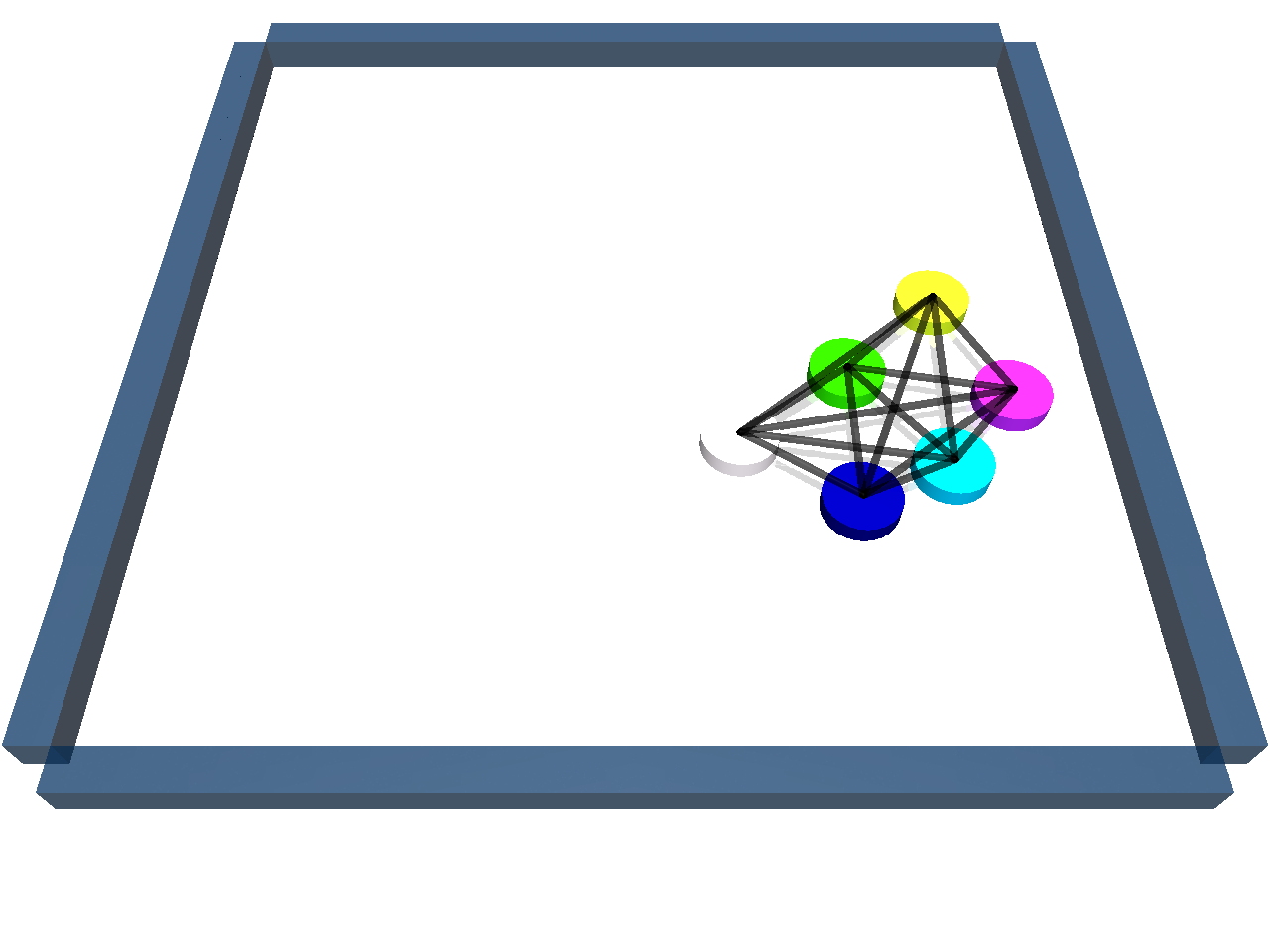}
    \end{subfigure}
    \begin{subfigure}{\subfigwidthrollouts}
        \includegraphics[width=\imagewidthrollouts]{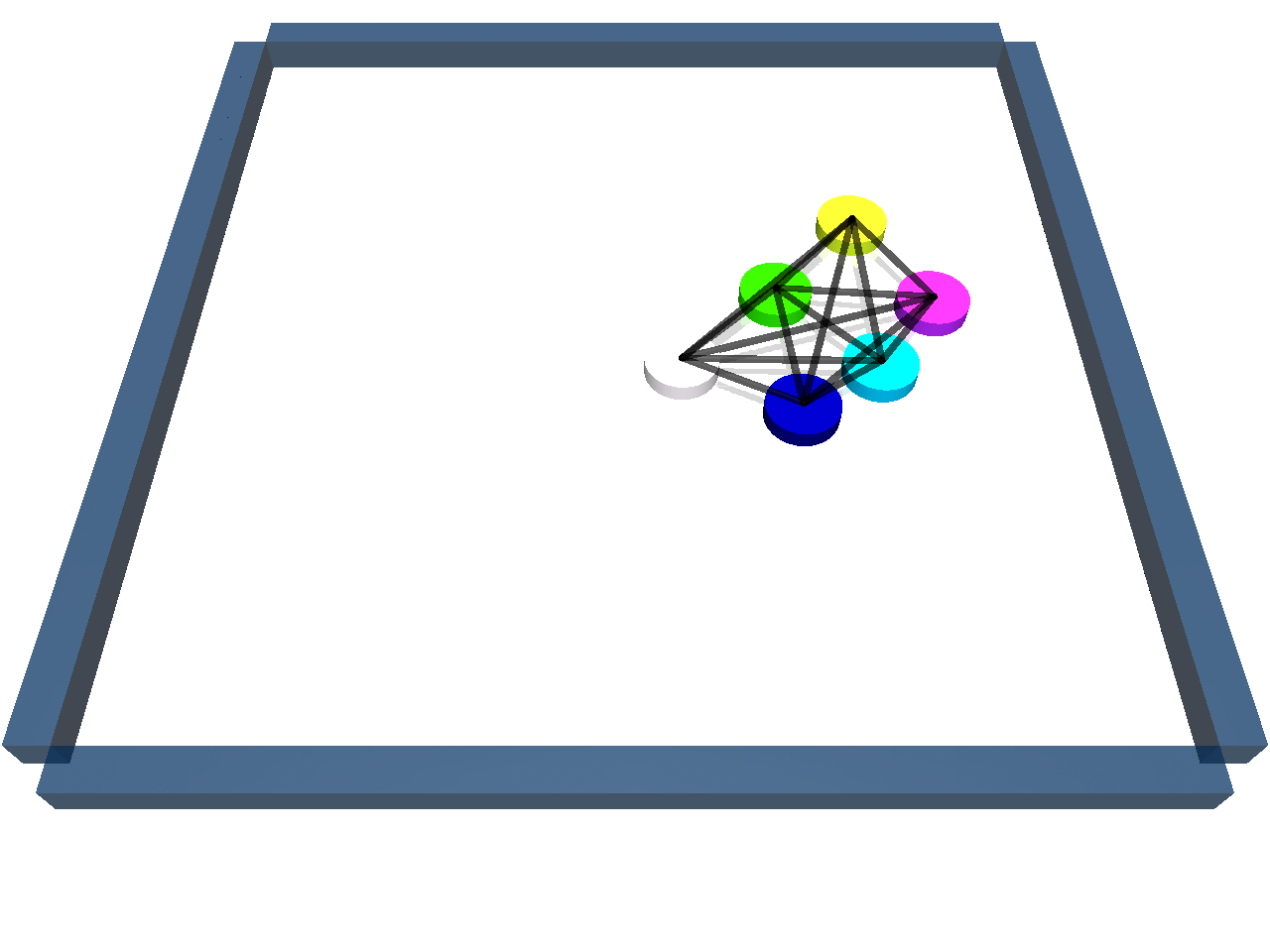}
    \end{subfigure}
    \hfill
    \begin{subfigure}{\rownamewidth}
    BoxBath\\(GT)
    \end{subfigure}
    \begin{subfigure}{\subfigwidthrollouts}
        \includegraphics[width=\imagewidthrollouts]{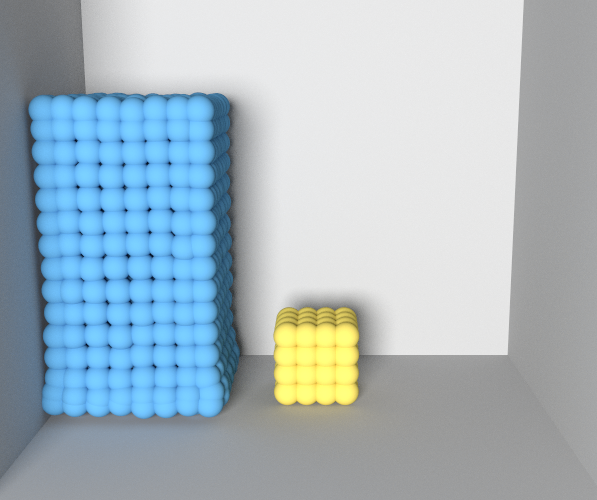}
    \end{subfigure}
    \begin{subfigure}{\subfigwidthrollouts}
        \includegraphics[width=\imagewidthrollouts]{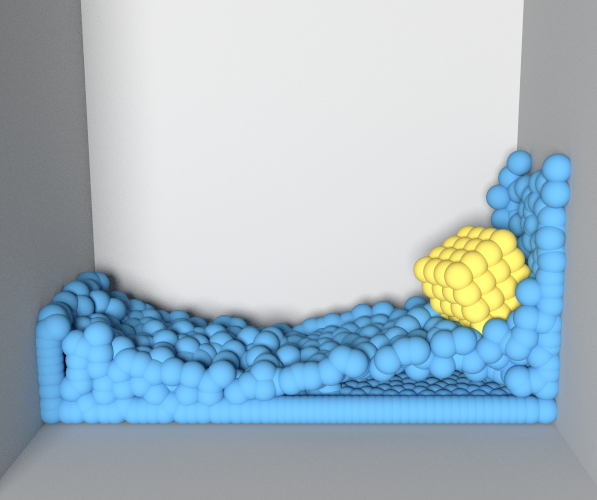}
    \end{subfigure}
    \begin{subfigure}{\subfigwidthrollouts}
        \includegraphics[width=\imagewidthrollouts]{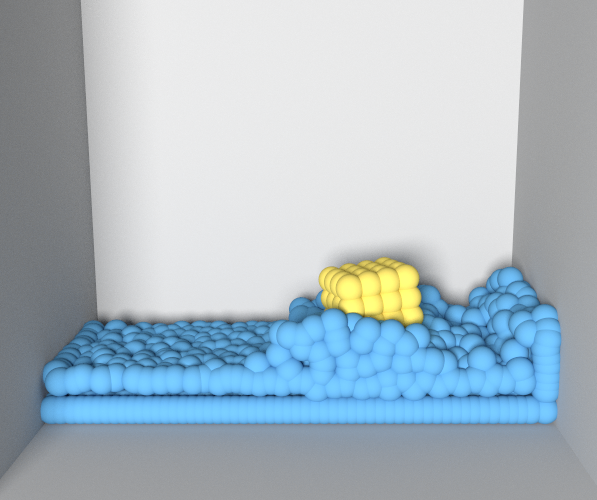}
    \end{subfigure}
    \begin{subfigure}{\subfigwidthrollouts}
        \includegraphics[width=\imagewidthrollouts]{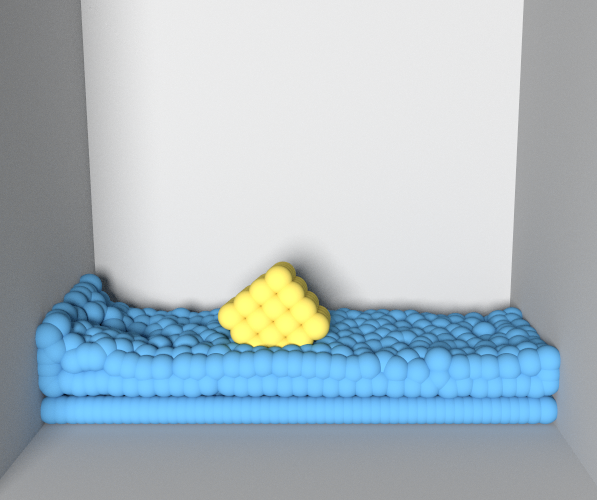}
    \end{subfigure}
    \begin{subfigure}{\subfigwidthrollouts}
        \includegraphics[width=\imagewidthrollouts]{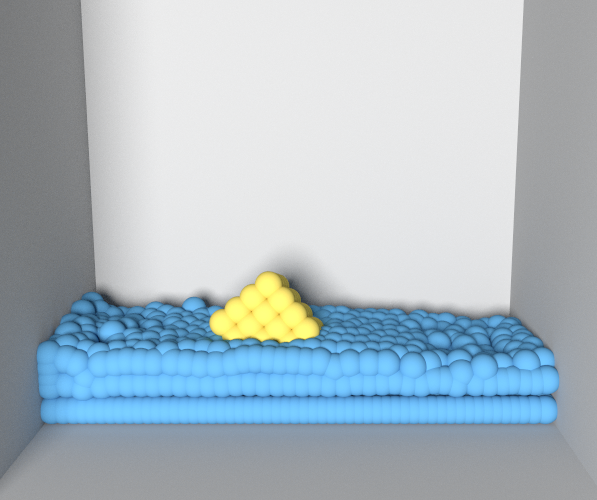}
    \end{subfigure}
    \hfill
    \begin{subfigure}{\rownamewidth}
    BoxBath\\(Rollout)
    \end{subfigure}
    \begin{subfigure}{\subfigwidthrollouts}
        \includegraphics[width=\imagewidthrollouts]{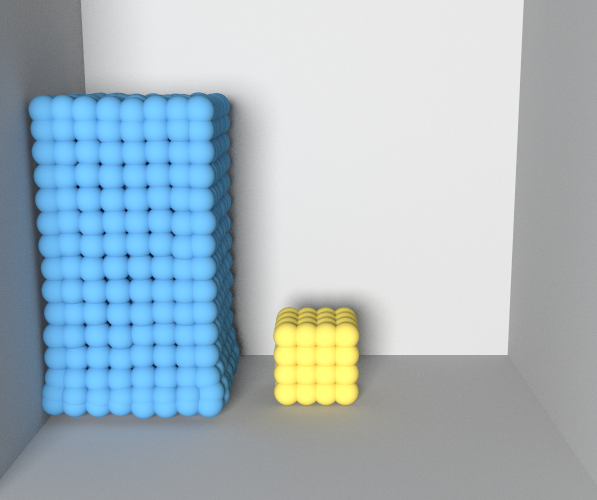}
    \end{subfigure}
    \begin{subfigure}{\subfigwidthrollouts}
        \includegraphics[width=\imagewidthrollouts]{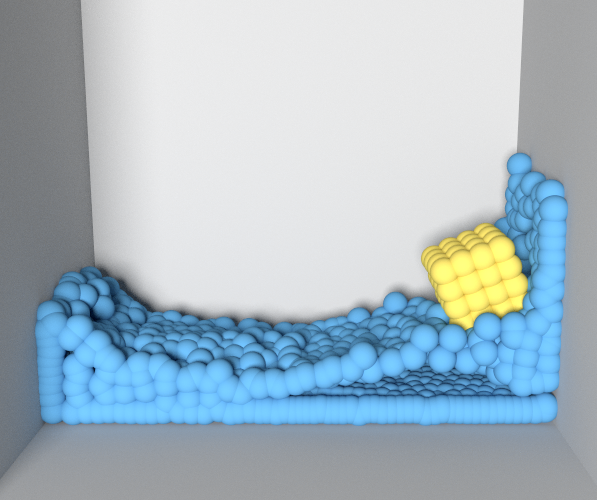}
    \end{subfigure}
    \begin{subfigure}{\subfigwidthrollouts}
        \includegraphics[width=\imagewidthrollouts]{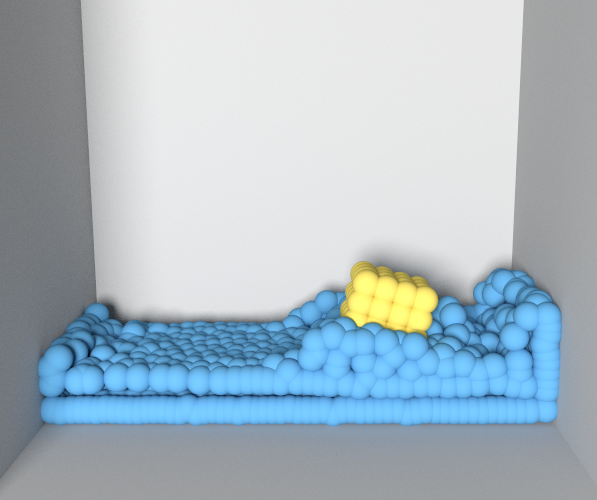}
    \end{subfigure}
    \begin{subfigure}{\subfigwidthrollouts}
        \includegraphics[width=\imagewidthrollouts]{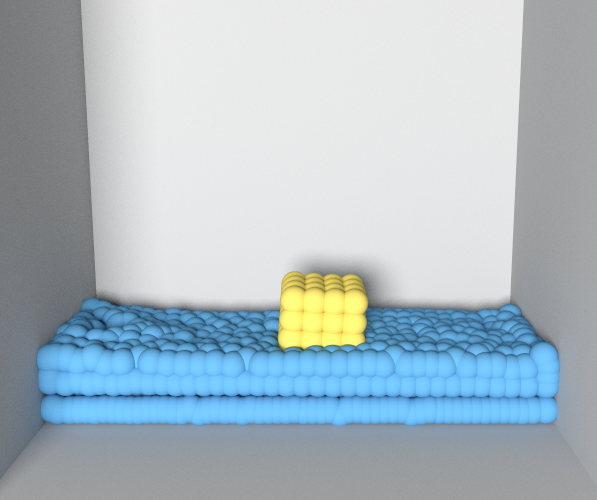}
    \end{subfigure}
    \begin{subfigure}{\subfigwidthrollouts}
        \includegraphics[width=\imagewidthrollouts]{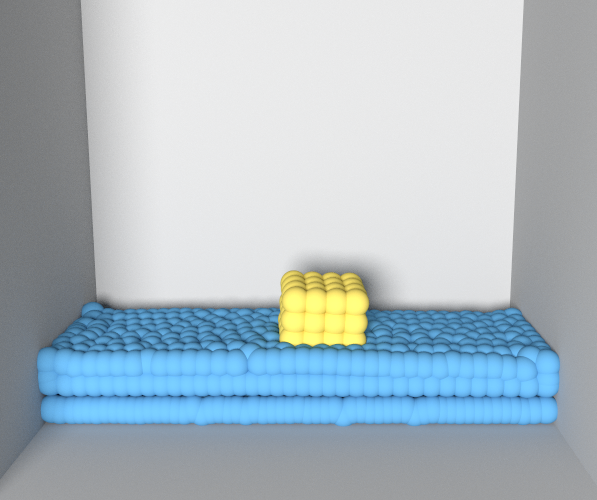}
    \end{subfigure}
\caption{Examples of the rollouts from \constraintmodel-GD models for our simulation environments.}
\label{fig:rollout_examples}
\end{figure}
\vfill

\pagebreak

\subsection{Investigating the convergence properties of the \constraintmodel-GD model}
\begin{figure}[H]
\centering
\begin{minipage}{0.7\textwidth}
\centering
\includegraphics[width=0.8\linewidth]{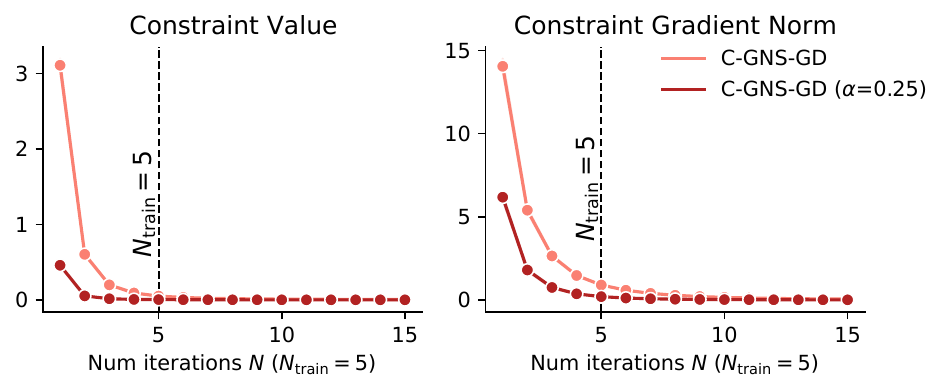}
\caption{\textbf{The constraint value and gradient norm across the number of solver iterations on the example from the \dataset{Rope} dataset}. The results are shown for two models: \constraintmodel-GD with the MSE loss on the last iteration only(pale red) and \constraintmodel-GD ($\alpha$=0.25) model with the additional decaying loss on intermediate states (dark red).}
\label{fig:constraint_convergence}
\end{minipage}
\end{figure}

In this section we investigate the convergence properties of the constraint function learned by \constraintmodel-GD on the examples from the \dataset{Rope} dataset, with and without the additional loss on intermediate states (see details in Section \ref{sec:suppl_model_implementation}). We train the models with 5 solver iterations and run the solver with up to 15 iterations at test time. Figure \ref{fig:constraint_convergence} demonstrates that in \constraintmodel-GD with both versions of the loss, the constraint value converges to the constant value, and the gradient norm converges to zero. The model with the loss on intermediate states with $\alpha=0.25$ converges in fewer iterations.

\subsection{Using other gradient-based solvers at test time}

\begin{figure}[H]
\centering
\begin{minipage}{0.8\textwidth}
\centering
\includegraphics[width=\linewidth]{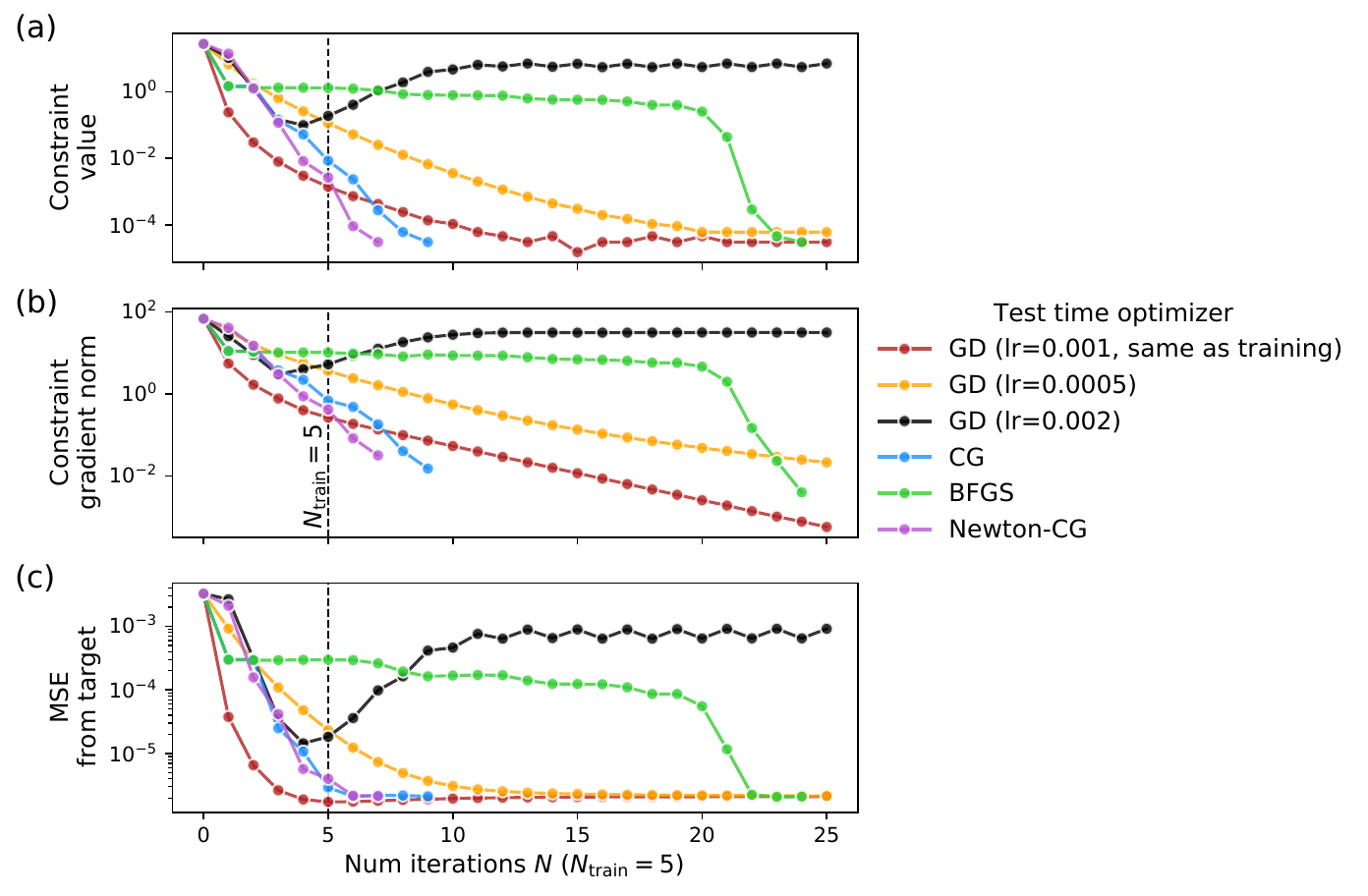}
\caption{\textbf{Generalization to different solvers at test time}. We run different solvers to minimize the constraint function learned by \constraintmodel-GD ($\alpha$=0.25) model with gradient descent solver with the~learning~rate~0.001. We start with the initial state from one of the examples in the \dataset{Rope} dataset. We ran the solvers until convergence or up to the maximum of 25 iterations. The figure show \textbf{(a)} the constraint value, shifted to have a minimum at zero \textbf{(b)}, constraint gradient norm, and \textbf{(c)} mean squared error between the state at the current iteration and the ground-truth. These plots are on log scale.}
\label{fig:rope_generalisation_optimizer}
\end{minipage}
\end{figure}

In Figure \ref{fig:rope_generalisation_optimizer} we further investigate whether the constraint function from the \constraintmodel-GD ($\alpha$=0.25) model can be optimized using a different gradient-based solver at test time. We used one of the test examples in the \dataset{Rope} dataset as the initial state for all the solvers we tested. We used solvers from SciPy \cite{virtanen2020scipy}, including Conjugate Gradient algorithm (CG, first order, \citet{NoceWrig06}), the Broyden–Fletcher–Goldfarb–Shanno algorithm (BFGS, second order method, \citet{NoceWrig06}) and the Newton Conjugate Gradient algorithm (Newton-CG, second order method, \citet{NoceWrig06}). We used default SciPy settings for these solvers. 

The model converges when using CG (blue), BFGS (green) and Newton-CG (purple) solvers, reaching a similar mean squared error as gradient descent (dark red), indicating that the learned constraint function is robust to the choice of optimization procedure. Note that Newton-CG solver also requires computing the Hessian of the constraint function, which we also obtained via auto-differentiation of the learned constraint. The convergence of Newton-CG suggests that the second order gradients of the learned constraint function are also well-behaved, even though they were never computed during training.

Next, we varied the learning rate (lr) of the gradient descent solver from the 0.001 value used during the training. As expected, halving the learning rate of gradient descent (lr=0.0005, orange) results in slower convergence, taking 15 iterations to reach a similar mean square error to GD(lr=0.001, dark red). On the other hand, with doubled learning rate (lr=0.002, black) the solver does not converge. We speculate that the model with lr=0.001 learned a constraint function that is sufficiently steep to converge to the minimum as quickly as possible with a learning rate of 0.001, and larger values of the learning rate may be detrimental. This is possible because the model is free to learn any scale for the constraint (and its gradients), which is equivalent to re-scaling the training learning rate. We informally tested this hypothesis by verifying that the model performance is not very sensitive to the training learning rate (although very large values make training more unstable). 

\newpage
\begin{figure}[H]
\centering
\begin{minipage}{0.8\textwidth}
\centering
\includegraphics[width=0.8\linewidth,trim={0 0 40, 0}, clip]{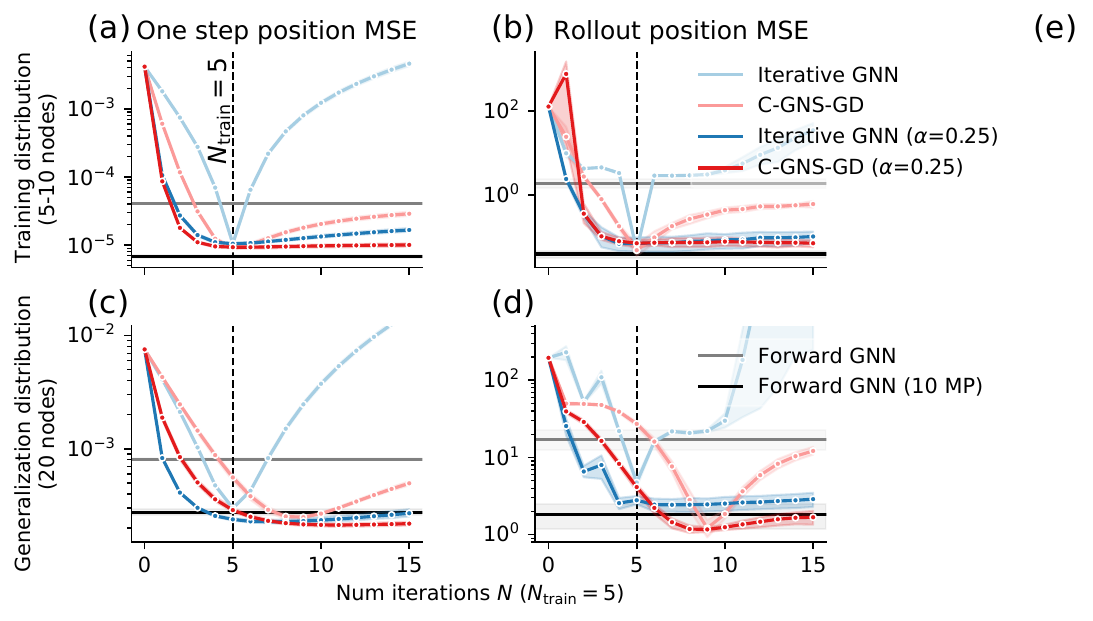}
\caption{\textbf{Generalization to more solver iterations and larger \dataset{Rope} systems at test time.} \textbf{(a-b)}~Test rollout MSE for ropes with the same lengths as those during training (5-10 nodes) \textbf{(c-d)}~Test rollout MSE for larger ropes (20 nodes) than during training. The x-axis indicate the number of solver iterations used at test time. The model was trained with 5 solver iterations. The y-axis represents MSE values. The horizontal black and grey lines show the performance of the Forward GNN models, which do not have the option to vary the number of iterations at test time. The results are shown for the models with the standard MSE loss on the last iteration (pale red and blue), as well as the models with exponentially decaying loss over all iterations (bright red and blue, see details in Section \ref{sec:suppl_model_implementation}).
}
\label{fig:suppl_rope_generalisation}
\end{minipage}
\end{figure}

\newpage
\subsection{Ablations to existing baselines}
\begin{figure}[H]
\centering
\includegraphics[width=\linewidth]{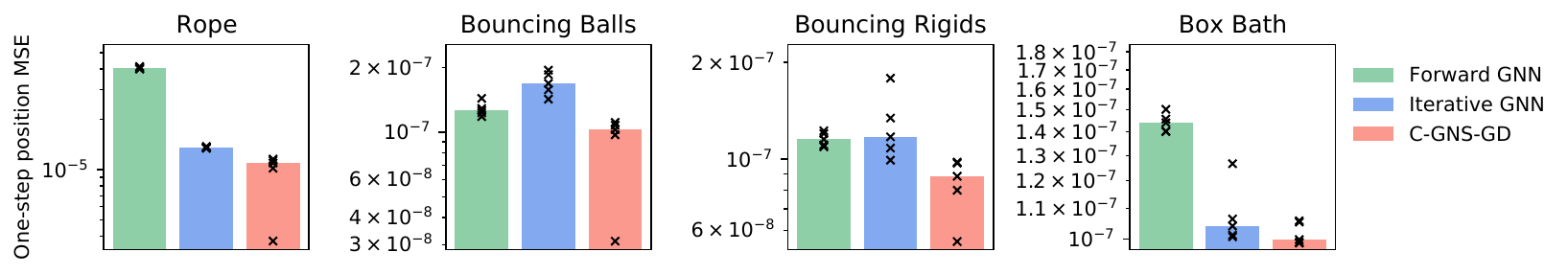}
\includegraphics[width=\linewidth]{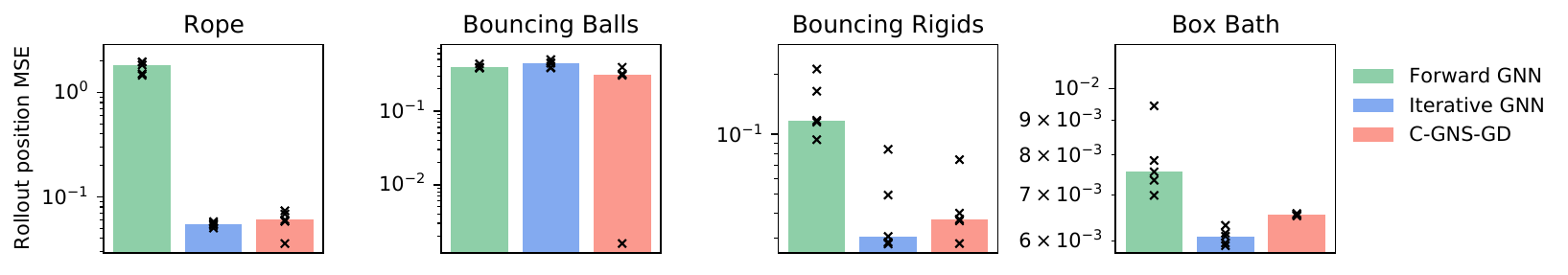}
\caption{Comparison to \forwardbaselinename and \iterativemodel. Top row: one-step test MSE error on node positions. Bottom row: full 160-step rollout MSE. The bar height represents the median MSEs over random seeds. The black cross marks show the MSE metric for each random seed. The black arrows indicates that a random seeds exceeds the upper y limit of the figure.}
\label{fig:model_ablations_forward_gnn}
\end{figure}

Figure \ref{fig:model_ablations_forward_gnn}
demonstrates the comparison of \constraintmodel-GD to \forwardbaselinename and \iterativemodel. Adding iterative refinement of the state, but computing the update directly (\iterativemodel) improves 1-step error on \dataset{Rope} and \dataset{BoxBath}, but suffers from higher variance across seeds on \dataset{Bouncing Balls} and \dataset{Bouncing Rigids}. Computing the update via constraint gradient (\constraintmodel-GD, ours) further improves 1-step error and improves the model stability. Both \iterativemodel and \constraintmodel-GD outperform \forwardbaselinename on the full rollout.
\begin{figure}[H]
\centering
\includegraphics[width=0.95\linewidth]{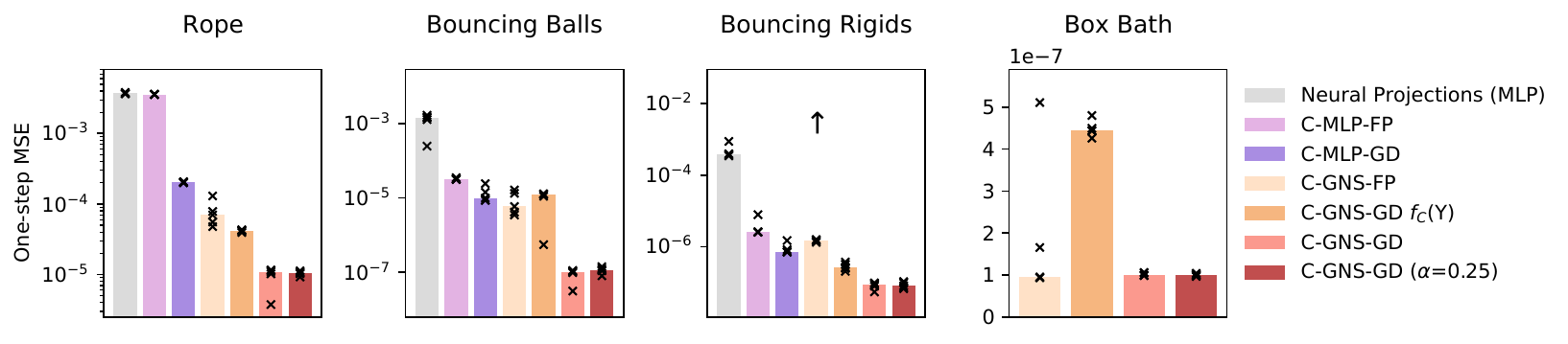}
\caption{\textbf{Ablations to Neural Projections}. One-step test MSE error on node positions. See Figure \ref{fig:model_ablations_yang} for full rollout MSE. The bar height represents the median MSEs over random seeds. The black cross marks show the MSE metric for each random seed. The black arrows indicates that a random seeds exceeds the upper y limit of the figure. The~upper~y~is set to 1e5~$\times$~the~median~MSE of \constraintmodel-GD.}
\label{fig:model_ablations_yang_suppl}
\end{figure}

\vfill
\clearpage

\begin{table}[H]
\begin{center}
\caption{\textbf{Median performance of the models on different datasets}.  The standard deviation from the median is shown over 5 random seeds. We do not show the results for the models where the median value, or standard deviation is more than 1000 times larger than the best model in each column. Note that the tables use different scales to demonstrate the errors on 1-step error, 10-step rollouts and full rollouts. Results are not shown for MLP models on \dataset{BoxBath}. We omit the results for the models where the median error is more than 4 orders of magnitude larger  than the median error of the \constraintmodel-GD model.}
\label{tab:suppl_tables}
\small
\begin{tabular}{l>{\centering\arraybackslash}p{27mm}>{\centering\arraybackslash}p{27mm}>{\centering\arraybackslash}p{25mm}>{\centering\arraybackslash}p{23mm}}
\hline & \\[-1.5ex]

\textbf{Model} & \multicolumn{4}{c}{\bf{One-step position MSE}} \\ \hline & \\[-1.5ex] & Rope (1e-5) & Bouncing Balls (1e-6) & Bouncing Rigids (1e-7) & Box Bath (1e-7) \\
\hline
Neural Projections & $ 370.525 \pm 6.120 $ & $ 1434.830 \pm 547.491 $ & \multicolumn{1}{c}{--} & \multicolumn{1}{c}{--} \\
ConstraintMLP-FP & $ 358.292 \pm 0.532 $ & $ 31.984 \pm 1.617 $ & $ 26.009 \pm 23.440 $ & \multicolumn{1}{c}{--} \\
ConstraintMLP-GD & $ 20.422 \pm 0.271 $ & $ 9.965 \pm 6.745 $ & $ 7.313 \pm 3.443 $ & \multicolumn{1}{c}{--} \\
\constraintmodel-FP & $ 6.983 \pm 2.965 $ & $ 5.896 \pm 5.930 $ & \multicolumn{1}{c}{--} & $ 0.954 \pm 1.885 $ \\
\constraintmodel-GD $f_C(Y)$ & $ 4.198 \pm 0.137 $ & $ 12.241 \pm 5.264 $ & $ 2.697 \pm 0.613 $ & $ 4.443 \pm 0.182 $ \\ \hline
\forwardbaselinename & $ 4.062 \pm 0.050 $ & $ 0.126 \pm 0.009 $ & $ 1.152 \pm 0.051 $ & $ 1.440 \pm 0.038 $ \\
\iterativemodel & $ 1.355 \pm 0.012 $ & $ 0.168 \pm 0.019 $ & $ 1.172 \pm 0.298 $ & $ 1.042 \pm 0.103 $ \\ \hline
\textbf{\constraintmodel-GD} & $ \mathbf{1.097 \pm 0.326} $ & $ \mathbf{0.103 \pm 0.032} $ & $ \mathbf{0.884 \pm 0.163} $ & $ \mathbf{0.998 \pm 0.038} $ \\
\constraintmodel-GD ($\alpha$=0.25) & $\mathbf{1.054 \pm 0.070} $ & $\mathbf{0.113 \pm 0.020} $ & $ \mathbf{0.819 \pm 0.154} $ & $ \mathbf{1.000 \pm 0.026} $ \\
\hline
\end{tabular}
\newline
\vspace*{0.5cm}
\newline
\begin{tabular}{l>{\centering\arraybackslash}p{27mm}>{\centering\arraybackslash}p{27mm}>{\centering\arraybackslash}p{25mm}>{\centering\arraybackslash}p{23mm}}
\hline & \\[-1.5ex]

\textbf{Model} & \multicolumn{4}{c}{\bf{Rollout position MSE
(10 steps)}} \\ \hline & \\[-1.5ex] & Rope (1e-4) & Bouncing Balls (1e-5) & Bouncing Rigids (1e-5) & Box Bath (1e-5) \\
\hline
Neural Projections & $ 9798.580 \pm 176.505 $ & \multicolumn{1}{c}{--} & \multicolumn{1}{c}{--} & \multicolumn{1}{c}{--} \\
ConstraintMLP-FP & $ 9558.814 \pm 9.186 $ & $ 704.608 \pm 16.157 $ & $ 48.944 \pm 392.478 $ & \multicolumn{1}{c}{--} \\
ConstraintMLP-GD & $ 41.000 \pm 5.062 $ & $ 756.406 \pm 177.790 $ & $ 6.933 \pm 3.332 $ & \multicolumn{1}{c}{--} \\
\constraintmodel-FP & $ 44.577 \pm 23.233 $ & $ 661.722 \pm 372.490 $ & \multicolumn{1}{c}{--} & $ 0.200 \pm 0.321 $ \\
\constraintmodel-GD $f_C(Y)$ & $ 18.399 \pm 1.090 $ & $ 615.769 \pm 298.710 $ & $ 0.927 \pm 0.279 $ & $ 43.099 \pm 14.758 $ \\ \hline
\forwardbaselinename & $ 275.851 \pm 73.352 $ & $ 2.853 \pm 0.277 $ & $ 12.168 \pm 3.459 $ & $ 0.386 \pm 0.061 $ \\
\iterativemodel & $ 2.506 \pm 1.454 $ & $ 2.260 \pm 0.429 $ & $ \mathbf{2.380 \pm 0.257} $ & $ \mathbf{0.174 \pm 0.020} $ \\ \hline
\textbf{\constraintmodel-GD} & $ 2.222 \pm 0.590 $ & $ \mathbf{0.613 \pm 0.333} $ & $ \mathbf{2.482 \pm 0.384} $ & $ 0.288 \pm 0.055 $ \\
\constraintmodel-GD ($\alpha$=0.25) & $ \mathbf{1.770 \pm 0.443} $ & $ 0.836 \pm 0.402 $ & $ \mathbf{2.108 \pm 0.943} $ & $ 0.240 \pm 0.036 $ \\
\hline
\end{tabular}
\newline
\vspace*{0.5cm}
\newline
\begin{tabular}{l>{\centering\arraybackslash}p{27mm}>{\centering\arraybackslash}p{27mm}>{\centering\arraybackslash}p{25mm}>{\centering\arraybackslash}p{23mm}}
\hline & \\[-1.5ex]

\textbf{Model} & \multicolumn{4}{c}{\bf{Rollout position MSE}} \\ \hline & \\[-1.5ex] & Rope (1e-1) & Bouncing Balls  & Bouncing Rigids (1e-1) & Box Bath (1e-2) \\
\hline
Neural Projections & \multicolumn{1}{c}{--} & $ 18.881 \pm 6.131 $ & \multicolumn{1}{c}{--} & \multicolumn{1}{c}{--} \\
ConstraintMLP-FP & \multicolumn{1}{c}{--} & $ 9.777 \pm 5.367 $ & $ 21.828 \pm 109.082 $ & \multicolumn{1}{c}{--} \\
ConstraintMLP-GD & \multicolumn{1}{c}{--} & $ 2.425 \pm 0.190 $ & $ 3.332 \pm 2.164 $ & \multicolumn{1}{c}{--} \\
\constraintmodel-FP & $ 2.804 \pm 3.432 $ & \multicolumn{1}{c}{--} & \multicolumn{1}{c}{--} & $ 0.689 \pm 0.099 $ \\
\constraintmodel-GD $f_C(Y)$ & $ 3.427 \pm 0.400 $ & $ 1.221 \pm 0.350 $ & $ 0.818 \pm 0.229 $ & $ 2.596 \pm 1.189 $ \\ \hline
\forwardbaselinename & $ 18.305 \pm 2.340 $ & $ 0.389 \pm 0.022 $ & $ 1.174 \pm 0.486 $ & $ 0.756 \pm 0.089 $ \\
\iterativemodel & $ 0.546 \pm 0.026 $ & $ 0.445 \pm 0.044 $ & $ \mathbf{0.306 \pm 0.253} $ & $ \mathbf{0.609 \pm 0.015} $ \\ \hline
\textbf{\constraintmodel-GD} & $ 0.602 \pm 0.131 $ & $ \mathbf{0.308 \pm 0.142} $ & $ 0.374 \pm 0.171 $ & $ 0.654 \pm 0.002 $ \\
\constraintmodel-GD ($\alpha$=0.25) & $ \mathbf{0.460 \pm 0.053} $ & $ \mathbf{0.335 \pm 0.037} $ & $ 0.429 \pm 0.249 $ & $ 0.660 \pm 0.011 $ \\
\hline
\end{tabular}
\end{center}
\end{table}

\newcommand{\subfigwidth}{0.21\linewidth}
\newcommand{\imagewidth}{0.99\linewidth}
\newcommand{\figspace}{0.03\linewidth}

\begin{figure}[H]
\centering
\captionsetup{width=.7\linewidth}
    \begin{subfigure}[t]{\subfigwidth}
        \includegraphics[width=\imagewidth]{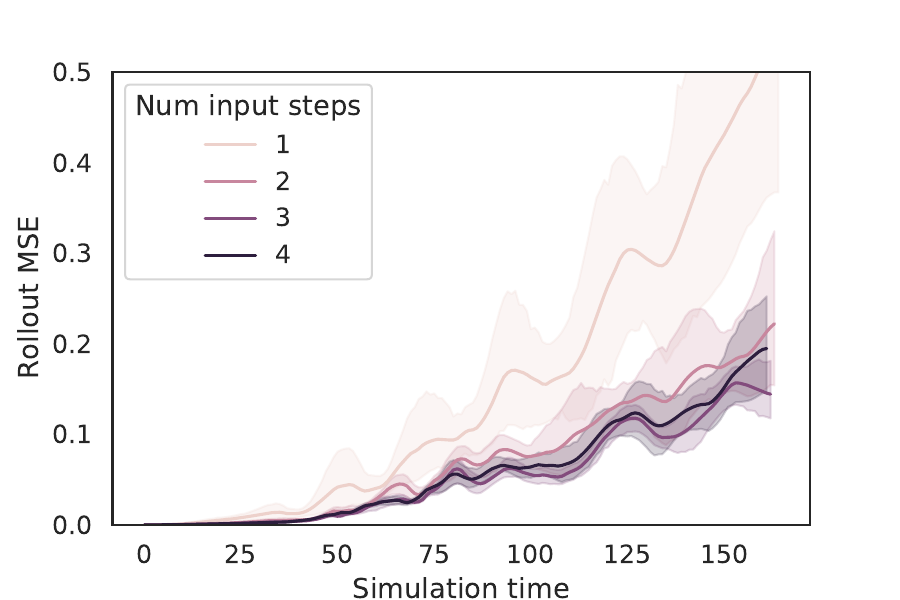}
        \caption{\# past time point in the history: position MSE}
    \end{subfigure} \hspace{\figspace}
    \begin{subfigure}[t]{\subfigwidth}
        \includegraphics[width=\imagewidth]{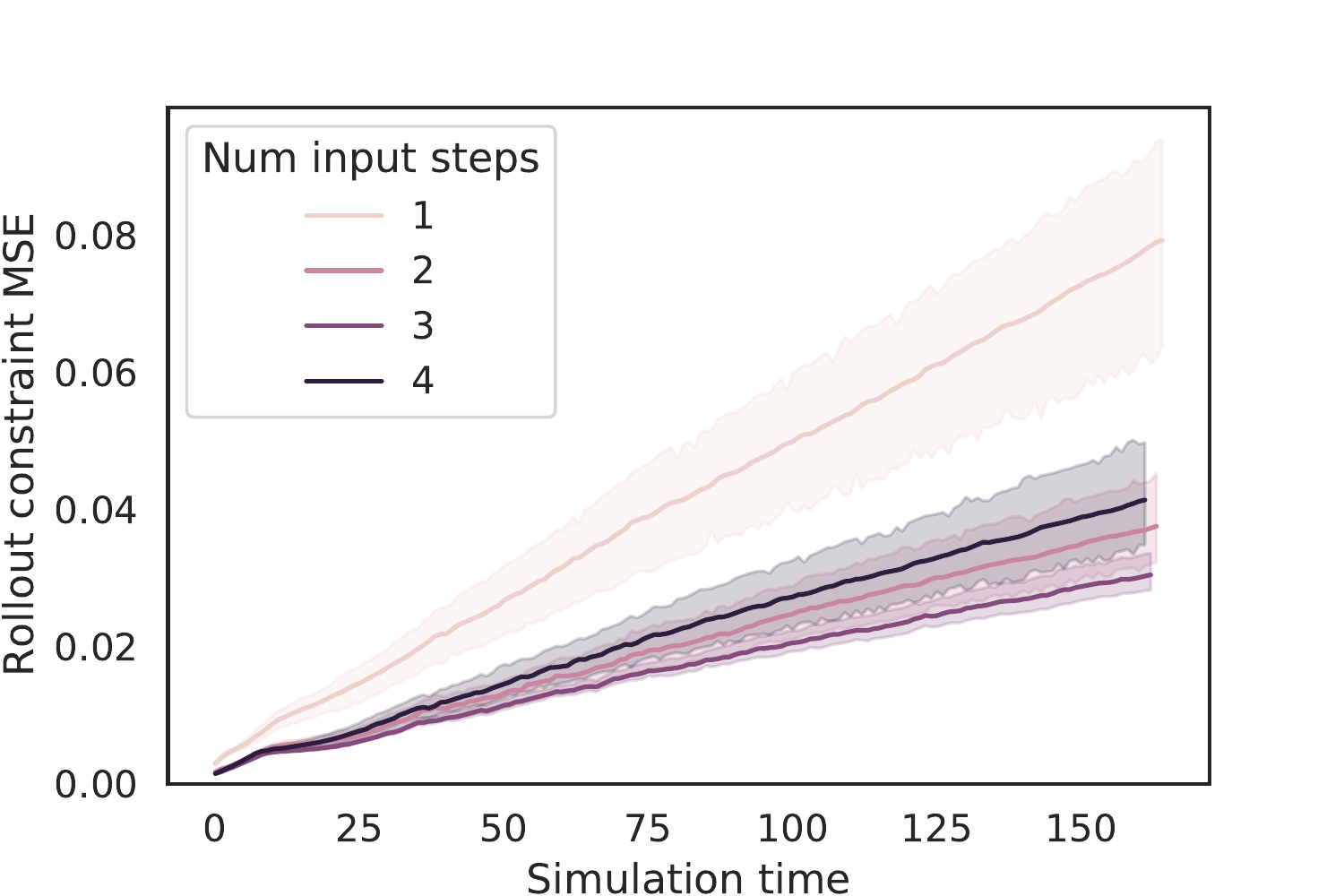}
        \caption{\# past time point in the history: constraint MSE}
    \end{subfigure} \hspace{\figspace}
    \begin{subfigure}[t]{\subfigwidth}
        \includegraphics[width=\imagewidth]{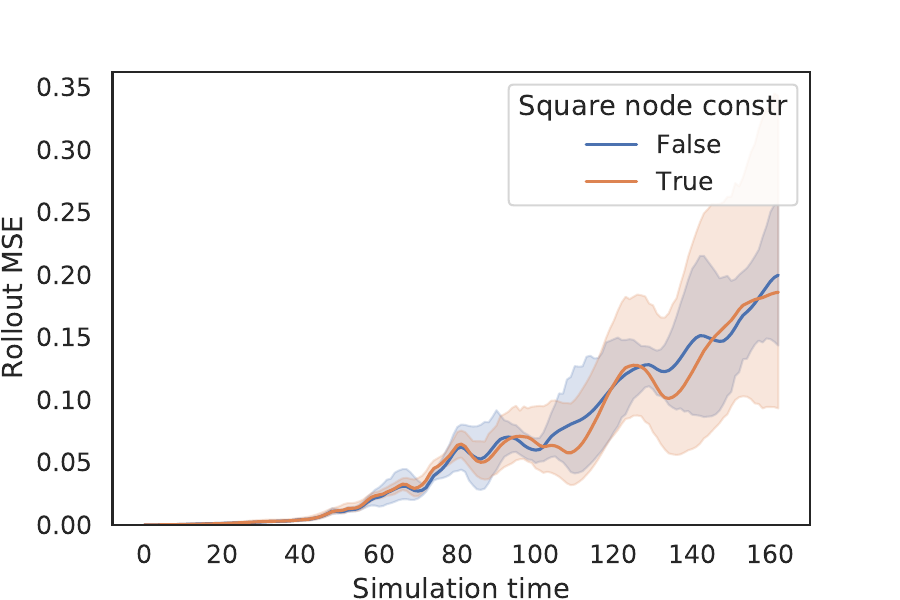}
        \caption{Squaring per-node outputs: position MSE}
    \end{subfigure} \hspace{\figspace}
    \begin{subfigure}[t]{\subfigwidth}
        \includegraphics[width=\imagewidth]{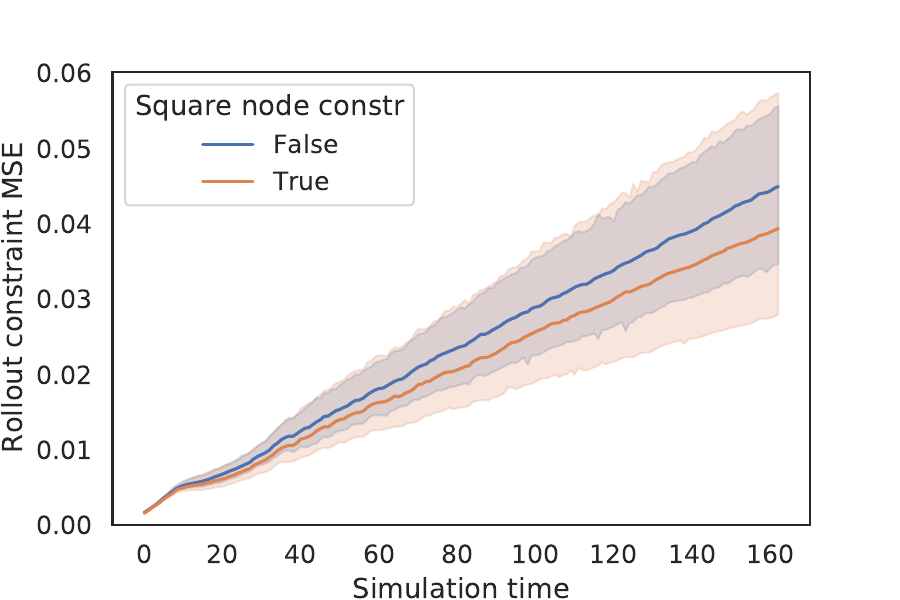}
        \caption{Squaring per-node outputs: constraint MSE}
    \end{subfigure} \hspace{\figspace}
\caption{Ablations of the modelling choices on the \dataset{Rope} dataset.}
    \label{fig:ablations}
\end{figure}


\end{document}